%% file: main_neurips_2025.tex
\documentclass{article}

\usepackage[final]{neurips_2025}

\usepackage[utf8]{inputenc} 
\usepackage[T1]{fontenc}    
\usepackage{hyperref}       
\usepackage{url}            
\usepackage{booktabs}       
\usepackage{amsfonts}       
\usepackage{nicefrac}       
\usepackage{microtype}      

\usepackage{xcolor}         
\input{tex/plots}

\usepackage[export]{adjustbox}

\input{tex/abbrev}
\input{tex/macro}
\usepackage{multirow}
\usepackage{enumitem}
\usepackage[normalem]{ulem}
\usepackage{xspace}
\usepackage{amsmath}

\usepackage{tikz}
\usepackage{pgfplots}
\usepackage{xspace}
\usepackage{amsmath}
\usepackage{colortbl}
\usepackage{enumitem}
\usepackage[normalem]{ulem}
\usepackage{wrapfig}
\usepackage{caption} 
\usepackage{graphicx}
\usepackage{multirow}
\usepackage{etoc}

\usepackage{makecell}
\usepackage{pifont}%
\newcommand{\cmark}{\ding{51}}%
\newcommand{\xmark}{\ding{55}}%

\newcommand{\parag}[1]{\smallskip\noindent\textbf{#1}\enspace}

\title{\ours: Looking into the Future with DINO}

\author{Efstathios Karypidis$^{1,3}$ \hspace{0.5em}
Ioannis~Kakogeorgiou$^{1}$ \hspace{0.5em} Spyros~Gidaris$^{2}$ \hspace{0.5em}  Nikos~Komodakis$^{1,4,5}$ \vspace{0.5em} 
\\
$^1$Archimedes, Athena Research Center, Greece \hspace{1.0em} $^2$valeo.ai \\
$^3$National Technical University of Athens \hspace{1.0em} $^4$University of Crete \hspace{1.0em} $^5$IACM-Forth
}

\begin{document}

\maketitle

\input{sec/0_Abstract}    
\input{sec/1_Intro}

\input{sec/2_Related}
\input{sec/3_Methodology}

\input{sec/4_Experiments}

\input{sec/5_Conclusion}

\input{sec/acknowledgements}

{
    \small
    \bibliographystyle{ieeenat_fullname}
    \bibliography{main}
}

\newpage
\appendix
\input{sec/supplementary}

\end{document}

%% file: tex/plots.tex
\usepackage[dvipsnames,svgnames,x11names]{xcolor}
\usepackage{tikz}
\usetikzlibrary{arrows.meta,shapes,calc,matrix,fit,positioning,backgrounds,decorations.markings}
\usepackage{pgfplots}
\usepackage{pgfplotstable}
\pgfplotsset{compat=1.9}
\usepackage{xstring}

\usepgfplotslibrary{external}

\IfBeginWith*{\jobname}{fig/extern/}{\finalcopy}{}

\tikzstyle{every picture}+=[
	remember picture,
	every text node part/.style={align=center},
	every matrix/.append style={ampersand replacement=\&},
]
\tikzstyle{tight} = [inner sep=0pt,outer sep=0pt]
\tikzstyle{node}  = [draw,circle,tight,minimum size=12pt,anchor=center]
\tikzstyle{op}    = [draw,circle,tight]
\tikzstyle{dot}   = [fill,draw,circle,inner sep=1pt,outer sep=0]
\tikzstyle{pt}    = [fill,draw,circle,inner sep=1.5pt,outer sep=.2pt]
\tikzstyle{box}   = [draw,rectangle,inner sep=3pt]
\tikzstyle{high}  = [black!60]
\tikzstyle{group} = [high,box,opacity=.5]
\tikzstyle{dim1}  = [fill opacity=.3,text opacity=1]
\tikzstyle{dim2}  = [fill opacity=.5,text opacity=1]
\tikzstyle{dim3}  = [fill opacity=.7,text opacity=1]
\tikzstyle{rectc} = [tight,transform shape]
\tikzstyle{rect}  = [rectc,anchor=south west]

\tikzset{every mark/.append style={solid}}
\pgfplotsset{
	grid=both, width=\columnwidth, try min ticks=5,
	every axis/.append style={font=\small},
	every axis plot/.append style={thick,mark=none,mark size=1.8,tension=0.18},
	legend cell align=left, legend style={fill opacity=0.8},
	xticklabel={\pgfmathprintnumber[assume math mode=true]{\tick}},
	yticklabel={\pgfmathprintnumber[assume math mode=true]{\tick}},
	nodes near coords math/.style={
		nodes near coords={\pgfmathprintnumber[assume math mode=true]{\pgfplotspointmeta}},
	},
}

\pgfplotsset{
	dash/.style={mark=o,dashed,opacity=0.6},
	dott/.style={mark=o,dotted,opacity=0.6},
	nolim/.style={enlargelimits=false},
	plain/.style={every axis plot/.append style={},nolim,grid=none},
}

\pgfdeclarelayer{bg4}
\pgfdeclarelayer{bg3}
\pgfdeclarelayer{bg2}
\pgfdeclarelayer{bg1}
\pgfdeclarelayer{fg1}
\pgfdeclarelayer{fg2}
\pgfdeclarelayer{fg3}
\pgfdeclarelayer{fg4}
\pgfsetlayers{bg4,bg3,bg2,bg1,main,fg1,fg2,fg3,fg4}

\pgfkeys{/tikz/.cd, aspect/.store in=\aspect, aspect=1}
\pgfkeys{/tikz/.cd, depth/.store in=\depth, depth=.5}
\pgfkeys{/tikz/.cd, stepx/.store in=\step, stepx=1}

\tikzstyle{geom} = [line join=bevel,aspect=1,depth=.5,z={(\depth*\aspect,\depth)}]
\tikzstyle{wire} = [geom,draw,thick]

\def\cx[#1,#2,#3]{#1}
\def\cy[#1,#2,#3]{#2}
\def\cz[#1,#2,#3]{#3}
\def\ex[#1,#2,#3]{#1,0,0}
\def\ey[#1,#2,#3]{0,#2,0}
\def\ez[#1,#2,#3]{0,0,#3}



%% file: tex/abbrev.tex
\newcommand{\Th}[1]{\textsc{#1}}
\newcommand{\mr}[2]{\multirow{#1}{*}{#2}}
\newcommand{\mc}[2]{\multicolumn{#1}{c}{#2}}

\newcommand{\red}[1]{{\textcolor{red}{#1}}}

\newcommand{\citeme}[1]{\red{[XX]}}
\newcommand{\refme}[1]{\red{(XX)}}

\makeatletter
\newcommand*\bdot{\mathpalette\bdot@{.7}}
\newcommand*\bdot@[2]{\mathbin{\vcenter{\hbox{\scalebox{#2}{$\m@th#1\bullet$}}}}}
\makeatother

\makeatletter
\DeclareRobustCommand\onedot{\futurelet\@let@token\@onedot}
\def\@onedot{\ifx\@let@token.\else.\null\fi\xspace}

\makeatother

%% file: tex/macro.tex
\newcommand{\ours}{\texttt{DINO-Foresight}\xspace}

\definecolor{ForestGreen}{RGB}{34,139,34}
\definecolor{Orange}{RGB}{1.0, 0.55, 0.0}
\definecolor{Purple}{RGB}{0.58, 0.44, 0.86}
\definecolor{VibrantMagenta}{RGB}{255,0,197}
\definecolor{DarkenedMagenta}{RGB}{225,0,167}

\definecolor{TableColor}{rgb}{0.882, 0.945, 0.882}

\newcolumntype{C}[1]{>{\centering\arraybackslash}p{#1}}

\newenvironment{customlegend}[1][]{%
    \begingroup
    \csname pgfplots@init@cleared@structures\endcsname
    \pgfplotsset{#1}%
}{%
    \csname pgfplots@createlegend\endcsname
    \endgroup
}%
\def\addlegendimage{\csname pgfplots@addlegendimage\endcsname}

\newcommand{\secref}[1]{\hyperref[#1]{Sec.~\ref{#1}}}
\newcommand{\sectionref}[1]{\hyperref[#1]{Section~\ref{#1}}}

\newcommand{\supplautoref}[1]{\hyperref[#1]{Suppl.~\autoref{#1}}}
\newcommand{\supplsec}[1]{\hyperref[#1]{Suppl. Sec.~\ref{#1}}}



%% file: sec/0_Abstract.tex
\begin{abstract}
Predicting future dynamics is crucial for applications like autonomous driving and robotics, where understanding the environment is key. Existing pixel-level methods are computationally expensive and often focus on irrelevant details. 
To address these challenges, we introduce \ours, a novel framework that operates in the semantic feature space of pretrained Vision Foundation Models (VFMs). Our approach trains a masked feature transformer in a self-supervised manner to predict the evolution of VFM features over time. By forecasting these features, we can apply off-the-shelf, task-specific heads for various scene understanding tasks. In this framework, VFM features are treated as a latent space, to which different heads attach to perform specific tasks for future-frame analysis. Extensive experiments show the very strong performance, robustness and scalability of our framework. Project page and code at \href{https://dino-foresight.github.io/}{https://dino-foresight.github.io/}.
\end{abstract}

%% file: sec/1_Intro.tex
\section{Introduction}
\label{sec:intro}

Predicting future states in video sequences is a key challenge in computer vision and machine learning, with important applications in autonomous systems like self-driving cars and robotics \citep{ULphysinter,dosovitskiy2017learning}. These systems must navigate dynamic environments safely, yet predicting future states remains difficult—especially in complex scenarios involving multi-object interactions over long time horizons.

Recent approaches focus on generating realistic RGB future frames using latent generative modeling \citep{rombach2022high}. These methods first compress RGB data into a latent space, such as continuous \citep{rombach2022high} or discrete \citep{esser2021taming} Variational Autoencoder (VAE) representations. Then, they train generative models—like diffusion models \citep{zheng2024genad,gao2024vista,ho2022imagen,ho2022video,harvey2022flexible,videoworldsimulators2024}, autoregressive models \citep{hu2023gaia,hong2022cogvideo,kondratyuk2024videopoet,wang2024emu3}, or masked video generation \citep{yu2023magvit,yu2024language}—to predict future states in this compressed space. While this reduces dimensionality and improves training stability \citep{rombach2022high}, VAE latents often lack semantic alignment, making them hard to interpret or use in downstream scene understanding tasks (see \autoref{tab:ablation_vfm_backbones}). Moreover, these methods must model both low-level appearance and high-level semantics, even though decision-making systems (e.g., self-driving cars) primarily need semantic scene understanding—what objects exist and where they are. Latent generative approaches may waste capacity on irrelevant low-level details, compromising temporal semantic accuracy.

In contrast, vision foundation models (VFMs) have revolutionized scene understanding with robust, transferable features \citep{oquab2024dinov,radford2021learning,venkataramanan2025franca, bardes2024revisiting}. This raises a key question: \emph{can VFM features serve as a semantically rich, high-dimensional latent space for precise future prediction?}

\begin{figure}[t]
\centering
\includegraphics[trim={0cm 0cm 0cm 0cm},width=0.89\linewidth]{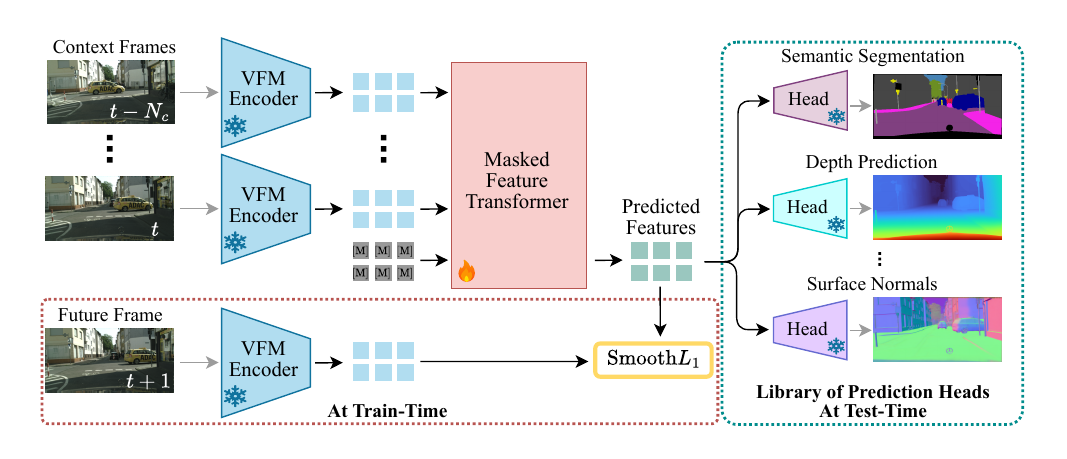}
\vspace{-4pt}
\caption{
\textbf{Forecasting VFM Features for Future Frames.}
At the core of our approach is the prediction of future VFM feature evolution. To this end, we train a masked  transformer model in a self-supervised manner to forecast these features from context frames, minimizing SmoothL1 loss between predicted and actual future features. By forecasting these rich and versatile features, task-specific prediction heads—such as semantic segmentation, depth, and surface normals—can be effortlessly employed at test time, enabling modular and efficient multi-task scene understanding.}
\label{fig:overview}
\end{figure}

In this work, we propose precisely this idea. Instead of predicting future low-level VAE latents, we forecast the temporal evolution of VFM features directly. This shift brings several important advantages: 
\textbf{(a) Beyond low-level latents.} It leverages semantically meaningful representations, inheriting strong scene understanding.  
\textbf{(b) Dynamic semantics over raw frames.} It avoids full-frame synthesis, letting the model focus on meaningful dynamics, reducing complexity and sidestepping challenges like multimodal pixel distributions. 
\textbf{(c) Scalable multi-task support and modular integration.} Forecasted features align with downstream tasks, enabling plug-and-play integration with pretrained classifiers, segmenters or task-specific heads—without the need to retrain the core feature predictor (see~\autoref{fig:overview}).
This represents a significant departure from prior semantic feature prediction methods~\citep{luc2017predicting}, which face significant challenges with multi-task scalability. Such approaches either require training a separate model for each task~\citep{nabavi2018future,chiu2020segmenting,terwilliger2019recurrent} or involve predicting features from multiple task-specific models simultaneously~\citep{hu2020probabilistic, karypidis2025advancing}, resulting in more complex and less scalable architectures.

In this work we show that forecasting high-dimensional VFM features is not only feasible but also achieves strong performance, offering a new path toward semantically grounded future prediction.

In summary, our contributions are:  
\textbf{(1)} We introduce \ours, a self-supervised method for semantic future prediction that builds on the key idea of forecasting VFM features—our core contribution. Unlike latent generative methods that rely on low-level VAE features, our approach delivers precise future scene understanding without modeling unnecessary appearance details, enabling a unified model for diverse scene understanding tasks.
\textbf{(2)} To realize this idea, we design an efficient masked feature transformer (see~\autoref{fig:overview}) that propagates multi-layer, high-resolution features critical for achieving strong task performance.
\textbf{(3)} 
Experimental results demonstrate a unique advantage of our approach - our single model successfully handles multiple future-frame understanding tasks (semantic segmentation, instance segmentation, depth prediction, and surface normal prediction) where previous approaches required multiple specialized models.
Furthermore, as we show in \hyperref[sec:intermediate_features]{Appendix Subsection~\ref*{sec:intermediate_features}} intermediate features within our masked transformer can further improve downstream task performance, highlighting its potential as a self-supervised learning strategy that enhances the already strong VFM features.

%% file: sec/2_Related.tex
\section{Related Work}
\label{sec:related}

\parag{Video Prediction} is an extensively studied problem.
 Early approaches based Convolutional LSTMs \citep{ nabavi2018future,Xu_2018_CVPR, wang2018eidetic,castrejon2019improved, lee2021video, wu2021motionrnn,gao2022simvp} struggled to maintain both visual quality and temporal consistency. Subsequent developments introduced generative adversarial networks and variational autoencoders \citep{yan2021videogpt, vondrick2016generating,castrejon2019improved, lee2018stochastic,   babaeizadeh2018stochastic}, alongside diffusion models \citep{ho2022imagen, ho2022video, gao2024vista,
harvey2022flexible}, to enhance spatial-temporal coherence and improve the quality of predictions. Furthermore, transformer-based models have also been adapted to videos, utilizing auto-regressive and masked modeling objectives to capture video dynamics \citep{yu2024language,yu2023magvit,gupta2023maskvit,wang2024emu3}.

\parag{Latent Generative Models} 
Current state-of-the-art video prediction methods~\citep{gupta2023maskvit,yu2023magvit,yu2024language,gao2024vista} build upon latent generative approaches \citep{rombach2022high}. These frameworks employ an autoencoder to compress RGB frames into a latent space, then train generative models to predict future states in this compressed representation. While sharing superficial similarities with our approach—including the use of latent spaces and masked transformers in some cases~\citep{gupta2023maskvit,yu2023magvit}—these methods differ fundamentally. Their latent spaces primarily encode low-level visual information, requiring reconstruction back to RGB space and forcing the model to simultaneously handle both appearance details and semantic changes.

Our key innovation lies in forecasting VFM features that natively encode high-level semantic information, enabling direct application of task-specific prediction heads. As demonstrated in \autoref{tab:ablation_vfm_backbones}, conventional VAE latent spaces cannot match this capability. Our experimental results in \autoref{tab:comparison_with_prior_work} further show these methods' limitations in predicting semantic scene evolution, highlighting the distinct advantages of VFM feature forecasting for scene understanding tasks. Concurrently to our work, DINO-WM \citep{zhou2025dinowm}, also leverages DINOv2 features for world modeling with action-conditioned planning in simulated environments, while our work targets multi-task dense semantic forecasting in real world scenarios.

\parag{Semantic Future Prediction}  
An emerging approach for future-frame prediction focuses on forecasting intermediate features from an encoder rather than raw RGB values \citep{nabavi2018future, lin2021predictive, saric2020warp, sun2019predicting, chen2019multi,jin2017video, luc2018predicting, vsaric2019single, hu2021apanet, vondrick2016anticipating, zhong2023anticipative}. This strategy models abstract encoder representations, which task-specific heads use for downstream tasks. Early methods in this paradigm include F2F~\citep{luc2018predicting}, which regresses Mask-RCNN’s feature pyramid, and F2MF~\citep{saric2020warp}, which improves feature prediction using flow-based warping. APANet~\citep{hu2021apanet} aggregates multi-level features via an auto-path mechanism for instance segmentation, while PFA~\citep{lin2021predictive} enhances global structures and suppresses local details for more predictable features. Recently, Futurist \citep{karypidis2025advancing} introduced a multi-modal semantic forecasting approach for semantic segmentation and depth maps. While effective, these methods often rely on task- or dataset-specific encoders, limiting practicality and scalability. To address this, we use VFM encoders, which, due to large-scale pre-training, perform well across diverse tasks and generalize effectively to new scenes without retraining.

\parag{Vision Foundation Models (VFMs)}  
VFMs have transformed computer vision, achieving strong performance across a range of visual tasks. Trained on large-scale datasets, these models learn rich, transferable visual representations. Notable examples include DINOv2 \citep{dino, oquab2024dinov}, a self-supervised model based on self-distillation; Franca \citep{venkataramanan2025franca}, a fully-open VFM for scalable self-supervised representation learning; CLIP and its variants \citep{radford2021learning, fang2023eva, sun2023eva, fang2024eva}, which align visual representations with natural language; SAM (Segment Anything Model) \citep{kirillov2023segment}, a foundation model for image segmentation and V-JEPA\citep{bardes2024revisiting}, a video representation learning method via feature prediction in latent space. In this work, we explore the potential of VFM features for semantic future prediction tasks, connecting static visual understanding with dynamic prediction.

\parag{Multi-Task Learning}
Multi-Task Learning (MTL) is a learning paradigm that enables simultaneous training of models on multiple related tasks~\citep{MRK19,Misra2016cross,adamtl2023,9336293}, promoting shared representations and improving performance across tasks. Traditional MTL frameworks often use parameter sharing~\citep{Kendall2018uncertainty,NEURIPS2018_432aca3a,bekoulis-etal-2018-adversarial} or task interaction allowing exchange of information~\citep{bragman2019stochastic,chen2023adamv,chen2023modsquad,Misra2016cross,Ruder_Bingel_Augenstein_Søgaard_2019}. Other approaches employ a strategy that incrementally increases the model's depth during training, enabling the network to learn task-specific representations in a more resource-efficient way~\cite{aich2023efficient,choi2023dynamic,pmlr-v119-guo20e,8099609,NEURIPS2022_dd3bd4e3}. Recently, the emergence of large-scale pretrained models has led to the introduction of adapter-based multi-task fine-tuning approaches~\citep{liang2022effective,NEURIPS2022_efb02f96}. In our work, we leverage VFM features to provide a unified, scalable and modular framework for future prediction. Our approach enables seamless integration of multiple tasks without retraining or complex adaptations.

%% file: sec/3_Methodology.tex
\section{Methodology}
\label{sec:method}

Our semantic future prediction approach builds on forecasting VFM features – powerful representations that excel at scene understanding tasks and generalize well to unseen environments. We realize this through a masked transformer model trained via self-supervision to predict the temporal evolution of VFM features. This forecasted feature space serves as a latent representation that can flexibly connect to various off-the-shelf task-specific heads for future-frame analysis.

\autoref{fig:overview} provides an overview of our approach, with the key components detailed in the following sections: \textbf{\sectionref{sec:target_features}} describes how target features are generated from multi-layer VFM features. \textbf{\sectionref{sec:training}} presents our formulation of VFM feature forecasting as a masked feature modeling problem and details the model architecture. \textbf{\sectionref{sec:high_res}} covers efficient training techniques for high-resolution VFM feature prediction. Finally, \textbf{\sectionref{sec:head}} introduces our modular framework for multi-task future-frame analysis and details how prediction heads are trained and integrated.

\subsection{Hierarchical Target Feature Construction}\label{sec:target_features}

\begin{wrapfigure}{r}{0.5\textwidth}
\vspace{-8pt}
\centering
\includegraphics[trim={0cm 0cm 0cm 0cm}]{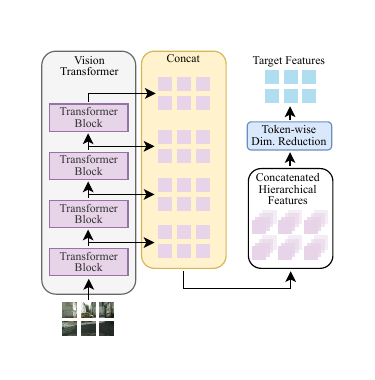}
\vspace{-4pt}
\caption{
\textbf{Hierarchical Target Feature Construction for the Feature Prediction Model.} Our framework constructs a feature space by extracting and concatenating multi-layer features from a frozen ViT encoder, capturing semantic information at varying abstraction levels. PCA is applied to reduce dimensionality, creating compact features.} 
\label{fig:hierarchical}
\vspace{-10pt}
\end{wrapfigure}

In~\autoref{fig:hierarchical}, 
we provide an overview of how the target feature space for the feature prediction model is constructed. 
Below, we outline the main steps involved.

\parag{Multi-Layer VFM Feature Extraction}
Scene understanding models often benefit from processing features across multiple layers of an image encoder \citep{ranftl2021vision, Long_2015_CVPR,lin2017feature, cheng2022masked,zhao2017pyramid,chen2017deeplab}, especially when the encoder is frozen, as in our approach. To fully leverage the pretrained VFMs, we propagate features extracted from 
multiple layers of the VFM. 
We ablate the impact of multi-layer features in \hyperref[sec:impact_multi_features]{Appendix Subsection~\ref*{sec:impact_multi_features}}.

Our framework uses VFMs based on the Vision Transformer (ViT) architecture, though the approach can be extended to other architectures. 
Given a sequence of $N$ image frames $\mathbf{X} \in \mathbb{R}^{N \times H' \times W' \times 3}$, let $\mathbf{F}^{(l)} \in \mathbb{R}^{N \times H \times W \times D_{enc}}$ represent the features extracted from layer $l$ of the ViT model. Here, $D_{enc}$ is the feature dimension, and $H \times W$ is the spatial resolution, which are consistent across layers in ViT-based models. 
To form the target feature space on which the feature prediction model operates,
we concatenate the features from $L$ layers along the channel dimension, resulting in $\mathbf{F}_{\text{concat}} \in \mathbb{R}^{N \times H \times W \times L \cdot D_{enc}}$. These concatenated features capture rich semantic information from the input images at multiple levels of abstraction.

\parag{Dimensionality Reduction}
The concatenated features $\mathbf{F}_{\text{concat}}$ have high dimensionality, so we apply dimensionality reduction to simplify the feature prediction task 
while retaining essential information. In this work, we use Principal Component Analysis (PCA) 
to reduce the dimensionality, transforming $\mathbf{F}_{\text{concat}}$ into a lower-dimensional representation $\mathbf{F}_{\text{PCA}} \in \mathbb{R}^{N \times H \times W \times D}$, where $D \ll L \cdot D_{enc}$. These PCA-reduced features, $\mathbf{F}_{\text{PCA}}$, serve as the target features on which the future feature prediction operates, i.e., $\mathbf{F}_{\text{TRG}} = \mathbf{F}_{\text{PCA}}$. 
In \hyperref[sec:impact_pca]{Appendix Subsection~\ref*{sec:impact_pca}}, 
we show that this PCA-based compression retains essential information and enhances downstream performance.

\subsection{Self-Supervised Future Feature Prediction with Masked Transformers} \label{sec:training}

\parag{Masked Feature Transformer Architecture}
Inspired by previous video generation models \citep{chang2022maskgit, gupta2023maskvit, karypidis2025advancing}, we implement the future feature prediction task using a self-supervised masked transformer architecture. The task involves predicting future frames in a video sequence consisting of $N$ frames, where $N_c$ are context frames and $N_p$ are future frames to be predicted, such that $N = N_c + N_p$.  Given the target features $\mathbf{F}_{\text{TRG}} \in \mathbb{R}^{N \times H \times W \times D}$ for these $N$ frames, 
the future-frame tokens are masked, and the transformer must predict these missing tokens by processing all the tokens from the entire sequence, i.e., all the $N \cdot H \cdot W$ tokens. 

The architecture begins with a token embedding stage, where each token is projected from $D$ dimensions into the transformer’s hidden dimension $D_{dec}$ through a linear layer. During training, the tokens corresponding to the future frames are replaced with a learnable $D_{\text{dec}}$-dimensional [MASK] vector. During inference, these [MASK] tokens are appended after the context frames. Each token also receives a position embedding to retain both temporal and spatial information across the sequence.

The tokens are then passed through a series of transformer layers. Standard self-attention layers in transformers have quadratic time complexity with respect to the number of tokens, making them computationally expensive for high-resolution, multi-frame sequences. To address this, we follow the approach from recent video transformers \citep{arnab2021vivit, gupta2023maskvit, karypidis2025advancing} and decompose the attention mechanism into temporal and spatial components. Temporal attention is applied across tokens with the same spatial position in different frames, capturing the dynamics over time. Spatial attention, on the other hand, operates within individual frames, focusing on spatial interactions. Thus, each transformer layer consists of a temporal Multi-Head Self-Attention (MSA) layer, a spatial MSA layer, and a feedforward MLP layer. After passing through the transformer layers, a linear prediction layer maps the output token embeddings from the hidden dimension $D_{\text{dec}}$ back to the feature dimension $D$, producing the predicted feature map $\mathbf{\tilde{F}}_{\text{TRG}} \in \mathbb{R}^{N \times H \times W \times D}$.

\parag{Training Objective}
We frame the future-frame prediction as a continuous regression problem and optimize a self-supervised training objective based on the SmoothL1 loss between the predicted features $\mathbf{\tilde{F}}_{\text{TRG}}$ and the ground truth features $\mathbf{F}_{\text{TRG}}$ at the masked locations.
The loss is defined as:
\begin{equation} \label{eq:mfm_loss}
\mathcal{L}_{\mathrm{MFM}} = \underset{x \in \mathcal{X}}{\mathbb{E}}\left[\sum_{p \in \mathcal{P}} L\left(\mathbf{F}_{\text{TRG}}(p), \mathbf{\tilde{F}}_{\text{TRG}}(p)\right)\right],
\end{equation}
where $\mathcal{X}$ denotes the training dataset,
$\mathcal{P} =[N_c+1:N] \times [1:H] \times [1:W]$ 
represents the set of positions across the $H \times W$ spatial dimensions of the $N_p$ future frames, and 
$\mathbf{F}_{\text{TRG}}(p)$ and $\mathbf{\tilde{F}}_{\text{TRG}}(p)$ are the ground truth and predicted feature vectors at position $p$, respectively. 
$L(\cdot, \cdot)$ computes the SmoothL1 loss between two feature vectors:
\begin{equation} \label{eq:smooth_loss}
L(x, y) = \sum_{d=1}^{D} \begin{cases} 
0.5 \frac{(x_d - y_d)^2}{ \beta}, & \text{if } |x_d - y_d| < \beta, \\
|x_d - y_d| - 0.5 \beta, & \text{otherwise.}
\end{cases}
\end{equation}
In our experiments, we set $\beta=0.1$.

\subsection{Compute-Efficient Training Strategies for High-Resolution Feature Forecasting}\label{sec:high_res}

Using high-resolution features is crucial for pixel-wise scene understanding tasks, such as segmentation or depth prediction, where low-resolution features struggle to capture small objects or fine spatial structures \citep{ranftl2021vision, 
cheng2022masked,chen2017deeplab,touvron2019fixing}. To achieve good performance on these tasks, we aim to forecast VFM features extracted from frames with a spatial resolution of \( H' \times W' = 448 \times 896 \). For a ViT with a patch size of \( 14 \times 14 \), as used in DINOv2~\citep{oquab2024dinov} and EVA-CLIP~\citep{fang2024eva}, this results in feature maps with a resolution of \( H \times W = 32 \times 64 \), corresponding to 2048 tokens per frame.

However, training ViTs on such high-resolution inputs is computationally expensive in terms of both time and memory \citep{dosovitskiy2020image}. To address this challenge while maintaining efficient training, we explore the following strategies:

\parag{Low-Resolution Training with High-Resolution Inference}
In this approach, we train on frames with a lower resolution of \( 224 \times 448 \), resulting in features with a resolution of \( 16 \times 32 \). During testing, we use high-resolution frames (\( 448 \times 896 \)) and adapt the position embeddings through interpolation. However, this strategy leads to suboptimal performance due to a distribution shift between the training and test data, which causes inaccurate feature forecasting.

\parag{Sliding-Window Approach for High-Resolution Inference}
Inspired by sliding-window techniques used in segmentation tasks~\citep{strudel2021segmenter}, this strategy trains the model with cropped feature maps. The ViT encoder extracts features from high-resolution frames (\( 448 \times 896 \)), producing high-resolution tokens (e.g., \( 32 \times 64 \) for a patch size of \( 14 \times 14 \)). During training, we sample local crops of size \( 16 \times 32 \), taken 
from the same spatial locations across frames. The model is trained on these cropped features. At inference time, the model processes the high-resolution features in a sliding-window manner using the same crop dimensions. This approach efficiently handles large inputs while avoiding the computational cost of full-resolution training.

\parag{Two-Phase Training with Resolution Adaptation}
This strategy employs a two-phase training process~\citep{oquab2024dinov, touvron2019fixing,kolesnikov2020big}. First, the model is trained on low-resolution frames (\( 224 \times 448 \)) for several epochs, focusing on learning broad feature forecasting. Then, the model is fine-tuned on high-resolution (\( 448 \times 896 \)) for a small number of epochs. This adaptation phase improves the model's ability to handle high-resolution features without incurring the computational cost of training from scratch at the higher resolution. As shown in our experiments, both strategies are effective, but the two-phase approach yields better feature forecasting performance. This is likely because the masked transformer has access to a larger spatial context when propagating VFM features in future frames.

\subsection{Modular Framework for Future-Frame Multi-Task Predictions}\label{sec:head}

Our framework supports a library of interchangeable task-specific prediction heads that operate on the preficted future freames, enabling flexible multi-task future understanding. The design is modular: each head operates independently, allowing tasks to be added or removed without retraining the core feature prediction model.

We focus on four pixel-wise prediction tasks: semantic segmentation, instance segmentation, depth prediction, and surface normals prediction. However, our framework is general and can support other scene understanding tasks, such as object detection and panoptic segmentation. 
For semantic segmentation, depth prediction, and surface normals prediction, we use the Dense Prediction Transformer~\citep{ranftl2021vision} (DPT) architecture, which is well-suited to our setup. 
DPT leverages multi-layer features from ViT-based encoders—consistent with the features predicted by our model—and progressively refines them into high-resolution predictions using convolutional fusion.
For the instance segmentation task, we use a Mask2Former~\citep{cheng2022masked} head.

Prediction heads can be trained directly on frozen VFM features and then applied ``off-the-shelf'' to future-frame features predicted by the model. Additionally, they can be trained to account for the PCA stage by applying PCA compression and decompression to the multi-layer features. This approach is useful for cases where prediction heads are trained on annotated 2D images, without requiring video data, and then added to the library for future-frame predictions.

%% file: sec/4_Experiments.tex
\section{Experiments} \label{sec:experiments}

\subsection{Experimental setup}

\parag{Data.} 
We assess our approach using the Cityscapes~\citep{Cordts_2016_CVPR} and nuScenes~\citep{nuscenes} datasets, both offering video sequences from urban driving environments. The Cityscapes dataset includes 2,975 training sequences, 500 for validation,  each with 30 frames captured at 16 fps and a resolution of 1024 × 2048 pixels. The 20th frame in each sequence is annotated for semantic segmentation with 19 classes. The nuScenes dataset comprises of 700 training scenes and 150 validation scenes, captured at a frame rate of 12 Hz, with each sequence extending for 20 seconds.

\parag{Implementation details.} 
By default, we use DINOv2-Reg with ViT-B/14 as the default VFM visual encoder for our method.
For the masked feature transformer we built upon \citep{besnier2023pytorch} implementation. We use 12 layers with a hidden dimension of \( d = 1152 \) and sequence length \( N = 5 \) (with \( N_c = 4 \) context frames and \( N_p = 1 \) future frame). For end-to-end training, we use the Adam optimizer \citep{adamopt} with momentum parameters \( \beta_1 = 0.9 \), \( \beta_2 = 0.99 \), and a learning rate of \( 6.4 \times 10^{-4} \) with cosine annealing. Training is conducted on 8 A100 40Gb GPUs with an effective batch size of 64.
We train DPT~\citep{ranftl2021vision} heads for the semantic segmentation, depth prediction, and surface normals estimation tasks, and a Mask2Former~\citep{cheng2022masked} head for the instance segmentation task. 
More implementation details in \hyperref[sec:appendix-implemdetails]{Appendix Subsection~\ref*{sec:appendix-implemdetails}}.

\input{tables/tab_comp_semsegm_instance}

\parag{Evaluation Metrics.}
To evaluate our method's performance we use the following metrics:
For \textbf{semantic segmentation}, we use mean Intersection over Union (mIoU) in two ways: (1) mIoU (ALL), which includes all semantic classes, and (2) MO-mIoU (MO), which considers only movable object classes like person, rider, car, truck, bus, train, motorcycle, and bicycle.
For \textbf{instance segmentation}, we measure performance using average precision at a 0.50 IoU threshold (AP50) and also the mean average precision over IoU thresholds from 0.50 to 0.95.
For \textbf{depth prediction}, we use mean Absolute Relative Error (AbsRel) and depth accuracy ($\delta_{1}$). For \textbf{surface normals}, we compute the mean angular error (m$\downarrow$) and the percentage of pixels with angular errors below 11.25° (11.25°$\uparrow$). Definitions of depth and surface normals metrics are provided in \hyperref[sec:appendix-evalmetrics]{Appendix Subsection~\ref*{sec:appendix-evalmetrics}}.

\parag{Evaluation Scenarios}
Following prior work \citep{luc2017predicting, nabavi2018future}, we evaluate our model on Cityscapes for \textbf{short-term prediction} (3 frames, ~0.18s) and \textbf{mid-term prediction} (9 frames, ~0.54s). On nuScenes, which has more static scenes (i.e., with less movement) than Cityscapes, we use \textbf{mid-term prediction} (9 frames, ~0.75s) and \textbf{long-term prediction} (18 frames, ~1.5s). Further details in \hyperref[sec:appendix-evalscenarios]{Appendix Subsection~\ref*{sec:appendix-evalscenarios}}.

\parag{Baselines} 
We evaluate our method against three baselines. The first, the \textbf{\texttt{Oracle}} baseline, directly accesses the target future frame, establishing an upper performance bound. The second, \textbf{\texttt{Copy-Last}}, copies the most recent context frame to predict the target frame, providing a lower performance bound. Both baselines use DINOv2-Reg ViT-B encoder with DPT heads for semantic segmentation, depth prediction, and surface normals estimation, and a Mask2Former head for instance segmentation. The third baseline leverages \textbf{\texttt{VISTA}} \citep{gao2024vista}, a state-of-the-art world model that uses video latent diffusion to generate future RGB frames from three context frames. This is a large-scale model, comprising 2.5 billion parameters and trained on 1,740 hours of driving videos, which we fine-tune on Cityscapes and nuScenes using the same frame-rate as our model. \texttt{VISTA} generates future RGB frames (with action-free conditioning), 
which are processed by a DINOv2-Reg encoder with DPT heads (identical to other baselines)
for semantic segmentation, depth and surface normals prediction.

\subsection{VFM feature forecasting results}

\parag{Comparison with State-of-the-Art}  
\autoref{tab:comparison_with_prior_work} compares our method with state-of-the-art approaches for semantic/instance segmentation, depth estimation, and surface normal forecasting. The results highlight our key advantage: a single feature prediction model that achieves competitive or superior performance across multiple scene understanding tasks. In contrast, prior works either require separate prediction models per task \citep{luc2018predicting} or handle at most two tasks simultaneously \citep{karypidis2025advancing}. 
This demonstrates the flexibility and practicality of our VFM feature forecasting approach.

\parag{Unified Representations for Multiple Tasks: VFM Features vs. RGB Pixels}
An alternative to forecasting VFM features is predicting future frames directly in RGB space, which also supports performing multiple downstream tasks through standard scene understanding models. For comparison, we fine-tune \texttt{VISTA}~\citep{gao2024vista} to generate five future frames from three context frames (8 frames in total). We process these synthesized frames (both short-term and mid-term) with DINOv2-Reg ViT-B and DPT heads (similar to our method's setup) for fair evaluation.
Despite \texttt{VISTA}'s large-scale training and model size (2.5B parameters), it achieves lower performance on semantic segmentation, depth estimation, and surface normal prediction. Extended evaluations on nuScenes (provided 
\hyperref[tab:ablation_vista_nuscenes]{Appendix Table~\ref*{tab:ablation_vista_nuscenes}}  ) show similar trends for depth and surface normal estimation.
Our approach is also far more computationally efficient: mid-term forecasting on Cityscapes' 500 validation scenes takes approximately 5 minutes versus \texttt{VISTA}'s 8.3 hours (both on a single A100 GPU). This highlights our method's advantages of operating in VFM feature space -- achieving accurate semantic prediction while being significantly more resource-efficient.

\input{tables/tab_found_mod_tasks}

\parag{Comparison of VFM Visual Encoders} 
In~\autoref{tab:ablation_vfm_backbones}, we evaluate our method using three VFM encoders to extract features for our feature prediction model: DINOv2 with registers~\citep{darcet2024vision,oquab2024dinov} (self-supervised), EVA2-CLIP~\citep{fang2024eva} (vision-language contrastive), and SAM~\citep{kirillov2023segment} (supervised instance segmentation). For each, we use the ViT-B variant. We also include \texttt{Copy-Last} and \texttt{Oracle} baselines for comparison. The results show that:  
\textbf{(1)} DINOv2 consistently outperforms other encoders across all tasks, achieving the best results for both short- and mid-term predictions.  
\textbf{(2)} This aligns with expectations, as the DINOv2-based \texttt{Oracle} also performs best in all cases.  
\textbf{(3)} Our model effectively predicts future-frame features for all VFMs, significantly improving over the \texttt{Copy-Last} baseline. 
Based on these findings, we select DINOv2 as our default VFM encoder.

\input{tables/tab_cont_disc_representations}
\input{tables/tab_highres}

\parag{Future Prediction: VFM features vs VAE-based latents} Additionally, in~\autoref{tab:ablation_vfm_backbones}, we evaluate using VAE latents~\citep{rombach2022high} (used in latent generative models) instead of VFM features, training DPT prediction heads on these latents. Results are significantly worse, as expected, since these latents lack high-level information, and even DPT oracles perform poorly. This highlights the advantage of forecasting VFM features over VAE latents.

\parag{Discrete vs. Continuous VFM Representations} Recent work in generative modeling has explored both discrete and continuous representations for image and video generation \citep{chang2022maskgit,yu2023magvit,yan2021videogpt,vqvae_2,li2024autoregressive,tschannen2024givt}. Recent findings favor continuous representations \citep{li2024autoregressive,tschannen2024givt}, showing that removing vector quantization can improve generation quality while retaining the benefits of sequence modeling. To further investigate this in the context of VFM feature forecasting, we employ 4M’s pretrained DINOv2 tokenizer \citep{4m21}, which encodes DINOv2 features (without register tokens) into discrete codes from a vocabulary of size 8192. We train a quantized variant of DINO-Foresight (ViT-B14, single-layer features) using a cross-entropy loss to predict these discrete codes, which are decoded back to DINOv2 features at inference time. Results are reported in~\autoref{tab:continuous_vs_discrete}. While the discretized variant achieves comparable oracle performance to the continuous VFM feature case, our continuous VFM feature forecasting approach yields superior future semantic prediction results. These findings suggest that preserving the rich, continuous representations from VFMs—without quantization—offers clear advantages for dense semantic forecasting tasks.

\parag{Training-Efficient Strategies for High-Resolution Feature Forecasting} 
In~\autoref{tab:highres-results}, we compare the resolution-adaptation strategies from~\sectionref{sec:high_res}, reporting results for future semantic segmentation. \textit{High-resolution features are essential}: comparing the low-resolution Oracle baseline (model (a)) with the high-resolution Oracle baseline (model (b)) highlights the importance of high-resolution features for strong segmentation performance. Consequently, forecasting low-resolution features (model (c)) results in significantly poorer segmentation than models predicting high-resolution features (models (e) and (f)).
Adapting a model trained on low-resolution features for high-resolution inputs by simply adjusting position embeddings during inference (model (d)) leads to suboptimal results, even underperforming compared to low-resolution forecasting. The other two adaptation strategies—Sliding Window (model (e)) and two-phase training with resolution increase (model (f))—achieve considerably better results, demonstrating their effectiveness. 
The two-phase approach is simpler and yields the best performance, so we adopt it as our default for high-resolution feature forecasting.

\parag{Additional Ablations and Analysis} We conduct comprehensive ablation studies in the appendix to further validate our approach. 
First, in \hyperref[sec:AdditionalVista]{Appendix Subsection~\ref*{sec:AdditionalVista}}, 
we demonstrate strong zero-shot generalization by training on Cityscapes and evaluating on nuScenes without fine-tuning. The resulting performance is only slightly worse than models trained directly on nuScenes, while surpassing all baselines. 
Second, in \hyperref[sec:Masking_LossF_ablations]{Appendix Subsection~\ref*{sec:Masking_LossF_ablations}}, 
we investigate the impact of different masking strategies (random vs. full masking) and loss functions (L1, MSE, SmoothL1, and SmoothL1 with cosine similarity). 
Third, in \hyperref[sec:Model_Data_Scalability]{Appendix Subsection~\ref*{sec:Model_Data_Scalability}}, 
we analyze scalability across model sizes (Small: 115M, Base: 258M, Large: 460M parameters) and data scale (Cityscapes alone vs. Cityscapes+nuScenes combined), demonstrating consistent performance improvements with increased capacity and data diversity. These results highlight the promising robustness and scalability of our VFM feature forecasting approach.

\subsection{Qualitative results.}

In~\autoref{fig:qualitative-results}, we present qualitative results from our method applied to semantic segmentation, depth estimation, and surface normal prediction tasks, with both short-term and mid-term future predictions. Our single VFM feature prediction model produces meaningful outputs across all tasks, demonstrating the benefits of leveraging the feature space of large-scale pre-trained VFMs for future prediction.

\begin{figure}[tb]
    \centering
    \include{sec/fig_qualitative_results}
    \vspace{-4pt}
\caption{\textbf{Future predictions for semantic segmentation, depth, and surface normals.}
Noisy segmentations at the bottom of the image (in both predicted and Oracle results) are due to unannotated regions in Cityscapes that are ignored during DPT training. This artifact affects only segmentation, not the predicted features, as evident in the clear depth and surface normal predictions.
}
    \label{fig:qualitative-results}
\vspace{-4pt}
\end{figure}

%% file: tables/tab_comp_semsegm_instance.tex
\newcommand{\graycell}[1]{\textcolor{gray}{#1}}
\begin{table}[t]
\scriptsize
\centering
\setlength{\tabcolsep}{1.95pt} 
\caption{\textbf{Comparison with state-of-the-art on multiple forecasting tasks.} Methods that do not handle a task are marked with `-'. For approaches requiring separate prediction models per task (e.g., \texttt{PFA}), we show multiple entries (semantic/instance).
ALL: mIoU of all classes. MO: mIoU of movable objects. \texttt{VISTA}$_{ft}$ is the \texttt{VISTA} model fine-tuned on Cityscapes. Compared approaches include: \texttt{3Dconv-F2F-RGB}~\citep{chiu2020segmenting}, \texttt{Dil10-S2S}~\citep{luc2017predicting}, \texttt{F2F}~\citep{luc2018predicting}, \texttt{ConvLSTM}~\citep{chiu2020segmenting}, \texttt{FeatReproj3D}~\citep{FeatReproj3D}, \texttt{Bayesian S2S}~\citep{Bayesian_s2s}, \texttt{3Dconv-F2F-SEG}~\citep{chiu2020segmenting}, \texttt{DeformF2F}~\citep{vsaric2019single}, \texttt{LSTM AM S2S}~\citep{chen2019multi}, \texttt{APANet}~\citep{hu2021apanet}, \texttt{LSTM M2M}~\citep{terwilliger2019recurrent}, \texttt{IndRNN-Stack}~\citep{indrnn_stack}, \texttt{DiffAttn-Fuse}~\citep{graber2022joint}, \texttt{F2MF}~\citep{saric2020warp}, and \texttt{PFA}~\citep{lin2021predictive}, \texttt{Futurist} \citep{karypidis2025advancing} and \texttt{VISTA} \citep{gao2024vista}.}

\vspace{2pt}

\begin{tabular}{l 
    C{0.7cm} C{0.7cm} C{0.7cm} C{0.7cm} 
    C{0.7cm} C{0.7cm} C{0.7cm} C{0.7cm} C{0.8cm} C{0.8cm} C{1.0cm}  C{1.0cm} 
}
\toprule
\multirow{3}{*}[-1.0ex]{\Th{Method}} 
& \multicolumn{4}{c}{\Th{Semantic Segmentation}} 
& \multicolumn{4}{c}{\Th{Instance Segmentation}} & \multicolumn{2}{c}{\Th{Depth}} & \multicolumn{2}{c}{\Th{Surface Normals}} \\
\cmidrule(lr){2-5} \cmidrule(lr){6-9} \cmidrule(lr){10-11} \cmidrule(lr){12-13}
& \multicolumn{2}{c}{\Th{Short}} & \multicolumn{2}{c}{\Th{Mid}} 
& \multicolumn{2}{c}{\Th{Short}} & \multicolumn{2}{c}{\Th{Mid}}
& \Th{Short} & \Th{Mid}
& \Th{Short} & \Th{Mid}\\
\cmidrule(lr){2-3} \cmidrule(lr){4-5} \cmidrule(lr){6-7} \cmidrule(lr){8-9} \cmidrule(lr){10-10} \cmidrule(lr){11-11} \cmidrule(lr){12-12} \cmidrule(lr){13-13}

& ALL & MO & ALL & MO 
& AP50 & AP & AP50 & AP & $\delta_{1}$ & $\delta_{1}$ &  11.25$^\circ$ &  11.25$^\circ$\\
\midrule

\texttt{3Dconv-F2F-RGB} & 57.0 & - & 40.8 & - & - & - & - & - & - & - & - & - \\

\texttt{Dil10-S2S} & 59.4 & 55.3 & 47.8 & 40.8 & - & - & - & - & - & - & - & - \\

\texttt{F2F} & - & 61.2 & - & 41.2 & 39.9 & 19.4 & 19.4 & 7.7 & - & - & - & - \\

\texttt{ConvLSTM} & 60.1 & - & - & - & - & - & - & - & - & - & - & -\\

\texttt{FeatReproj3D} & 61.5 & - & 45.4 & - & - & - & - & - & - & - & - & -\\

\texttt{Bayesian S2S} & 65.1 & - & 51.2 & - & - & - & - & - & - & - & - & -\\

\texttt{3Dconv-F2F-SEG} & 65.5 & - & 50.5 & - & - & - & - & - & - & - & - & -\\

\texttt{DeformF2F} & 65.5 & 63.8 & 53.6 & 49.9 & - & - & - & - & - & - & - & -\\

\texttt{LSTM AM S2S} & 65.8 & - & 51.3 & -  & - & - & - & - & - & - & - & -\\

\texttt{CPConvLSTM} & - & - & - & - & 44.3 & 22.1 & 25.6 & 11.2 & - & - & - & -\\

\texttt{APANet} & - & 64.9 & - & 51.4 & 46.1 & 23.2 & \underline{29.2} & \underline{12.9} & - & - & - & -\\

\texttt{LSTM M2M} & 67.1 & 65.1 & 51.5 & 46.3  & - & - & - & - & - & - & - & -\\

\texttt{IndRNN-Stack} & 67.6 & 60.8 & 58.1 & 52.1  & - & - & - & - & - & - & - & -\\

\texttt{DiffAttn-Fuse} & 67.9 & 61.2 & 58.1 & 51.7  & - & - & - & - & - & - & - & -\\

\texttt{F2MF} & 69.6 & 67.7 & 57.9 & 54.6  & - & - & - & - & - & - & - & -\\

\texttt{PFA} (semantic)  & 71.1 & 69.2 & \underline{60.3} & 56.7  & - & - & - & - & - & - & - & -\\

\texttt{PFA} (instance)  & - & - & - & -  & \underline{48.7} & \underline{24.9} & 

\textbf{30.5} & \textbf{14.8} & - & - & - & -\\

\texttt{Futurist} & \textbf{73.9} & \textbf{74.9} & \textbf{62.7} & \textbf{61.2} & - & - & - & - & \textbf{96.0} & \textbf{91.9} & - & - \\
\midrule
\graycell{\texttt{Oracle}} & \graycell{77.0} & \graycell{77.4} & \graycell{77.0} & \graycell{77.4} & \graycell{66.2} & \graycell{40.4} & \graycell{66.2} & \graycell{40.4} & \graycell{89.1} & \graycell{89.1} & \graycell{95.3} & \graycell{95.3} \\

\texttt{VISTA}$_{ft}$ & 64.9 & 62.1 & 53.9 & 51.0 & 33.1 & 17.7 & 19.8 & 9.0 & 86.4 & 82.8 & 93.0 & 90.0 \\ 
\ours (ours) 
& \underline{71.8} & \underline{71.7} & 59.8  & \underline{57.6}
& \textbf{50.5} & \textbf{26.6} & 27.3 & 12.6 & \underline{88.6} & \underline{85.4} 
& \textbf{94.4}& \textbf{91.3}\\

\bottomrule
\end{tabular}
\label{tab:comparison_with_prior_work}
\end{table}

%% file: tables/tab_found_mod_tasks.tex
\begin{table}[t]
\caption{\textbf{Comparison of VFM encoders across tasks}. 
For each encoder (DINOv2, EVA2-CLIP, SAM), we show performance on segmentation (ALL, MO), depth estimation ($\delta_{1}$ accuracy, AbsRel error), and surface normal prediction (m, percentage within 11.25°).} 
\scriptsize
\centering
\setlength{\tabcolsep}{1.5pt} 
\begin{tabular}{ll C{0.75cm}C{0.75cm}C{0.75cm}C{0.75cm} @{\hspace{1.5em}}C{0.75cm}C{0.75cm}C{0.75cm}C{0.75cm} C{0.75cm}C{0.75cm}C{0.75cm}C{0.75cm}}
\toprule
\multirow{3}{*}[-1.0ex]{\Th{Encoder}} & \multirow{3}{*}[-1.0ex]{\Th{Method}} & \multicolumn{4}{c}{\Th{Segmentation}} & \multicolumn{4}{c}{\Th{Depth}} & \multicolumn{4}{c}{\Th{Surface Normals}} \\
\cmidrule(lr){3-6} \cmidrule(lr){7-10} \cmidrule(lr){11-14}
& & \multicolumn{2}{c}{\Th{Short}} & \multicolumn{2}{c}{\Th{Mid}} & \multicolumn{2}{c}{\Th{Short}} & \multicolumn{2}{c}{\Th{Mid}} & \multicolumn{2}{c}{\Th{Short}} & \multicolumn{2}{c}{\Th{Mid}} \\
\cmidrule(r){3-4} \cmidrule(lr){5-6} \cmidrule(lr){7-8} \cmidrule(lr){9-10} \cmidrule(lr){11-12} \cmidrule(lr){13-14}
& & ALL$\uparrow$ & MO$\uparrow$ & ALL$\uparrow$ & MO$\uparrow$ & $\delta_{1}$ $\uparrow$ & AbsR$\downarrow$ & $\delta_{1}$$\uparrow$ & AbsR$\downarrow$ & m$\downarrow$ & 11.25$^\circ$$\uparrow$ & m$\downarrow$ & 11.25$^\circ$$\uparrow$ \\
\midrule
\mr{2}{\Th{DINOv2}} 
& Oracle & 77.0 & 77.4 & 77.0 & 77.4 & 89.1 & .108 & 89.1 & .108 & 3.24 & 95.3 & 3.24 & 95.3 \\
& Copy Last & 54.7 & 52.0 & 40.4 & 32.3 & 84.1 & .154 & 77.8 & .212 & 4.41 & 89.2 & 5.39 & 84.0 \\
\cite{oquab2024dinov}& Prediction & \textbf{71.8} & \textbf{71.7} & \textbf{59.8} & \textbf{57.6} & \textbf{88.6} & \textbf{.114} & \textbf{85.4} & \textbf{.136} & \textbf{3.39} & \textbf{94.4} & \textbf{4.00} & \textbf{91.3} \\
\midrule
\mr{2}{\Th{EVA2-CLIP}} 
& Oracle & 71.0 & 69.5 & 71.0 & 69.5 & 85.2 & .123 & 85.2 & .123 & 3.37 & 94.5 & 3.37 & 94.5 \\
& Copy Last & 51.9 & 47.7 & 38.5 & 29.5 & 81.2 & .161 & 75.6 & .216 & 4.52 & 88.5 & 5.44 & 83.6 \\
\cite{fang2024eva}& Prediction & 66.3 & 64.2 & 54.5 & 49.6 & 85.1 & .122 & 82.5 & .145 & 3.56 & 93.4 & 4.18 & 90.1 \\
\midrule
\mr{2}{\Th{SAM}} 
& Oracle & 69.8 & 63.9 & 69.8 & 63.9 & 84.8 & .143 & 84.8 & .143 & 3.01 & 96.0 & 3.01 & 96.0 \\
& Copy Last & 49.4 & 41.8 & 36.8 & 26.0 & 78.3 & .211 & 73.4 & .267 & 4.84 & 87.4 & 5.77 & 82.4 \\
\cite{kirillov2023segment}& Prediction & 65.3 & 59.3 & 52.5 & 43.9 & 81.3 & .178 & 77.6 & .209 & 3.80 & 92.8 & 4.49 & 89.2 \\
\midrule
\midrule
\mr{3}{\graycell{\Th{VAE}}}
& \graycell{Oracle} & \graycell{47.3}  & \graycell{34.7} & \graycell{47.4}  & \graycell{35.2} & \graycell{61.5}  & \graycell{.251}  & \graycell{61.5}  & \graycell{.251} & \graycell{5.3}  & \graycell{86.1} & \graycell{5.3}  & \graycell{86.1} \\
& \graycell{Copy Last} & \graycell{37.1}  & \graycell{25.6}  & \graycell{28.5}  & \graycell{16.5} & \graycell{60.7}  & \graycell{.252}  & \graycell{59.2}  & \graycell{.286} & \graycell{5.8}  & \graycell{83.2}  & \graycell{6.3}  & \graycell{80.1}  \\
\graycell{\cite{rombach2022high}}& \graycell{Prediction} & \graycell{33.4}  & \graycell{17.9}  & \graycell{24.7}  & \graycell{9.8} & \graycell{64.1}  & \graycell{.281}  & \graycell{61.4}  & \graycell{.394} & \graycell{6.5}  & \graycell{80.5}  & \graycell{8.0}  & \graycell{73.2} \\
\bottomrule
\end{tabular}
\label{tab:ablation_vfm_backbones}
\end{table}

%% file: tables/tab_cont_disc_representations.tex
\begin{table}[!h]
\footnotesize
\centering
\caption{\textbf{Continuous vs. Discrete VFM Representations}. Comparison of continuous DINOv2 features (our approach) against 4M's DINOv2 tokenizer with discrete codes. Unlike other tables and for fair comparison with the 4M tokenized variant, we use the DINOv2 ViT-B w/o Reg model and extract features from the last layer only. Results on semantic segmentation forecasting.}
\setlength{\tabcolsep}{2.5pt}
\begin{tabular}{ccccccccc} \toprule
 \multirow{3}{*}[-1.0ex]{\makecell{\Th{Method}}} & \multicolumn{4}{c}{\Th{Continuous}} & \multicolumn{4}{c}{\Th{Discrete (4M tokenized)}} \\ \cmidrule(lr){2-5} \cmidrule(lr){6-9}
 & \multicolumn{2}{c}{ \Th{Short} } & \multicolumn{2}{c}{ \Th{Mid} } & \multicolumn{2}{c}{ \Th{Short} } & \multicolumn{2}{c}{ \Th{Mid} } \\ 
\cmidrule(lr){2-3} \cmidrule(lr){4-5} \cmidrule(lr){6-7} \cmidrule(lr){8-9}
 & \Th{ALL}$\uparrow$ & \Th{MO}$\uparrow$ & \Th{ALL}$\uparrow$ & \Th{MO}$\uparrow$ & \Th{ALL}$\uparrow$ & \Th{MO}$\uparrow$ & \Th{ALL}$\uparrow$ & \Th{MO}$\uparrow$\\
\midrule
Oracle      & 72.9 & 74.0 & 72.9 & 74.0 & 70.2 & 71.4 & 70.2 & 71.4 \\
Copy Last   & 54.7 & 51.9 & 40.5 & 32.2 & 53.7 & 51.0 & 40.0 & 31.6 \\
Prediction  & 68.9 & 69.3 & 57.3 & 55.0 & 61.7 & 60.9 & 53.7 & 51.0 \\
\bottomrule
\end{tabular}
\label{tab:continuous_vs_discrete}
\end{table}

%% file: tables/tab_highres.tex
\begin{table}[h]
\caption{\textbf{Strategies for Training-Efficient High-Resolution Feature Forecasting}.}
\footnotesize
\centering
\setlength{\tabcolsep}{4pt}
\begin{tabular}{llcccc}
\toprule
\Th{Resolutions} & \Th{Adaptation} & \multicolumn{2}{c}{\Th{Short-term}} & \multicolumn{2}{c}{\Th{Mid-term}} \\
(Train$\rightarrow$Test) & \Th{Approach}  & \Th{ALL} & \Th{MO} & \Th{ALL} & \Th{MO} \\
\midrule
Oracle\\
(a) 224$\rightarrow$224 & N/A & 68.24 & 66.41 & 68.24 & 66.41 \\
(b) 448$\rightarrow$448 & N/A & 76.97 & 77.40 & 76.97 & 77.40 \\
\midrule
Forecasting\\
(c) 224$\rightarrow$224 & N/A & 64.50 & 62.63 & 55.49 & 52.62 \\
(d) 224$\rightarrow$448 & Pos. interp. & 64.34 & 64.29 & 48.31 & 44.60 \\
(e) 224$\rightarrow$448$_{224}$ & Sliding win. & 71.26 & 71.11 & 58.75 & 56.78 \\
(f) (224$\&$448)$\rightarrow$448 & Two-phase & \textbf{71.81} & \textbf{71.71} & \textbf{59.78} & \textbf{57.65} \\
\bottomrule
\end{tabular}%
\label{tab:highres-results}
\vspace{-10pt}
\end{table}

%% file: sec/fig_qualitative_results.tex
\scriptsize
\centering
\newcommand{\resultsfig}[1]{\includegraphics[width=0.113\textwidth,valign=c]{#1}}
\newcommand{\resultsfigtwo}[1]{\includegraphics[width=0.113\textwidth,valign=c]{#1}}
\setlength{\tabcolsep}{1pt}

\resizebox{\textwidth}{!}{
\begin{tabular}{@{}c@{\hspace{8pt}}c@{}}
\begin{tabular}{@{}ccccc@{}}
    & \mc{2}{\textbf{Scene Frankfurt (1)}}
    & \mc{2}{\textbf{Scene  Munster (15)}} \\
    \mc{5}{\vspace{-1.6ex}}\\
    & Last Context Frame 
    & Future Frame 
    & Last Context Frame 
    & Future Frame \\
    &
    \resultsfigtwo{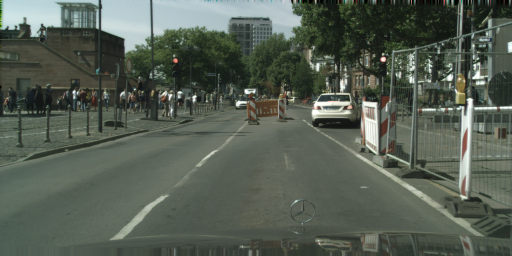}
    &
    \resultsfigtwo{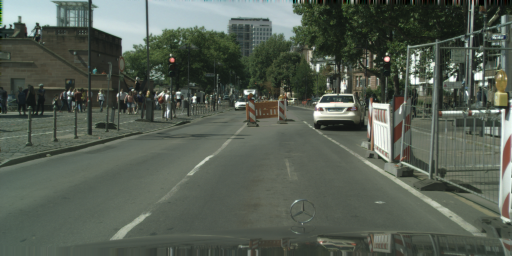}
     &
     \resultsfigtwo{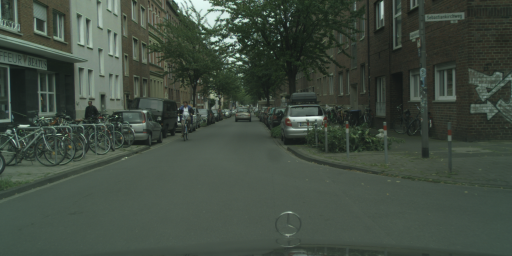}
     & 
    \resultsfigtwo{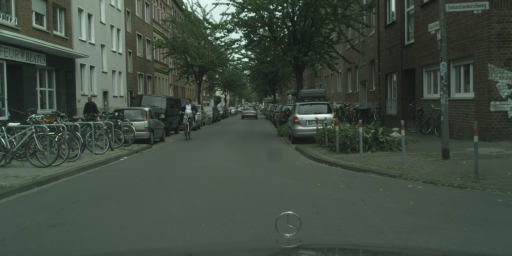}
     \\
    \mc{5}{\vspace{-1.6ex}}\\
    & Ours
    & Oracle
    & Ours
    & Oracle \\
    \raisebox{-.45\height}{\rotatebox{90}{\centering Segm.}}
    &
    \resultsfig{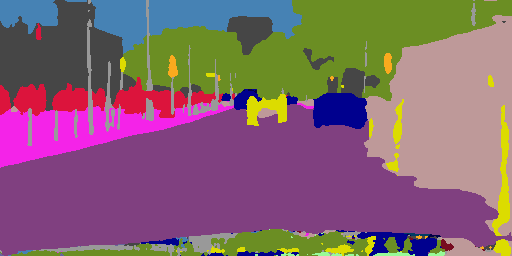} & 
    \resultsfig{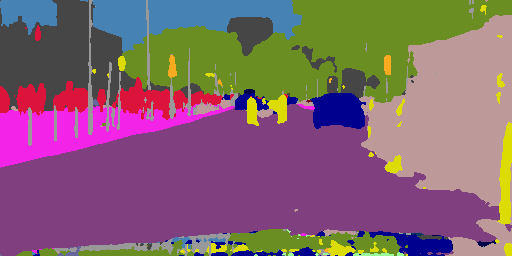} &
    \resultsfig{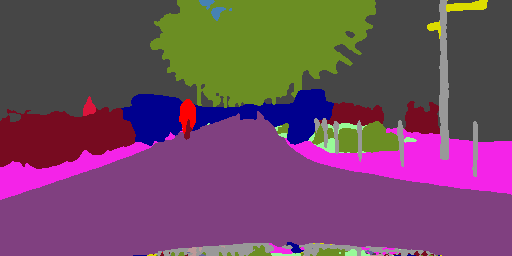} & 
    \resultsfig{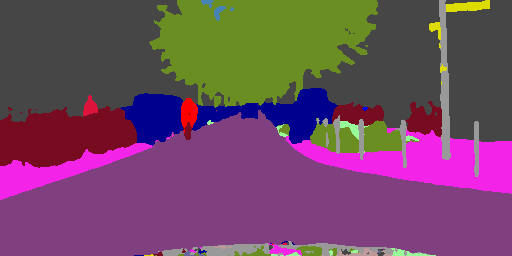}\\
    \mc{5}{\vspace{-1.1ex}}\\
    \raisebox{-.45\height}{\rotatebox{90}{\centering Depth}}
    &
    \resultsfig{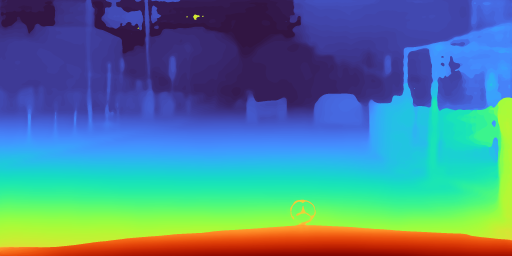} & 
    \resultsfig{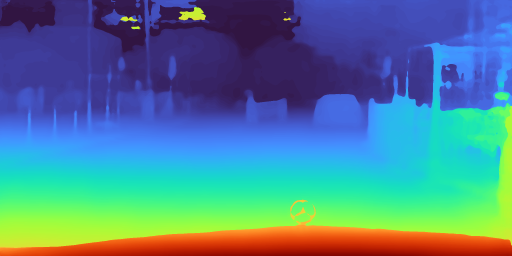} &
    \resultsfig{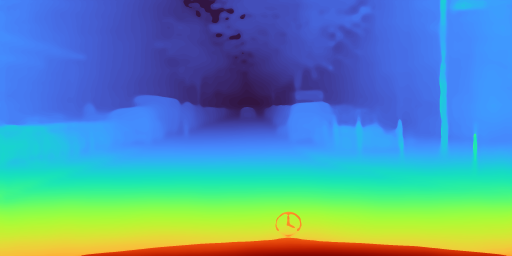} & 
    \resultsfig{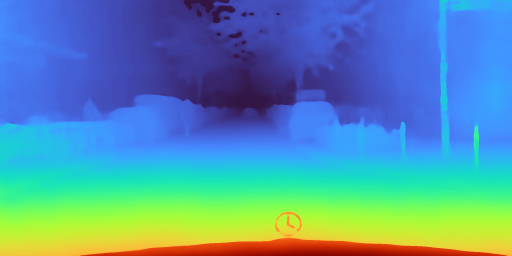}\\
    \mc{5}{\vspace{-1.1ex}}\\
    \raisebox{-.45\height}{\rotatebox{90}{\centering Normals}}
    &
    \resultsfig{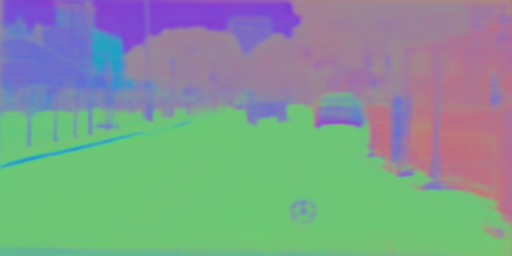} & 
    \resultsfig{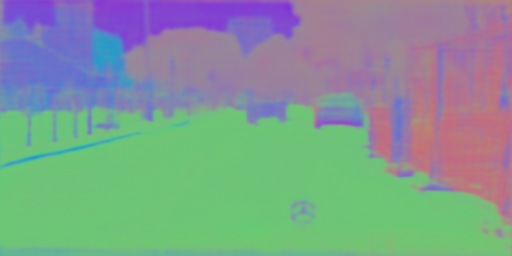} &
    \resultsfig{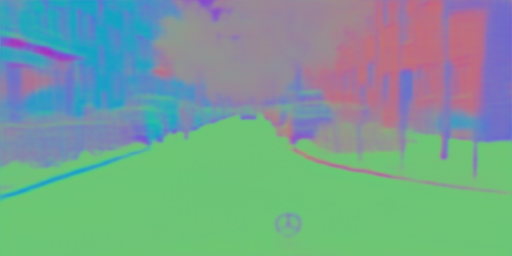} & 
    \resultsfig{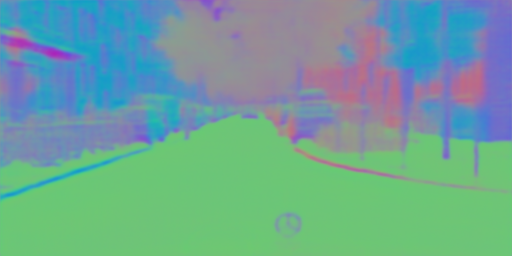}\\
    &
    \mc{4}{(a) Short-Term}\\
\end{tabular}
&
\begin{tabular}{@{}ccccc@{}}
    & \mc{2}{\textbf{Scene Frankfurt (1)}}
    & \mc{2}{\textbf{Scene  Munster (15)}} \\
    \mc{5}{\vspace{-1.6ex}}\\
    & Last Context Frame 
    & Future Frame
    & Last Context Frame 
    & Future Frame \\
    & 
    \resultsfigtwo{figs/Predictions_Dino/frankfurt_000001_017098_leftImg8bit}
    &
    \resultsfigtwo{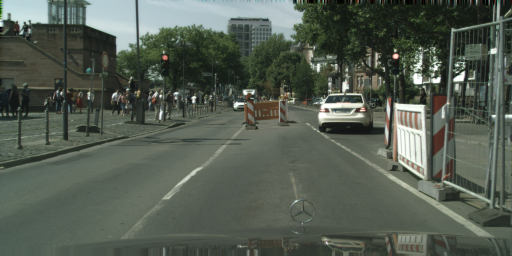}
    & 
    \resultsfigtwo{figs/Predictions_Dino/munster_000015_000016_leftImg8bit}
    & 
    \resultsfigtwo{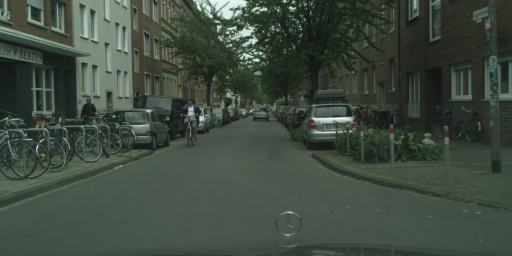}
     \\
    \mc{5}{\vspace{-1.6ex}}\\
    & Ours
    & Oracle
    & Ours
    & Oracle \\
    \raisebox{-.45\height}{\rotatebox{90}{\centering Segm.}}
    &
    \resultsfig{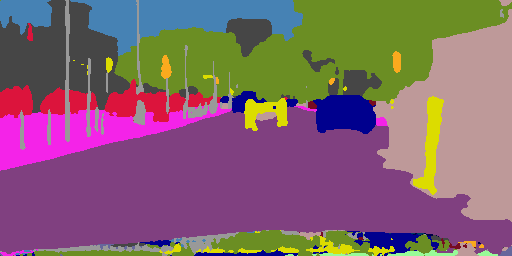} & 
    \resultsfig{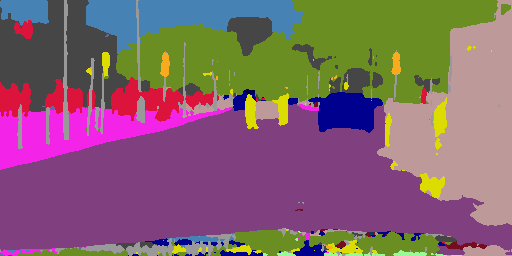} &
    \resultsfig{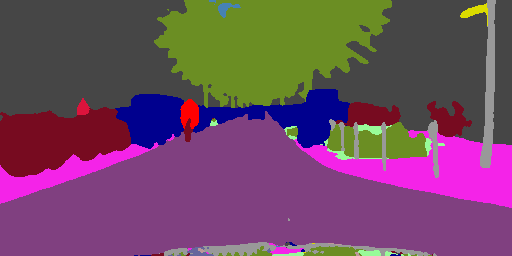} & 
    \resultsfig{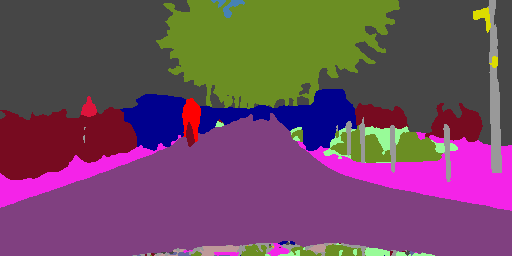}\\
    \mc{5}{\vspace{-1.1ex}}\\
    \raisebox{-.45\height}{\rotatebox{90}{\centering Depth}}
    &
    \resultsfig{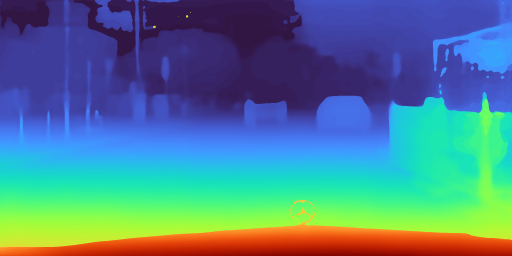} & 
    \resultsfig{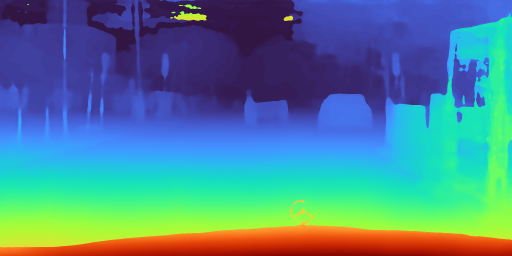} &
    \resultsfig{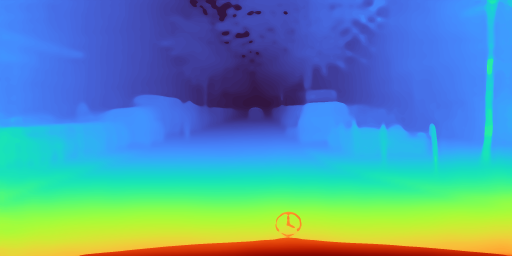} & 
    \resultsfig{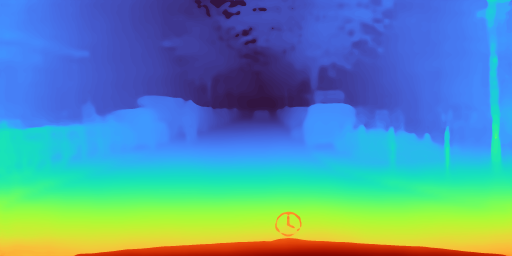}\\
    \mc{5}{\vspace{-1.1ex}}\\
    \raisebox{-.45\height}{\rotatebox{90}{\centering Normals}}
    &
    \resultsfig{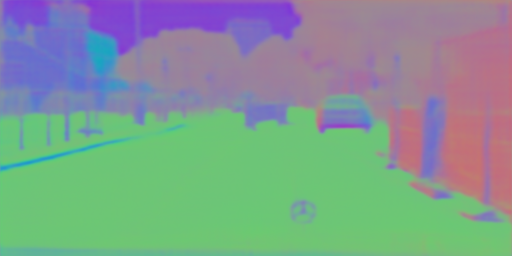} & 
    \resultsfig{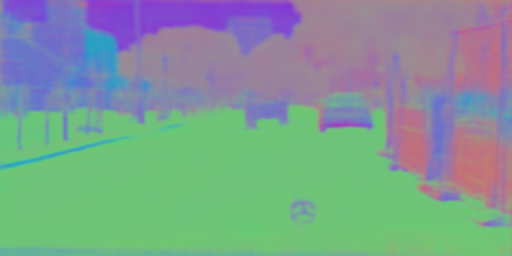} &
    \resultsfig{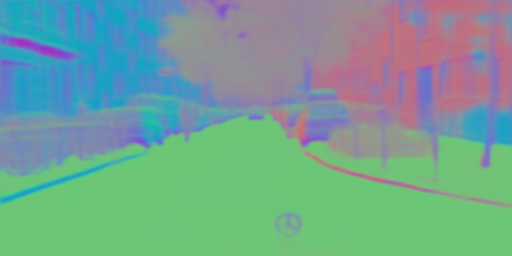} & 
    \resultsfig{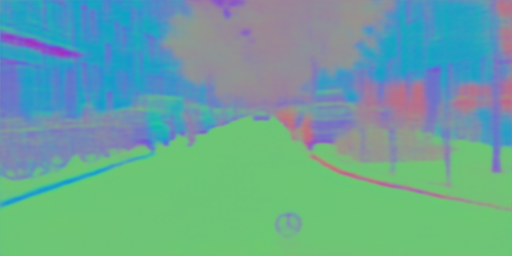}\\
    &
    \mc{4}{(b) Mid-Term}\\
\end{tabular}
\end{tabular}
}

%% file: sec/5_Conclusion.tex
\section{Conclusion}
\label{sec:conclusion}

In this work, we introduced \ours, a self-supervised framework for semantic future prediction that shifts the paradigm from forecasting low-level latent representations to predicting high-dimensional, semantically rich VFM features. This shift offers several key advantages: it enhances scene understanding by leveraging structured semantic information, avoids the complexity of full-frame synthesis, and enables scalable, modular integration with downstream tasks.

To realize this approach, we designed a masked feature transformer that efficiently propagates high-resolution, multi-layer VFM features over time. Our experiments show that forecasting such features is not only feasible but also highly effective—demonstrating strong performance across diverse future-frame understanding tasks including semantic segmentation, instance segmentation, depth estimation, and surface normal prediction. Unlike prior methods that rely on multiple task-specific models, our single model handles all tasks seamlessly, validating the scalability and versatility of our framework.

Overall, this work lays the foundation for a new class of unified and modular future prediction systems, grounded in semantic reasoning rather than low-level reconstruction.

%% file: sec/acknowledgements.tex
\paragraph{Acknowledgements}

This work has been partially supported by project MIS 5154714 of the National Recovery and Resilience Plan Greece 2.0 funded by the European Union under the NextGenerationEU Program. Hardware resources were granted with the support of GRNET. Also, this work was performed using HPC resources from GENCI-IDRIS (Grants 2023-A0141014182, 2023-AD011012884R2, and 2024-AD011012884R3).

%% file: sec/supplementary.tex
\newpage

{\huge \textbf{Appendix}}
\section{Additional Results}
\label{sec:additional_results}

\subsection{Impact of Dimensionality Reduction}
\label{sec:impact_pca}

In our work, we examine a PCA-based dimensionality reduction method and find that compressing features in this way does not compromise performance on semantic segmentation and depth prediction downstream tasks (\autoref{tab:impact_pca}). In fact, reducing the dimensionality simplifies the modeling process and leads to improved performance. Specifically, for semantic segmentation forecasting, PCA enhances short-term predictions—particularly for moving objects—while its effect on mid-term predictions is negligible. Similarly, for depth forecasting, dimensionality reduction consistently boosts performance across all metrics for both short- and mid-term predictions.
\input{supp_tables/supp_tab_pca}

\subsection{Emerging Visual Representations in the Future-Frame Masked Feature Transformer} \label{sec:intermediate_features}

Self-supervised representation learning has achieved remarkable progress, with numerous studies focusing on extracting robust visual features from unlabeled images and videos~\citep{bardes2024revisiting, tong2022videomae, omnimae, ryali2023hiera, wang2023videomaev2, simclr, dino, mae, byol,wei2022masked, spot_cvpr, AttMask, gidaris2024moca, venkataramanan2025franca, sirko2025dip}. Inspired by these advancements, we investigate the potential of our future-frame masked feature transformer as a self-supervised method for enhancing VFM visual features. Specifically, we train DPT heads for semantic segmentation and depth prediction, using not only the features predicted by the masked feature transformer but also additional features extracted from intermediate transformer layers. We examine features from the 6th, 9th, 10th, 11th, and 12th (last) layers of the transformer to assess whether these intermediate representations can further improve the strong VFM features predicted by our masked transformer.
\input{sec/Intermfeats_future}

Results, shown in~\autoref{fig:supp_intermediate_features}, indicate that incorporating intermediate transformer features from the masked transformer enhances segmentation performance, with one exception (6th-layer features for short-term segmentation). Notably, the best segmentation results are achieved using features from the 9th layer. Although the improvements are modest, this aligns with expectations given the strength of the predicted VFM features alone.
Similar results are observed for future depth prediction, where intermediate features also led to performance gains, with the best depth results achieved using 6th-layer features (see~\autoref{fig:supp_intermediate_features}).
While exploring self-supervised learning was not the primary aim of our work, we find these results intriguing, as they suggest that future prediction methods hold promise as self-supervised visual representation learners. We hope this work can spark further research into this direction.

\subsection{Additional Comparisons with \texttt{VISTA}}
\label{sec:AdditionalVista}
In \autoref{tab:ablation_vista_base_supp}, we present a comprehensive evaluation comparing our method against \texttt{VISTA}~\citep{gao2024vista} for semantic segmentation, instance segmentation, depth and surface normal estimation on the Cityscapes dataset. We extend this evaluation to the nuScenes~\citep{nuscenes} dataset in \autoref{tab:ablation_vista_nuscenes} for depth and surface normals. 
Notably, \ours demonstrates strong zero-shot generalization capabilities. When trained only on Cityscapes and directly evaluated on nuScenes without fine-tuning, performance degradation is minimal, while still significantly outperforming all baselines. The results show that \ours consistently surpasses the performance of \texttt{VISTA} on both datasets.
\input{supp_tables/supp_tab_mod_tasks}
\input{supp_tables/supp_tab_mod_tasks_nuscenes}

\subsection{Impact of Multi-Layer Features}
\label{sec:impact_multi_features}

Scene understanding models benefit from utilizing features from multiple layers of a frozen image encoder. To fully exploit the pretrained DINO features, we integrate representations from several layers into the DPT head. \autoref{tab:comparison_layer_feats} presents an ablation comparing multi-layer DINO features to using only the final layer (layer 12) for semantic segmentation on Cityscapes. The results demonstrate that aggregating features from layers 3, 6, 9, and 12 enhances performance, with ALL (all semantic classes) and MO (movable objects classes) scores rising from 72.1/73.4 to 77.0/77.4.

\input{supp_tables/supp_tab_layerfeats}

\subsection{Masking strategy and loss function ablations} 
\label{sec:Masking_LossF_ablations}

We evaluate the impact of masking strategies and loss functions on forecasting performance. As shown in \autoref{tab:masking_ablation}, full masking (masking all future features) consistently outperforms random masking across all metrics, for semantic segmentation and depth forecasting. This demonstrates that masking all future features forces the model to learn more robust temporal dynamics. Regarding loss functions (\autoref{tab:loss_ablation}), we find that our framework is robust to loss function choice, with L1, MSE, SmoothL1, and SmoothL1+Cosine achieving comparable performance.

\input{supp_tables/supp_tab_masking_strategies}
\input{supp_tables/supp_tab_loss_functions}

\subsection{Model Size and Data Scale Scalability}
\label{sec:Model_Data_Scalability}
We investigate how performance scales with model size and training data diversity. As shown in \autoref{tab:model_scaling}, we evaluate three model variants—Small (115M), Base (258M), and Large (460M) parameters—by modifying the hidden dimension and number of attention heads while keeping dataset size and training duration fixed. Results demonstrate consistent performance improvements with increased model capacity, particularly for mid-term predictions, indicating that larger models better capture complex temporal dynamics in VFM features. Regarding data scale (\autoref{tab:data_scaling}), we combine Cityscapes and nuScenes datasets with equal-probability sampling during training and evaluate on Cityscapes. The model trained on combined datasets achieves consistent improvements across all tasks compared to training on Cityscapes alone, with gains particularly pronounced for mid-term semantic segmentation. These findings demonstrate that our framework effectively scales with both increased model capacity and diverse training data, motivating further exploration with larger models and datasets to fully realize the potential of forecasting VFM features for multi-task scene understanding.

\input{supp_tables/supp_tab_model_scalability}
\input{supp_tables/supp_tab_data_scalability}


\subsection{More Visualizations}
In~\autoref{fig:qualitative-results-v1} and \ref{fig:qualitative-results-v2}, we present additional qualitative results illustrating the prediction of semantic segmentation, depth maps, and surface normals. Specifically, we compare our method, \ours, against the \texttt{Oracle}, which involves using future RGB frames as inputs for different prediction heads, as well as \texttt{VISTA} \cite{gao2024vista}. 

As illustrated in~\autoref{fig:qualitative-results-v1},
\ours maintains superior integrity of motion dynamics and geometric consistency across frames, resulting in more accurate predictions

In~\autoref{fig:qualitative-rollout-1} and \ref{fig:qualitative-rollout-2}, we offer additional qualitative results derived from utilizing \ours for the prediction of semantic segmentation and depth maps and surface normals over extended time intervals. These outcomes are achieved through the use of autoregressive rollouts. Beginning with a series of four context frames (\(X_{t-9}\) to \(X_t\)), the model is capable of predicting up to 48 subsequent frames, equivalent to 2.88 seconds, with predictions occurring at an interval of every third frame.  Our model consistently delivers accurate predictions over the entire forecasted duration, effectively capturing motion dynamics and maintaining consistency across different modalities. This performance underscores its robustness and versatility, which are related to its capability of predicting the features of a foundation model. As a final remark, it is important to note that the noisy segmentation predictions observed at the bottom of the images, present in both the predicted and \texttt{Oracle} results, are attributed to unannotated regions in the Cityscapes dataset that are disregarded during DPT training. This artifact impacts only the segmentation outcomes of DPT head and does not affect the predicted future features, as evidenced by the clear and accurate depth and surface normal predictions.

\section{Limitations and Future Work}
\label{sec:limitations}

In this work, we introduced \ours, a simple yet effective method for semantic future prediction based on forecasting VFM features. Our approach delivers strong results while opening several exciting directions for future research.

First, our method uses a straightforward masked transformer with SmoothL1 loss. While forecasting VFM features avoids the challenges of modeling complex pixel distributions, our current implementation is deterministic. However, our framework can easily be extended to capture uncertainty—for example, by adding a diffusion loss (as in MAR~\citep{li2024autoregressive}) or modeling tokens with a Gaussian mixture model (as in GIVT~\citep{tschannen2024givt}). These extensions would better handle future ambiguity while maintaining the simplicity of our approach.

Although we explored strategies to reduce training compute for high-resolution feature prediction, inference-time compute demands remain unchanged. Future work could address this by adopting hierarchical transformer architectures~\citep{ryali2023hiera}, which would not only improve efficiency but also enable the model to handle even higher feature resolutions.

Another promising direction is scaling \ours to larger datasets and models. Our experiments on model size (115M to 460M parameters) and data diversity (Cityscapes+nuScenes)  demonstrate consistent performance improvements, particularly for mid-term predictions, suggesting that further scaling to even larger models and more diverse datasets could yield substantial gains in forecasting performance.. Furthermore, the flexibility of our framework allows seamless integration of newer VFM encoders, such as RADIOv2.5~\citep{heinrich2025radiov25improvedbaselinesagglomerative}, which combines multiple VFMs into a single, more powerful model, enhancing its multi-task future scene understanding capabilities.

Overall, these research directions highlight the flexibility and growth potential of our approach, paving the way for further advancements in semantic future prediction.

\section{Broader Impact}
\label{sec:broader_impact}
Our work enables efficient and scalable semantic future prediction by forecasting semantically rich VFM features. This allows flexible integration with different scene understanding tasks without retraining, making it useful for applications like autonomous driving and robotics. While we do not foresee risks in our approach, we must remain mindful that the pretrained Vision Foundation Models we build upon may carry biases, which could potentially influence our semantic future predictions.

\section{Implementation Details}
\label{sec:appendix-implemdetails}

We provide implementation details for the heads trained on different downstream tasks. The DPT head is used for semantic segmentation, depth estimation, and surface normal estimation. We adopt the DPT \cite{ranftl2021vision} implementation from Depth Anything~\cite{yang_depth,yang2024depth}, setting the feature dimensionality to 256 and configuring $\texttt{dptoutchannels = [128, 256, 512, 512]}$. 
For all tasks, models are trained for 100 epochs with a batch size of 128 ($16 \times 8$ GPUs). The learning rate is set to 0.0016, using the AdamW optimizer with linear warmup for the first 10 epochs, and weight decay is 0.0001. For semantic segmentation, we use a polynomial scheduler and cross-entropy loss with 19 classes. For depth estimation, we use a cosine annealing scheduler and cross-entropy loss, with 256 classes. For surface normal estimation, we employ a polynomial scheduler and a loss function combining cosine similarity and $L_2$ loss with weighted averaging, using 3 classes.

For the Mask2Former head used in instance segmentation, we implement our approach using the official Mask2Former~\cite{cheng2022masked} and Detectron2~\cite{wu2019detectron2} codebases. The main difference compared to the official Mask2Former configuration for Cityscapes instance segmentation is the input feature maps. In our approach, the four multi-scale feature maps expected by Mask2Former are derived from the PCA features. These PCA features are first projected to 128, 256, 512, and 1024 dimensions and then resized so their spatial resolutions are $\times4$, $\times2$, $\times1$, and $\times0.5$ relative to the original resolution of the DINOv2 ViT-B outputs. We train using the AdamW optimizer, with a batch size of 64 ($8 \times 8$ GPUs), learning rate of 0.00032, weight decay of 0.05, and 67,500 iterations, with a polynomial scheduler.

Regarding Vista, we fine-tuned the model with 8 frames in total (3 as context and 5 future frames) for cityscapes, while for Nuscenes we used 9 frames in total (3 as context and 6 future frames) to support long-term forecasting.

\section{Definitions of Evaluation Metrics}
\label{sec:appendix-evalmetrics}
For \textbf{depth prediction}, we use two metrics: the mean Absolute Relative Error (AbsRel), defined as \(\frac{1}{M} \sum_{i=1}^M \frac{|a_i - b_i|}{b_i}\), where \( a_i \) and \( b_i \) are the predicted and ground truth disparities at pixel \( i \), and \( M \) is the number of pixels. We also evaluate depth accuracy using \( \delta_{1} \), the percentage of pixels where \(\max\left(\frac{a_i}{b_i}, \frac{b_i}{a_i}\right) < 1.25\). For \textbf{surface normal evaluation}, we compute the mean angular error m$\downarrow$ as \(\frac{1}{N} \sum_{i=1}^{N} \cos^{-1}\left( \frac{\mathbf{n}_i \cdot \tilde{\mathbf{n}}_i}{\|\mathbf{n}_i\| \, \|\tilde{\mathbf{n}}_i\|} \right)\), where \(\mathbf{n}_i\) and \(\tilde{\mathbf{n}}_i\) are the predicted and ground truth normals, respectively. Furthermore, we measure precision through the percentage of pixels with angular errors below 11.25°, calculated as \(\left( \frac{1}{N} \sum_{i=1}^{N} \mathbb{I}(\theta_i < 11.25^\circ) \right) \)

\section{Details of Evaluation Scenarios}
\label{sec:appendix-evalscenarios}
Regarding Cityscapes, the target frame for both short and mid term prediction is 20. We subsample sequences by a factor of 3 before inputting to the model. For short-term prediction, the model uses frames 8, 11, 14, and 17 as context to predict frame 20 (with context length \(N_c=4\) and \(N_p=1\)). For mid-term prediction, the model uses frames 2, 5, 8, and 11 as context and predicts frame 20 auto-regressively through frames 14 and 17. We calculate segmentation metrics on the 20th frame using Cityscapes ground truth. For depth and surface normals, we rely on pseudo-annotations from DepthAnythingV2~\cite{yang2024depth} and Lotus~\cite{he2025lotus}, respectively, due to the lack of true annotations in Cityscapes. Regarding nuScenes, the target frame for both mid and long term prediction is 29. Again, we subsample sequences by a factor of 3 before input to the model. For mid-term prediction, the model uses frames 11, 14, 17, and 20 as context and predicts frame 29 auto-regressively through frames 23 and 26. For long-term prediction, the model uses frames 2, 5, 8, and 11 as context and predicts frame 29 auto-regressively through frames 14,17,20,23 and 26.  Again for depth and surface normals, we rely on pseudo-annotations from DepthAnythingV2~\cite{yang2024depth} and Lotus~\cite{he2025lotus}, respectively, due to the lack of true annotations in nuScenes.

\newpage
\begin{figure*}[!h]
    \centering
    \include{sec/supp_fig_qualitative_vista_frankfurt}
\caption{\textbf{Visualization of future predictions for semantic segmentation, depth, and surface normals.} The illustrated scene is Frankfurt (01 (017082-017111)).} 
    \label{fig:qualitative-results-v1}
\end{figure*}


\begin{figure*}[!h]
    \centering
    \include{sec/supp_fig_qualitative_vista_munster}
\caption{\textbf{Visualization of future predictions for semantic segmentation, depth, and surface normals}. The illustrated scene is Munster (15).} 
    \label{fig:qualitative-results-v2}
\end{figure*}

\begin{figure*}[!h]
    \centering
    \include{sec/supp_fig_rollout}
    \vspace{-10pt}
    \caption{\textbf{Long-term semantic segmentation, depth and surface normal predictions.} The illustrated scene is Frankfurt (01 (011791-011820)). \ours consistently preserves motion dynamics and intricate details in complex scenes over extended time horizons.}
    \label{fig:qualitative-rollout-1}
\end{figure*}

\begin{figure*}[!h]
    \centering
    \include{sec/supp_fig_rollout_2}
    \vspace{-10pt}
    \caption{\textbf{Long-term semantic segmentation, depth and surface normal predictions}. The illustrated scene is Frankfurt (01 (006570-006599)).
    \ours excels in predicting the motion of the nearby car but faces challenges with distant, low-motion objects, highlighting areas for future improvement.}
    \label{fig:qualitative-rollout-2}
\end{figure*}

%% file: supp_tables/supp_tab_pca.tex
\begin{table}[!h]
\footnotesize
\centering
\caption{\textbf{Impact of Dimensionality Reduction}. Reduction is performed using PCA. Results on semantic segmentation and depth forecasting.}
\setlength{\tabcolsep}{2.5pt}
\begin{tabular}{ccccccccc} \toprule
 \multirow{3}{*}[-1.0ex]{\makecell{\Th{Dim.}\\\Th{Reduction}}} & \multicolumn{4}{c}{\Th{Segmentation}} & \multicolumn{4}{c}{\Th{Depth}} \\ \cmidrule(lr){2-5} \cmidrule(lr){6-9}
 & \multicolumn{2}{c}{ \Th{Short} } & \multicolumn{2}{c}{ \Th{Mid} } & \multicolumn{2}{c}{ \Th{Short} } & \multicolumn{2}{c}{ \Th{Mid} } \\ 
\cmidrule(lr){2-3} \cmidrule(lr){4-5} \cmidrule(lr){6-7} \cmidrule(lr){8-9}
 & \Th{ALL}$\uparrow$ & \Th{MO}$\uparrow$ & \Th{ALL}$\uparrow$ & \Th{MO}$\uparrow$ & $\delta_{1}$ $\uparrow$ & AbsR$\downarrow$ & $\delta_{1}$$\uparrow$ & AbsR$\downarrow$\\
\midrule
\xmark & 71.3 & 70.4 & \textbf{59.9} & \textbf{57.6} & 87.9 & .122 & 84.8 & .147\\
\cmark & \textbf{71.8} & \textbf{71.7} & 59.8 & \textbf{57.6} & \textbf{88.6} & \textbf{.114} & \textbf{85.4} & \textbf{.136}\\
\bottomrule
\end{tabular}%
\label{tab:impact_pca}
\end{table}

%% file: sec/Intermfeats_future.tex
\begin{figure*}[!h]
    \centering  
    \begin{tabular}{c@{\hspace{0.3cm}}c@{\hspace{0.5cm}}c@{\hspace{0.3cm}}c}
    \multicolumn{4}{c}{
        \centering
        \begin{tikzpicture}
            \centering
            \begin{customlegend}[legend columns=3, legend style={align=center, column sep=.7ex, nodes={scale=0.9, transform shape}, draw=white!90!black}, legend entries={Predicted + Intermediate Transformer Features, Predicted VFM features only}]
                \addlegendimage{blue, fill=blue!60, area legend}
                \addlegendimage{black, dashed, thick, line legend}
            \end{customlegend}
        \end{tikzpicture}
    }      
    \\
    \centering
    \begin{tikzpicture}
        \tikzstyle{every node}=[font=\scriptsize]
        \begin{axis}[
            width=0.24\linewidth,
            height=0.24\linewidth,
            xlabel={Transformer Layer},
            ylabel={mIoU},
            font=\footnotesize,
            ymin=71.4, ymax=72.2,
            xmin=0.5, xmax=5.5,
            ytick={71.5,72.0},
            xtick={1,2,3,4,5},
            xticklabels={6,9,10,11,12},
            xlabel shift=-5 pt,
            ylabel shift=-3 pt,                
            grid=both,
            legend cell align={left},
            axis y line*=left,
            axis x line*=bottom,
            bar width=8pt,
            ybar=2pt,
            scaled y ticks=false,
            yticklabel style={
                /pgf/number format/.cd,
                fixed,
                fixed zerofill,
                precision=1
            },
            tick label style={font=\footnotesize}, 
            label style={font=\footnotesize}, 
            title={(a) Short-term, Segm.},
            title style={font=\footnotesize},
            title style={yshift=-22.ex,},             
        ]
        \addplot[blue, fill=blue!60] coordinates {
            (1,71.52765656)
            (2,72.11769867)
            (3,71.94857025)
            (4,71.94340515)
            (5,71.94602203)
        };
        \draw[black, dashed, thick] (axis cs:0.5,71.84496307) -- (axis cs:5.5,71.84496307);
        \end{axis}
    \end{tikzpicture}
    & 
    \hspace{-1.3em} 
    \centering
    \begin{tikzpicture}
        \tikzstyle{every node}=[font=\scriptsize]
        \begin{axis}[
            width=0.24\linewidth,
            height=0.24\linewidth,
            xlabel={Transformer Layer},
            font=\footnotesize,
            ymin=59.4, ymax=60.30,
            xmin=0.5, xmax=5.5,
            ytick={59.5,60.0},
            xtick={1,2,3,4,5},
            xticklabels={6,9,10,11,12},
            xlabel shift=-5 pt,
            grid=both,
            legend cell align={left},
            axis y line*=left,
            axis x line*=bottom,
            bar width=8pt,
            ybar=2pt,
            scaled y ticks=false,
            yticklabel style={
                /pgf/number format/.cd,
                fixed,
                fixed zerofill,
                precision=1
            },
            tick label style={font=\footnotesize}, 
            label style={font=\footnotesize}, 
            title={(b) Mid-term, Segm},
            title style={font=\footnotesize},
            title style={yshift=-22.ex,},              
        ]
        \addplot[blue, fill=blue!60] coordinates {
            (1,59.99101257) 
            (2,60.12539291) 
            (3,59.95279694) 
            (4,59.99740982) 
            (5,59.83141708) 
        };
        \draw[black, dashed, thick] (axis cs:0.5,59.71506119) -- (axis cs:5.5,59.71506119);
        \end{axis}
    \end{tikzpicture}
    \hfill
    &
    \begin{tikzpicture}
        \tikzstyle{every node}=[font=\scriptsize]
        \begin{axis}[
            width=0.24\linewidth,
            height=0.24\linewidth,
            xlabel={Transformer Layer},
            ylabel={AbsRel Reduction},
            ymin=-0.0003, ymax=0.0050,
            xmin=0.5, xmax=5.5,
            font=\footnotesize,
            ytick={0.000,0.002,0.004},
            yticklabels={.000,.002,.004},
            xtick={1,2,3,4,5},
            xticklabels={6,9,10,11,12},
            xlabel shift=-5 pt,
            ylabel shift=-3 pt, 
            grid=both,
            legend cell align={left},
            axis y line*=left,
            axis x line*=bottom,
            bar width=8pt,
            ybar=2pt,
            scaled y ticks=false,
            yticklabel style={
                /pgf/number format/.cd,
                fixed,
                fixed zerofill,
                precision=3
            },
            tick label style={font=\footnotesize}, 
            label style={font=\footnotesize}, 
            title={(c) Short-term, Depth},
            title style={font=\footnotesize},
            title style={yshift=-22.ex,},             
        ]
        \addplot[blue, fill=blue!60] coordinates {
            (1,0.00475172) 
            (2,0.00125166) 
            (3,0.00358317) 
            (4,0.00231268) 
            (5,0.00229064) 
        };        
        \draw[black, dashed, thick] (axis cs:0.5,0.) -- (axis cs:5.5,0.);
        \end{axis}
    \end{tikzpicture}    
    &
    \hspace{-1.3em} 
    \begin{tikzpicture}
        \tikzstyle{every node}=[font=\scriptsize]
        \begin{axis}[
            width=0.24\linewidth,
            height=0.24\linewidth,
            xlabel={Transformer Layer},
            ymin=-0.0003, ymax=0.0050,
            xmin=0.5, xmax=5.5,
            font=\footnotesize,
            ytick={0.000,0.002,0.004},
            yticklabels={.000,.002,.004},
            xtick={1,2,3,4,5},
            xticklabels={6,9,10,11,12},
            xlabel shift=-5 pt,
            ylabel shift=-3 pt,           
            grid=both,
            legend cell align={left},
            axis y line*=left,
            axis x line*=bottom,
            bar width=8pt,
            ybar=2pt,
            scaled y ticks=false,
            yticklabel style={
                /pgf/number format/.cd,
                fixed,
                fixed zerofill,
                precision=3
            },
            tick label style={font=\footnotesize}, 
            label style={font=\footnotesize}, 
            title={(d) Mid-term, Depth},
            title style={font=\footnotesize},
            title style={yshift=-22.ex,},              
        ]
        \addplot[blue, fill=blue!60] coordinates {
            (1,0.0033007) 
            (2,0.0017571) 
            (3,0.00201985) 
            (4,0.00098474) 
            (5,0.00105399) 
        };
        \draw[black, dashed, thick] (axis cs:0.5,0.0) -- (axis cs:5.5,0.0);
        \end{axis}
    \end{tikzpicture}    
    \end{tabular}
    \vspace{-10pt}
    \caption{\textbf{Impact of Intermediate Transformer Features on Future Segmentation and Depth Prediction.} Results are shown for semantic segmentation and depth prediction heads using two feature sets: only the VFM features predicted by the masked feature transformer (dashed line) and combined features from both predicted and intermediate transformer layers (blue bars). We evaluate DPT heads trained on features from the 6th, 9th, 10th, 11th, and 12th layers. For segmentation (barplots (a) and (b)), we report mIoU across all classes. For depth (barplots (c) and (d)), we show the reduction in AbsRel metric (higher is better) when adding intermediate layer features.}
    \label{fig:supp_intermediate_features}
    \vspace{-4pt}
\end{figure*}

%% file: supp_tables/supp_tab_mod_tasks.tex
\begin{table*}[!ht]
\footnotesize
\centering
\caption{\textbf{Comparison across tasks on Cityscapes}. 
We use DINOv2 encoder and show performance on segmentation (ALL, MO), instance segmentation (AP50, AP), depth estimation ($\delta_{1}$ accuracy, AbsRel error), and surface normal prediction (m, percentage within 11.25°). \texttt{VISTA$_{ft}$} is the \texttt{VISTA} model fine-tuned on Cityscapes.} 
\resizebox{\columnwidth}{!}{
\begin{tabular}{l C{0.75cm}C{0.75cm}C{0.75cm}C{0.75cm} @{\hspace{1.5em}}C{0.75cm}C{0.75cm}C{0.75cm}C{0.75cm} @{\hspace{1.5em}}C{0.75cm}C{0.75cm}C{0.75cm}C{0.75cm} C{0.75cm}C{0.75cm}C{0.75cm}C{0.75cm}}
\toprule
 \multirow{3}{*}[-1.0ex]{\Th{Method}} & \multicolumn{4}{c}{\Th{Semantic Segm.}} & \multicolumn{4}{c}{\Th{Instance Segm}} & \multicolumn{4}{c}{\Th{Depth}} & \multicolumn{4}{c}{\Th{Surface Normals}} \\
\cmidrule(lr){2-5} \cmidrule(lr){6-9} \cmidrule(lr){10-13} \cmidrule(lr){14-17}
& \multicolumn{2}{c}{\Th{Short}} & \multicolumn{2}{c}{\Th{Mid}} & \multicolumn{2}{c}{\Th{Short}} & \multicolumn{2}{c}{\Th{Mid}} & \multicolumn{2}{c}{\Th{Short}} & \multicolumn{2}{c}{\Th{Mid}} & \multicolumn{2}{c}{\Th{Short}} & \multicolumn{2}{c}{\Th{Mid}} \\
\cmidrule(r){2-3} \cmidrule(lr){4-5} \cmidrule(lr){6-7} \cmidrule(lr){8-9} \cmidrule(lr){10-11} \cmidrule(lr){12-13} \cmidrule(lr){14-15} \cmidrule(lr){16-17}
& ALL$\uparrow$ & MO$\uparrow$ & ALL$\uparrow$ & MO$\uparrow$ & AP50$\uparrow$ & AP$\uparrow$ & AP50$\uparrow$ & AP$\uparrow$ & $\delta_{1}$ $\uparrow$ & AbsR$\downarrow$ & $\delta_{1}$$\uparrow$ & AbsR$\downarrow$ & m$\downarrow$ & 11.25$^\circ$$\uparrow$ & m$\downarrow$ & 11.25$^\circ$$\uparrow$ \\
\midrule
\graycell{\texttt{Oracle}} & \graycell{77.0} & \graycell{77.4} & \graycell{77.0} & \graycell{77.4} & \graycell{66.2} & \graycell{40.4} & \graycell{66.2} & \graycell{40.4} & \graycell{89.1} & \graycell{.108} & \graycell{89.1} & \graycell{.108} & \graycell{3.24} & \graycell{95.3} & \graycell{3.24} & \graycell{95.3} \\
\texttt{Copy Last} & 54.7 & 52.0 & 40.4 & 32.3 & 24.7 & 10.4 & 9.5 & 2.8 & 84.1 & .154 & 77.8 & .212 & 4.41 & 89.2 & 5.39 & 84.0 \\
\Th{\texttt{VISTA$_{ft}$}}& 64.9 & 62.1 & 53.9 & 51.0 & 33.1 & 17.7 & 19.8 & 9.0 & 86.4 & .124 & 82.8 & .153 & 3.75 & 93.0 & 4.30 & 90.0 \\
\ours & \textbf{71.8} & \textbf{71.7} & \textbf{59.8} & \textbf{57.6} & \textbf{50.5} & \textbf{26.6} & \textbf{27.3} & \textbf{12.6} & \textbf{88.6} & \textbf{.114} & \textbf{85.4} & \textbf{.136} & \textbf{3.39} & \textbf{94.4} & \textbf{4.00} & \textbf{91.3} \\
\bottomrule
\end{tabular}
}
\label{tab:ablation_vista_base_supp}
\end{table*}

%% file: supp_tables/supp_tab_mod_tasks_nuscenes.tex
\begin{table}[!ht]
\footnotesize
\centering
\caption{\textbf{Comparison across tasks at nuScenes}. 
We use DINOv2 encoder and show performance on depth estimation ($\delta_{1}$ accuracy, AbsRel error) and surface normal prediction (m, percentage within 11.25°). \texttt{VISTA$_{ft}$} is the \texttt{VISTA} model fine-tuned on nuScenes. The last row evaluates zero-shot generalization: \ours trained on Cityscapes and directly evaluated on nuScenes.} 
\setlength{\tabcolsep}{1.pt} 
\begin{tabular}{lcccccccc}
\toprule
 \multirow{3}{*}[-1.0ex]{\Th{Method}} & \multicolumn{4}{c}{\Th{Depth}} & \multicolumn{4}{c}{\Th{Surface Normals}} \\
\cmidrule(lr){2-5} \cmidrule(lr){6-9}

& \multicolumn{2}{c}{\Th{Mid}} & \multicolumn{2}{c}{\Th{Long}} & \multicolumn{2}{c}{\Th{Mid}} & \multicolumn{2}{c}{\Th{Long}} \\
\cmidrule(r){2-3} \cmidrule(lr){4-5} \cmidrule(lr){6-7} \cmidrule(lr){8-9}

& $\delta_{1}\uparrow$ & AbsR$\downarrow$ & $\delta_{1}\uparrow$ & AbsR$\downarrow$ & m$\downarrow$ & 11.25$^\circ$$\uparrow$ & m$\downarrow$ & 11.25$^\circ$$\uparrow$ \\
\midrule
\graycell{\texttt{Oracle}} &  \graycell{82.6} & \graycell{.206} & \graycell{82.6} & \graycell{.206} & \graycell{3.09} & \graycell{97.1} & \graycell{3.09} & \graycell{97.1} \\
\texttt{{Copy Last}} & 73.4 & .353 & {68.4} & {.468} & 4.66 & 88.6 & {5.31} & {85.3} \\
\texttt{\Th{VISTA$_{ft}$}} & 74.6 & .337 & 70.8 & .421 & 4.46 & 90.8 & 4.96 & 88.2 \\
{\ours}  & \textbf{80.7} & \textbf{.218} & {\textbf{76.3}} & {\textbf{.299}} & \textbf{3.59} & \textbf{93.9} & {\textbf{4.22}} & {\textbf{90.6}}\\
\midrule
{\ours (zero-shot)} &  \underline{78.4} & \underline{.269} & {\underline{72.1}} & {\underline{.377}} & \underline{4.03} & \underline{92.3} & {\underline{4.77}} & {\underline{88.4}} \\
\bottomrule
\end{tabular}
\label{tab:ablation_vista_nuscenes}
\end{table}

%% file: supp_tables/supp_tab_layerfeats.tex
\begin{table}[h]
\footnotesize
\centering
\caption{\textbf{DINO+DPT Features Ablation}. Ablation of multi-layer DINO features as input to the DPT head versus using only the last layer features, evaluated on semantic segmentation on Cityscapes.}
\setlength{\tabcolsep}{4.5pt}
\begin{tabular}{lcc} \toprule
\mr{2}{\Th{Layers}} & \multicolumn{2}{c}{ \Th{Segmentation} } \\ 
\cmidrule(lr){2-3}
 & \Th{ALL} & \Th{MO}\\
\midrule
12  & 72.1 & 73.4 \\
3,6,9,12  & 77.0 & 77.4 \\ 
\bottomrule
\end{tabular}%

\label{tab:comparison_layer_feats}
\end{table}

%% file: supp_tables/supp_tab_masking_strategies.tex
\begin{table*}[!ht]
\footnotesize
\centering
\caption{\textbf{Impact of Masking Strategies on Cityscapes}. 
We compare random masking versus full masking for semantic segmentation, depth estimation, and surface normal prediction. Full masking demonstrates consistently superior performance across all metrics.} 
\resizebox{\columnwidth}{!}{
\begin{tabular}{l C{0.75cm}C{0.75cm}C{0.75cm}C{0.75cm} @{\hspace{1.5em}}C{0.75cm}C{0.75cm}C{0.75cm}C{0.75cm} C{0.75cm}C{0.75cm}C{0.75cm}C{0.75cm}}
\toprule
 \multirow{3}{*}[-1.0ex]{\Th{Masking Strategy}} & \multicolumn{4}{c}{\Th{Semantic Segmentation}} & \multicolumn{4}{c}{\Th{Depth}} & \multicolumn{4}{c}{\Th{Surface Normals}} \\
\cmidrule(lr){2-5} \cmidrule(lr){6-9} \cmidrule(lr){10-13}
& \multicolumn{2}{c}{\Th{Short}} & \multicolumn{2}{c}{\Th{Mid}} & \multicolumn{2}{c}{\Th{Short}} & \multicolumn{2}{c}{\Th{Mid}} & \multicolumn{2}{c}{\Th{Short}} & \multicolumn{2}{c}{\Th{Mid}} \\
\cmidrule(r){2-3} \cmidrule(lr){4-5} \cmidrule(lr){6-7} \cmidrule(lr){8-9} \cmidrule(lr){10-11} \cmidrule(lr){12-13}
& ALL$\uparrow$ & MO$\uparrow$ & ALL$\uparrow$ & MO$\uparrow$ & $\delta_{1}$ $\uparrow$ & AbsR$\downarrow$ & $\delta_{1}$$\uparrow$ & AbsR$\downarrow$ & m$\downarrow$ & 11.25$^\circ$$\uparrow$ & m$\downarrow$ & 11.25$^\circ$$\uparrow$ \\
\midrule
Random Masking & 70.5 & 70.2 & 58.0 & 55.7 & 88.2 & .121 & 84.7 & .148 & 3.48 & 94.0 & 4.11 & 90.8 \\
Full Masking & \textbf{71.8} & \textbf{71.7} & \textbf{59.8} & \textbf{57.6} & \textbf{88.6} & \textbf{.114} & \textbf{85.4} & \textbf{.136} & \textbf{3.39} & \textbf{94.4} & \textbf{4.00} & \textbf{91.3} \\
\bottomrule
\end{tabular}
}
\label{tab:masking_ablation}
\end{table*}

%% file: supp_tables/supp_tab_loss_functions.tex
\begin{table*}[!ht]
\footnotesize
\centering
\caption{\textbf{Loss Function Comparison on Cityscapes}. 
We evaluate different loss functions (L1, MSE, SmoothL1, SmoothL1+Cosine) for semantic segmentation, depth estimation, and surface normal prediction. Results demonstrate that our framework is robust to loss function choice, with all variants achieving comparable performance.} 
\resizebox{\columnwidth}{!}{
\begin{tabular}{l C{0.75cm}C{0.75cm}C{0.75cm}C{0.75cm} @{\hspace{1.5em}}C{0.75cm}C{0.75cm}C{0.75cm}C{0.75cm} C{0.75cm}C{0.75cm}C{0.75cm}C{0.75cm}}
\toprule
 \multirow{3}{*}[-1.0ex]{\Th{Loss Function}} & \multicolumn{4}{c}{\Th{Semantic Segmentation}} & \multicolumn{4}{c}{\Th{Depth}} & \multicolumn{4}{c}{\Th{Surface Normals}} \\
\cmidrule(lr){2-5} \cmidrule(lr){6-9} \cmidrule(lr){10-13}
& \multicolumn{2}{c}{\Th{Short}} & \multicolumn{2}{c}{\Th{Mid}} & \multicolumn{2}{c}{\Th{Short}} & \multicolumn{2}{c}{\Th{Mid}} & \multicolumn{2}{c}{\Th{Short}} & \multicolumn{2}{c}{\Th{Mid}} \\
\cmidrule(r){2-3} \cmidrule(lr){4-5} \cmidrule(lr){6-7} \cmidrule(lr){8-9} \cmidrule(lr){10-11} \cmidrule(lr){12-13}
& ALL$\uparrow$ & MO$\uparrow$ & ALL$\uparrow$ & MO$\uparrow$ & $\delta_{1}$ $\uparrow$ & AbsR$\downarrow$ & $\delta_{1}$$\uparrow$ & AbsR$\downarrow$ & m$\downarrow$ & 11.25$^\circ$$\uparrow$ & m$\downarrow$ & 11.25$^\circ$$\uparrow$ \\
\midrule
L1 & 71.7 & 71.7 & 59.7 & 57.6 & 88.6 & .118 & \textbf{85.6} & .138 & 3.41 & 94.3 & 4.00 & 91.3 \\
MSE & 71.7 & \textbf{71.9} & \textbf{60.0} & \textbf{57.8} & 88.6 & .117 & 85.4 & .138 & 3.40 & \textbf{94.4} & \textbf{3.99} & \textbf{91.4} \\
SmoothL1+Cos & 71.7 & 71.4 & 59.8 & 57.5 & \textbf{88.7} & .116 & 85.5 & .137 & 3.40 & \textbf{94.4} & \textbf{3.98} & \textbf{91.4} \\
SmoothL1 & \textbf{71.8} & 71.7 & 59.8 & 57.6 & 88.6 & \textbf{.114} & 85.4 & \textbf{.136} & \textbf{3.39} & \textbf{94.4} & 4.00 & 91.3 \\
\bottomrule
\end{tabular}
}
\label{tab:loss_ablation}
\end{table*}

%% file: supp_tables/supp_tab_model_scalability.tex
\begin{table*}[!ht]
\footnotesize
\centering
\caption{\textbf{Model Size Scalability on Cityscapes}. 
We evaluate three model sizes—Small (115M), Base (258M), and Large (460M) parameters—across semantic segmentation, depth estimation, and surface normal prediction. Results demonstrate consistent performance improvements with increased model capacity.} 
\resizebox{\columnwidth}{!}{
\begin{tabular}{l c c C{0.75cm}C{0.75cm}C{0.75cm}C{0.75cm} @{\hspace{1.5em}}C{0.75cm}C{0.75cm}C{0.75cm}C{0.75cm} C{0.75cm}C{0.75cm}C{0.75cm}C{0.75cm}}
\toprule
 \multirow{3}{*}[-1.0ex]{\Th{Model Variant}} & \multirow{3}{*}[-1.0ex]{\Th{Hidden Dim}} & \multirow{3}{*}[-1.0ex]{\Th{Att Heads}} & \multicolumn{4}{c}{\Th{Semantic Segm.}} & \multicolumn{4}{c}{\Th{Depth}} & \multicolumn{4}{c}{\Th{Surface Normals}} \\
\cmidrule(lr){4-7} \cmidrule(lr){8-11} \cmidrule(lr){12-15}
& & & \multicolumn{2}{c}{\Th{Short}} & \multicolumn{2}{c}{\Th{Mid}} & \multicolumn{2}{c}{\Th{Short}} & \multicolumn{2}{c}{\Th{Mid}} & \multicolumn{2}{c}{\Th{Short}} & \multicolumn{2}{c}{\Th{Mid}} \\
\cmidrule(r){4-5} \cmidrule(lr){6-7} \cmidrule(lr){8-9} \cmidrule(lr){10-11} \cmidrule(lr){12-13} \cmidrule(lr){14-15}
& & & ALL$\uparrow$ & MO$\uparrow$ & ALL$\uparrow$ & MO$\uparrow$ & $\delta_{1}$ $\uparrow$ & AbsR$\downarrow$ & $\delta_{1}$$\uparrow$ & AbsR$\downarrow$ & m$\downarrow$ & 11.25$^\circ$$\uparrow$ & m$\downarrow$ & 11.25$^\circ$$\uparrow$ \\
\midrule
Small (115M) & 768 & 6 & 71.1 & 70.8 & 59.3 & 57.3 & 87.7 & .125 & 84.8 & .142 & 3.58 & 93.7 & 4.13 & 90.8 \\
Base (258M) & 1152 & 8 & 71.8 & \textbf{71.7} & 59.8 & 57.6 & \textbf{88.6} & \textbf{.114} & 85.4 & \textbf{.136} & \textbf{3.39} & \textbf{94.4} & \textbf{4.00} & 91.3 \\
Large (460M) & 1536 & 12 & \textbf{71.9} & \textbf{71.7} & \textbf{60.2} & \textbf{58.3} & \textbf{88.6} & .116 & \textbf{85.5} & .137 & 3.40 & \textbf{94.4} & \textbf{4.00} & \textbf{91.5} \\
\bottomrule
\end{tabular}
}
\label{tab:model_scaling}
\end{table*}

%% file: supp_tables/supp_tab_data_scalability.tex
\begin{table*}[!ht]
\footnotesize
\centering
\caption{\textbf{Data Scale Scalability}. 
We compare training on Cityscapes alone versus training on combined Cityscapes+nuScenes data. Results demonstrate that increasing training data diversity leads to improved performance across all tasks and temporal horizons.} 
\resizebox{\columnwidth}{!}{
\begin{tabular}{l C{0.75cm}C{0.75cm}C{0.75cm}C{0.75cm} @{\hspace{1.5em}}C{0.75cm}C{0.75cm}C{0.75cm}C{0.75cm} C{0.75cm}C{0.75cm}C{0.75cm}C{0.75cm}}
\toprule
 \multirow{3}{*}[-1.0ex]{\Th{Training Data}} & \multicolumn{4}{c}{\Th{Semantic Segmentation}} & \multicolumn{4}{c}{\Th{Depth}} & \multicolumn{4}{c}{\Th{Surface Normals}} \\
\cmidrule(lr){2-5} \cmidrule(lr){6-9} \cmidrule(lr){10-13}
& \multicolumn{2}{c}{\Th{Short}} & \multicolumn{2}{c}{\Th{Mid}} & \multicolumn{2}{c}{\Th{Short}} & \multicolumn{2}{c}{\Th{Mid}} & \multicolumn{2}{c}{\Th{Short}} & \multicolumn{2}{c}{\Th{Mid}} \\
\cmidrule(r){2-3} \cmidrule(lr){4-5} \cmidrule(lr){6-7} \cmidrule(lr){8-9} \cmidrule(lr){10-11} \cmidrule(lr){12-13}
& ALL$\uparrow$ & MO$\uparrow$ & ALL$\uparrow$ & MO$\uparrow$ & $\delta_{1}$ $\uparrow$ & AbsR$\downarrow$ & $\delta_{1}$$\uparrow$ & AbsR$\downarrow$ & m$\downarrow$ & 11.25$^\circ$$\uparrow$ & m$\downarrow$ & 11.25$^\circ$$\uparrow$ \\
\midrule
Cityscapes & 71.8 & 71.7 & 59.8 & 57.6 & \textbf{88.6} & \textbf{.114} & 85.4 & \textbf{.136} & \textbf{3.39} & \textbf{94.4} & 4.00 & 91.3 \\
Cityscapes+nuScenes & \textbf{72.3} & \textbf{72.2} & \textbf{61.0} & \textbf{59.4} & \textbf{88.6} & .117 & \textbf{85.7} & \textbf{.136} & 3.41 & \textbf{94.4} & \textbf{3.95} & \textbf{91.6} \\
\bottomrule
\end{tabular}
}
\label{tab:data_scaling}
\end{table*}

%% file: sec/supp_fig_qualitative_vista_frankfurt.tex
{
\small
\centering
\newcommand{\resultsfig}[1]{\includegraphics[width=0.3\textwidth,valign=c]{#1}}
\newcommand{\resultsfigtwo}[1]{\includegraphics[width=0.3\textwidth,valign=c]{#1}}

\setlength{\tabcolsep}{1pt}

\begin{tabular}{@{}cccc@{}}

    \mc{4}{\vspace{-1.6ex}}\\
    & Last Context Frame 
    & Future Frame 
    & VISTA (Output)

    \\

    &
    \resultsfigtwo{figs/Predictions_Dino/frankfurt_000001_017098_leftImg8bit}
    &
    \resultsfigtwo{figs/Predictions_Dino/frankfurt_000001_017101_leftImg8bit}
     &
     \resultsfigtwo{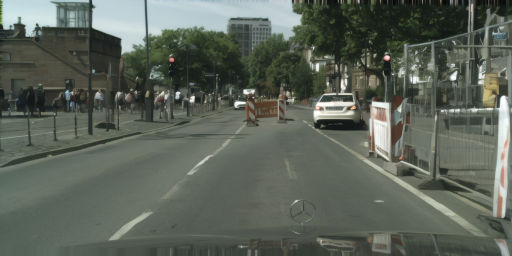}

     \\

    \mc{4}{\vspace{-1.6ex}}\\

    & Ours
    & Oracle
    & VISTA
    \\
    
    \raisebox{-.45\height}{\rotatebox{90}{\centering Segm.}}
    &
    \resultsfig{figs/Predictions_Dino/frankfurt_000001_017101_pred_segm_tplus3} & 
    \resultsfig{figs/Predictions_Dino/frankfurt_000001_017101_oracle_segm_tplus3} &
    \resultsfig{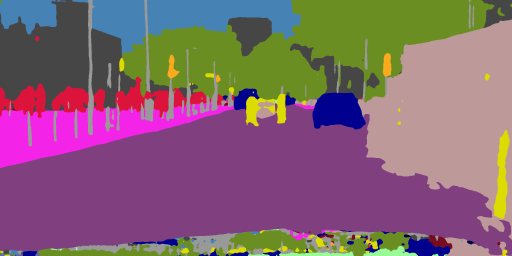} \\

    \mc{4}{\vspace{-2.1ex}}\\

    \raisebox{-.45\height}{\rotatebox{90}{\centering Depth}}
    &
    \resultsfig{figs/Predictions_Dino/frankfurt_000001_017101_pred_depth_tplus3} & 
    \resultsfig{figs/Predictions_Dino/frankfurt_000001_017101_oracle_depth_tplus3} &
    \resultsfig{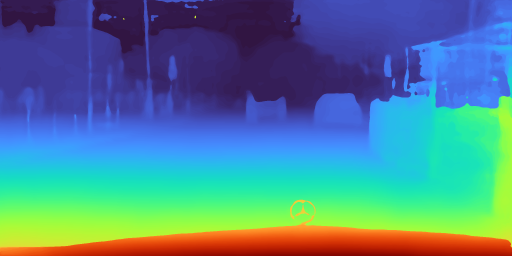} \\

    \mc{4}{\vspace{-2.1ex}}\\

    \raisebox{-.45\height}{\rotatebox{90}{\centering Surf. Normals}}
    &
    \resultsfig{figs/Predictions_Dino/frankfurt_000001_017101_pred_surface_normals_tplus3} & 
    \resultsfig{figs/Predictions_Dino/frankfurt_000001_017101_oracle_surface_normals_tplus3} &
    \resultsfig{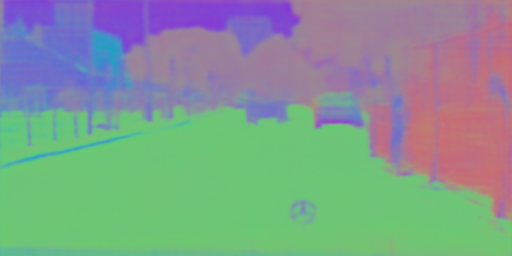} \\

    &
    \mc{3}{(a) Short-Term
    }\\

    \mc{4}{\vspace{-1.6ex}}\\

    & Last Context Frame 
    & Future Frame
    & VISTA (Output)

    \\

    & 
    \resultsfigtwo{figs/Predictions_Dino/frankfurt_000001_017098_leftImg8bit}
    &
    \resultsfigtwo{figs/Predictions_Dino/frankfurt_000001_017107_leftImg8bit}
    & 
    \resultsfigtwo{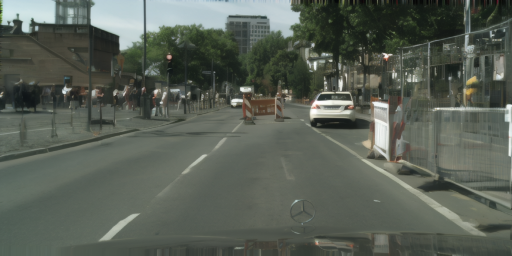}
    \\

    \mc{4}{\vspace{-1.6ex}}\\

    & Ours
    & Oracle
    & VISTA
    \\
    
    \raisebox{-.45\height}{\rotatebox{90}{\centering Segm.}}
    &
    \resultsfig{figs/Predictions_Dino/frankfurt_000001_017101_pred_segm_tplus9} & 
    \resultsfig{figs/Predictions_Dino/frankfurt_000001_017101_oracle_segm_tplus9} &
    \resultsfig{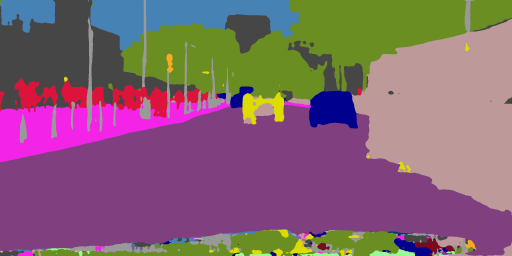} \\

    \mc{4}{\vspace{-2.1ex}}\\

    \raisebox{-.45\height}{\rotatebox{90}{\centering Depth}}
    &
    \resultsfig{figs/Predictions_Dino/frankfurt_000001_017101_pred_depth_tplus9} & 
    \resultsfig{figs/Predictions_Dino/frankfurt_000001_017101_oracle_depth_tplus9} &
    \resultsfig{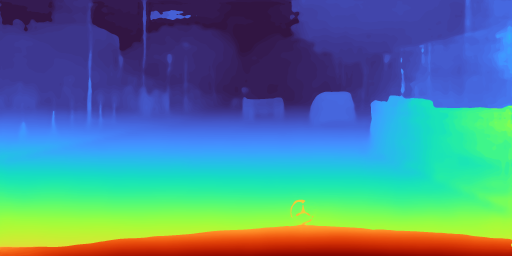} \\

    \mc{4}{\vspace{-2.1ex}}\\

    \raisebox{-.45\height}{\rotatebox{90}{\centering Surf. Normals}}
    &
    \resultsfig{figs/Predictions_Dino/frankfurt_000001_017101_pred_surface_normals_tplus9} & 
    \resultsfig{figs/Predictions_Dino/frankfurt_000001_017101_oracle_surface_normals_tplus9} &
\resultsfig{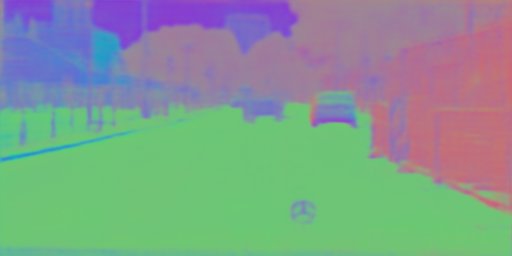} \\

    &
    \mc{3}{(b) Mid-Term
    }\\

\end{tabular}
}

%% file: sec/supp_fig_qualitative_vista_munster.tex
{
\small
\centering
\newcommand{\resultsfig}[1]{\includegraphics[width=0.3\textwidth,valign=c]{#1}}
\newcommand{\resultsfigtwo}[1]{\includegraphics[width=0.3\textwidth,valign=c]{#1}}

\setlength{\tabcolsep}{1pt}

\begin{tabular}{@{}cccc@{}}

    \mc{4}{\vspace{-1.6ex}}\\
    & Last Context Frame 
    & Future Frame 
    & VISTA (Output)

    \\

    &
    \resultsfigtwo{figs/Predictions_Dino/munster_000015_000016_leftImg8bit}
    &
    \resultsfigtwo{figs/Predictions_Dino/munster_000015_000019_leftImg8bit}
     &
\resultsfigtwo{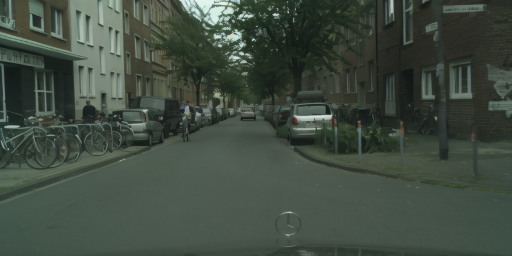}

     \\

    \mc{4}{\vspace{-1.6ex}}\\

    & Ours
    & Oracle
    & VISTA
    \\
    
    \raisebox{-.45\height}{\rotatebox{90}{\centering Segm.}}
    &
    \resultsfig{figs/Predictions_Dino/munster_000015_000019_pred_segm_tplus3} & 
    \resultsfig{figs/Predictions_Dino/munster_000015_000019_oracle_segm_tplus3} &
    \resultsfigtwo{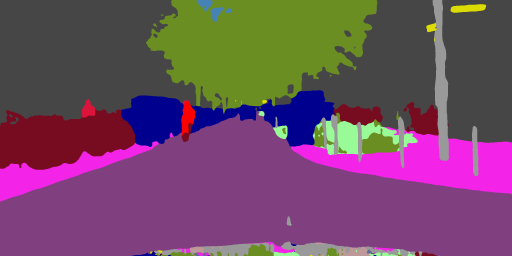} \\

    \mc{4}{\vspace{-2.1ex}}\\

    \raisebox{-.45\height}{\rotatebox{90}{\centering Depth}}
    &
    \resultsfig{figs/Predictions_Dino/munster_000015_000019_pred_depth_tplus3} & 
    \resultsfig{figs/Predictions_Dino/munster_000015_000019_oracle_depth_tplus3} &
    \resultsfigtwo{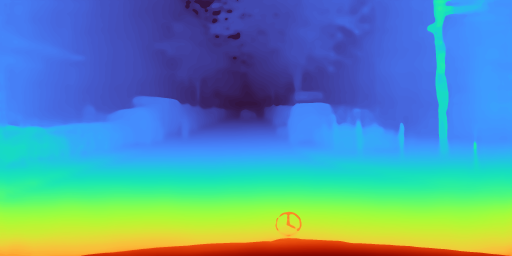} \\

    \mc{4}{\vspace{-2.1ex}}\\

    \raisebox{-.45\height}{\rotatebox{90}{\centering Surf. Normals}}
    &
    \resultsfig{figs/Predictions_Dino/munster_000015_000019_pred_surface_normals_tplus3} & 
    \resultsfig{figs/Predictions_Dino/munster_000015_000019_oracle_surface_normals_tplus3} &
\resultsfigtwo{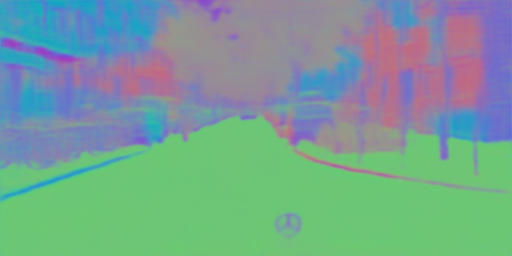} \\

    &
    \mc{3}{(a) Short-Term
    }\\

    \mc{4}{\vspace{-1.6ex}}\\

    & Last Context Frame 
    & Future Frame
    & VISTA (Output)

    \\

    & 
    \resultsfigtwo{figs/Predictions_Dino/munster_000015_000016_leftImg8bit}
    &
    \resultsfigtwo{figs/Predictions_Dino/munster_000015_000025_leftImg8bit}
    & 
\resultsfigtwo{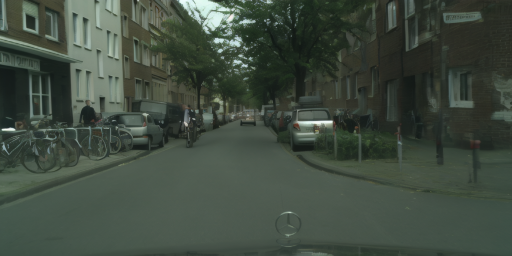}
    \\

    \mc{4}{\vspace{-1.6ex}}\\

    & Ours
    & Oracle
    & VISTA
    \\
    
    \raisebox{-.45\height}{\rotatebox{90}{\centering Segm.}}
    &
    \resultsfig{figs/Predictions_Dino/munster_000015_000019_pred_segm_tplus9} & 
    \resultsfig{figs/Predictions_Dino/munster_000015_000019_oracle_segm_tplus9} &
\resultsfigtwo{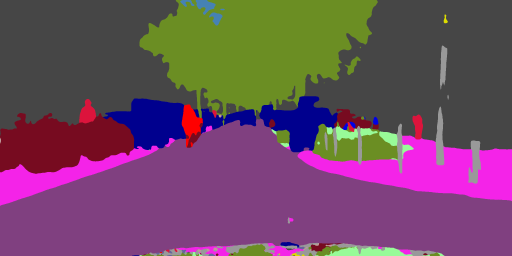} \\

    \mc{4}{\vspace{-2.1ex}}\\

    \raisebox{-.45\height}{\rotatebox{90}{\centering Depth}}
    &
    \resultsfig{figs/Predictions_Dino/munster_000015_000019_pred_depth_tplus9} & 
    \resultsfig{figs/Predictions_Dino/munster_000015_000019_oracle_depth_tplus9} &
\resultsfigtwo{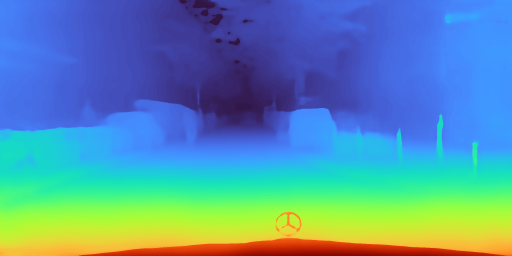} \\

    \mc{4}{\vspace{-2.1ex}}\\

    \raisebox{-.45\height}{\rotatebox{90}{\centering Surf. Normals}}
    &
    \resultsfig{figs/Predictions_Dino/munster_000015_000019_pred_surface_normals_tplus9} & 
    \resultsfig{figs/Predictions_Dino/munster_000015_000019_oracle_surface_normals_tplus9} &
\resultsfigtwo{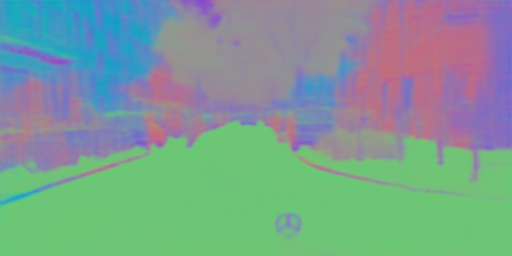} \\

    &
    \mc{3}{(b) Mid-Term
    }\\

\end{tabular}
}

%% file: sec/supp_fig_rollout.tex
{
\small
\centering
\newcommand{\resultsfignew}[1]{\includegraphics[width=0.18\textwidth,valign=c]{#1}}
\setlength{\tabcolsep}{1pt}

\begin{tabular}{@{}cccc@{}}
    
    \mc{4}{\large{Context Frames}} \\
    $X_{t-9}$ & $X_{t-6}$& $X_{t-3}$ & $X_{t}$ \\

    \resultsfignew{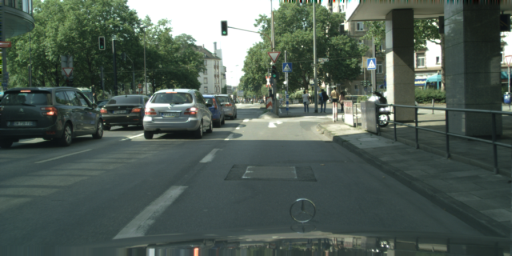} & 
    \resultsfignew{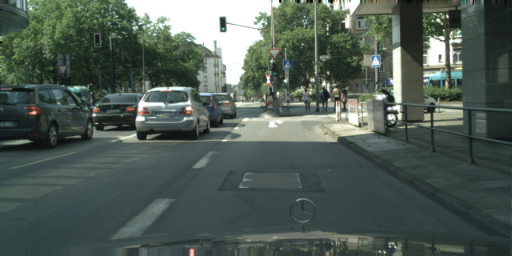} & 
    \resultsfignew{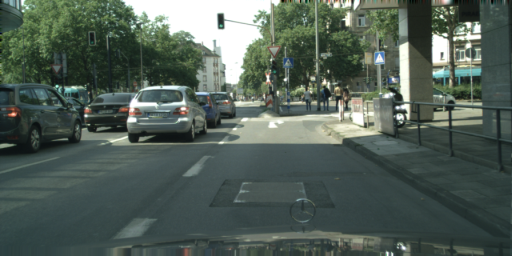} & 
    \resultsfignew{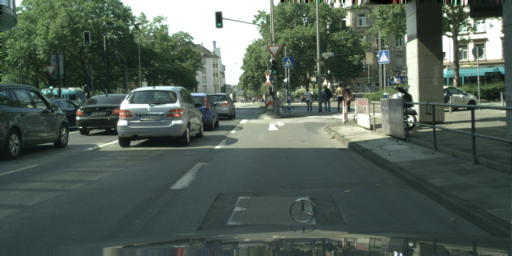}  \\

    \mc{4}{\vspace{-.5ex}}\\

    \mc{4}{\large{Predicted Frames}} \\

    \mc{4}{\vspace{-1.5ex}}\\

    $X_{t+3} (0.18s)$ & $X_{t+6} (0.36s)$& $X_{t+9} (0.54s)$ & $X_{t+12} (0.72s)$ \\

    \resultsfignew{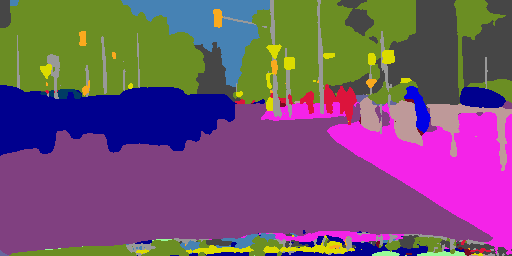} & 
    \resultsfignew{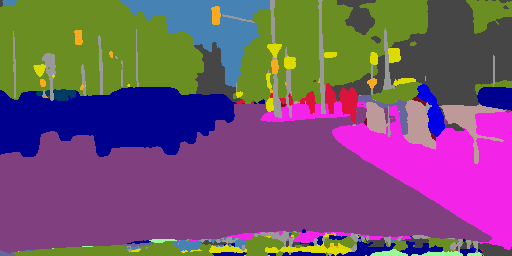} & 
    \resultsfignew{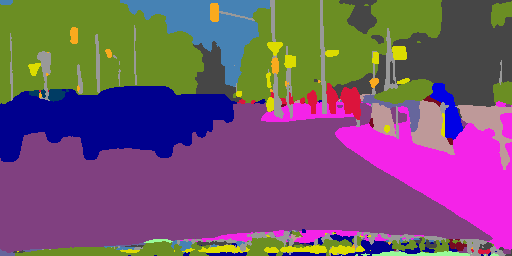} & 
    \resultsfignew{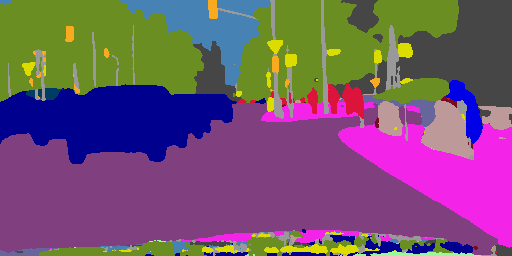}  \\

    \mc{4}{\vspace{-2.0ex}}\\

\resultsfignew{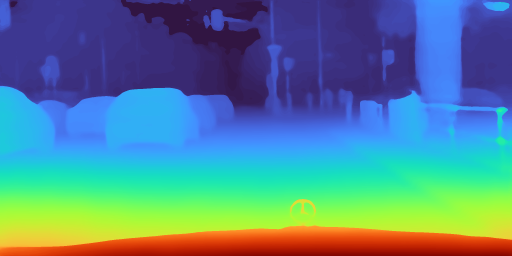} & 
    \resultsfignew{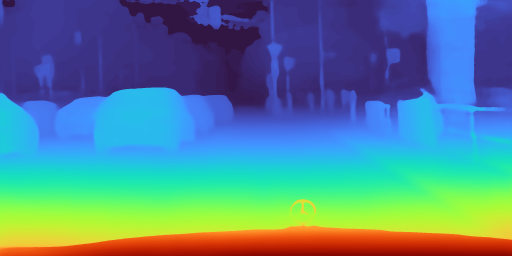} & 
    \resultsfignew{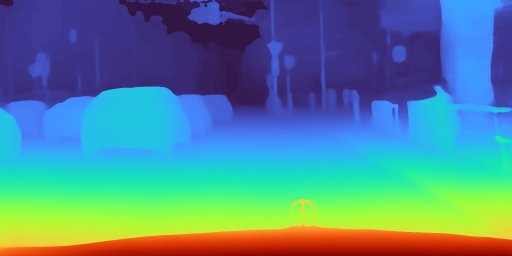} & 
    \resultsfignew{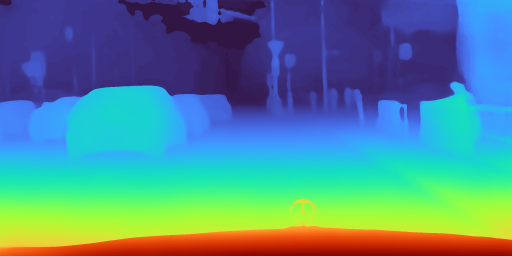}  \\
    
    \mc{4}{\vspace{-2.0ex}}\\

    \resultsfignew{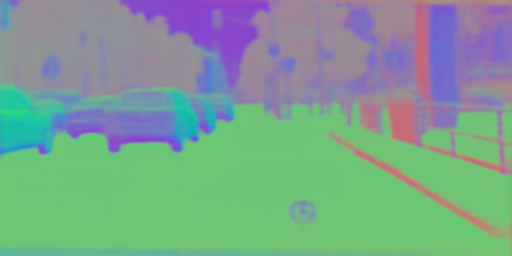} & 
    \resultsfignew{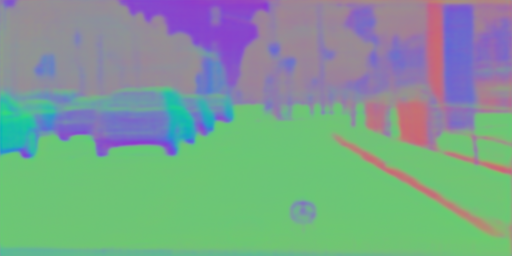} & 
    \resultsfignew{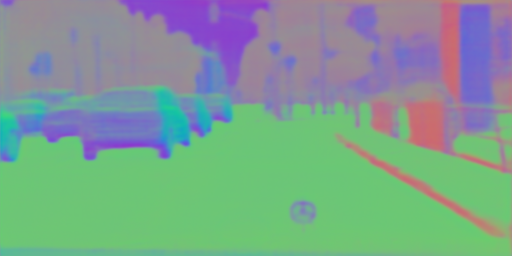} & 
    \resultsfignew{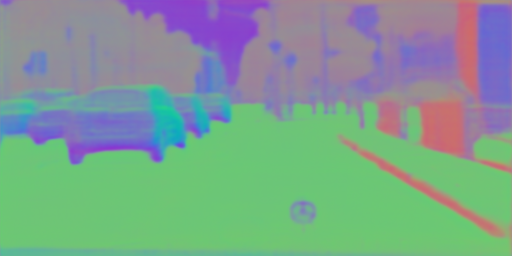}  \\

    \mc{4}{\vspace{-1.5ex}}\\
    
    $X_{t+15} (0.9s)$ & $X_{t+18}(1.08s)$& $X_{t+21} (1.26s)$ & $X_{t+24} (1.44s)$ \\

    \resultsfignew{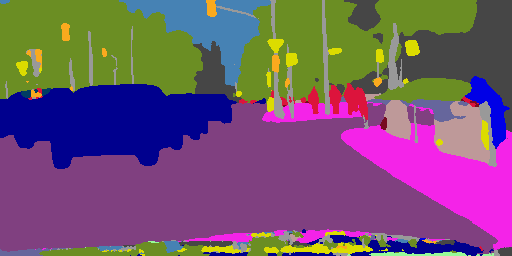} & 
    \resultsfignew{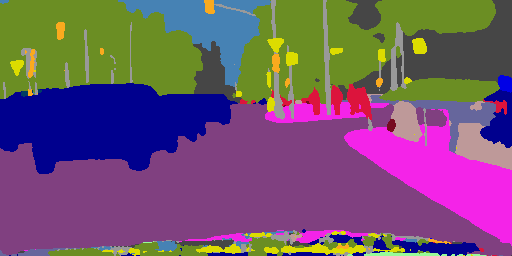} & 
    \resultsfignew{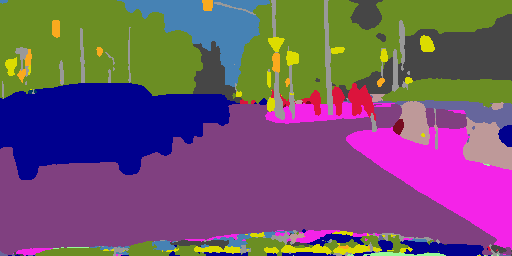} & 
    \resultsfignew{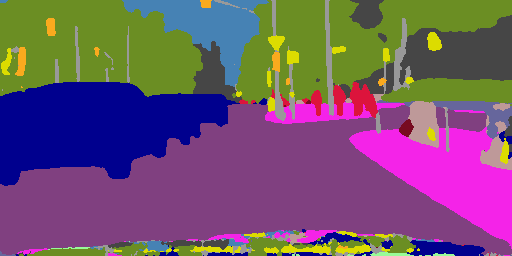}  \\

    \mc{4}{\vspace{-2.0ex}}\\

\resultsfignew{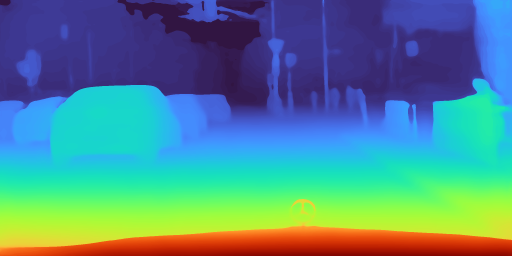} & 
    \resultsfignew{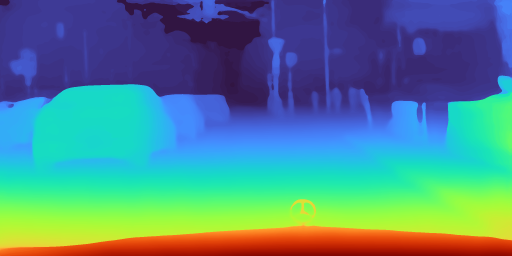} & 
    \resultsfignew{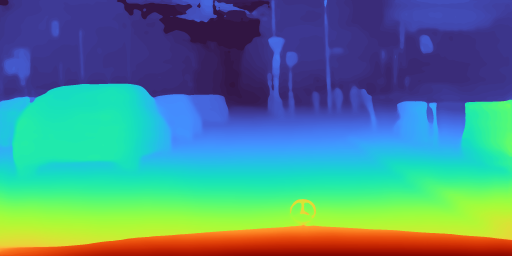} & 
    \resultsfignew{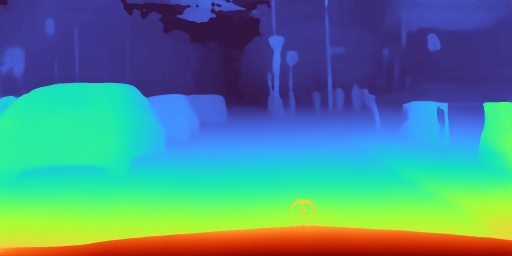}  \\

    \mc{4}{\vspace{-2.0ex}}\\

\resultsfignew{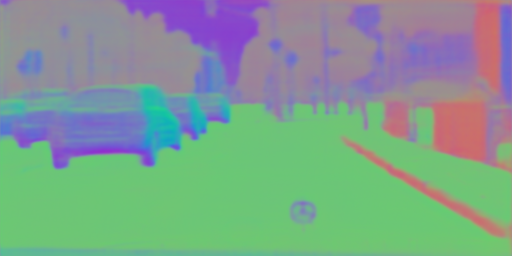} & 
    \resultsfignew{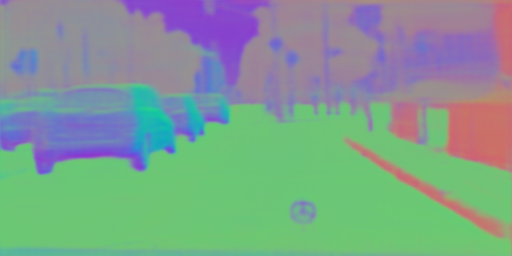} & 
    \resultsfignew{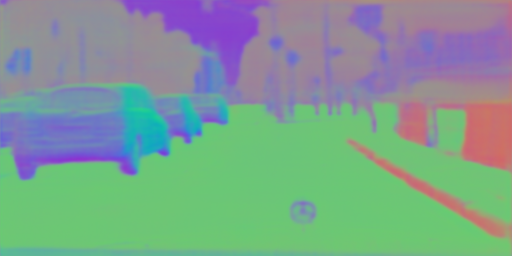} & 
    \resultsfignew{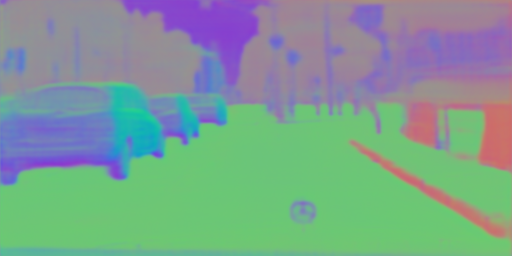}  \\

    \mc{4}{\vspace{-1.5ex}}\\

     $X_{t+27}  (1.62s)$ & $X_{t+30}  (1.8s)$& $X_{t+33}  (1.98s)$ & $X_{t+36}  (2.16s)$ \\

\resultsfignew{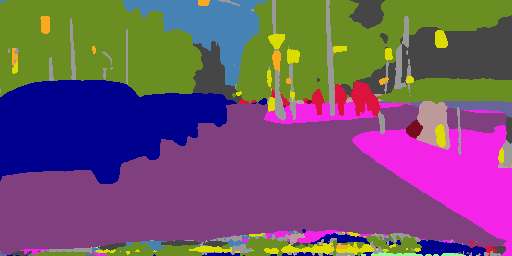} & 
    \resultsfignew{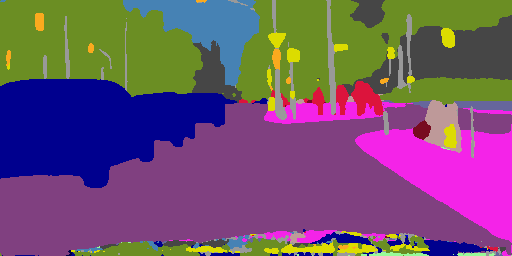} & 
    \resultsfignew{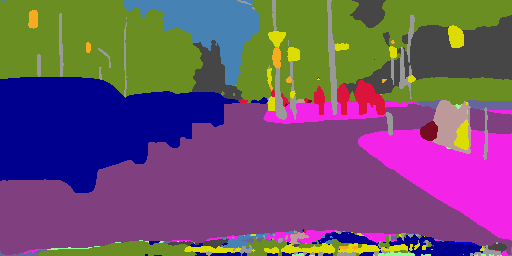} & 
    \resultsfignew{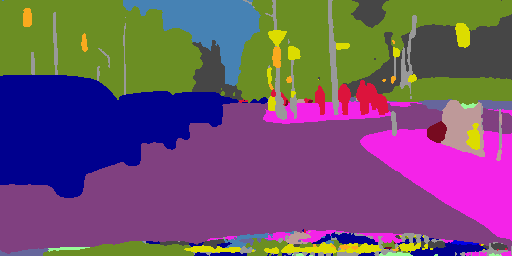}  \\

    \mc{4}{\vspace{-2.0ex}}\\

\resultsfignew{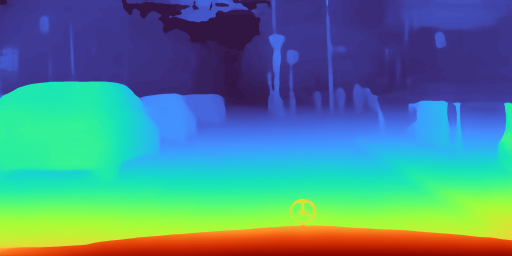} & 
    \resultsfignew{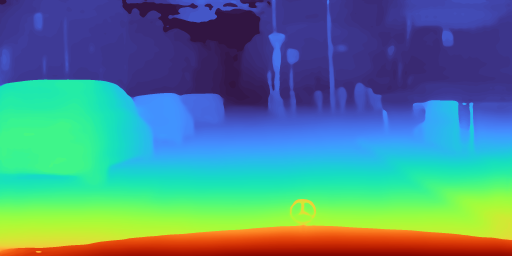} & 
    \resultsfignew{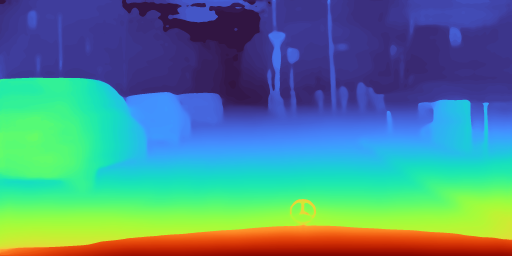} & 
    \resultsfignew{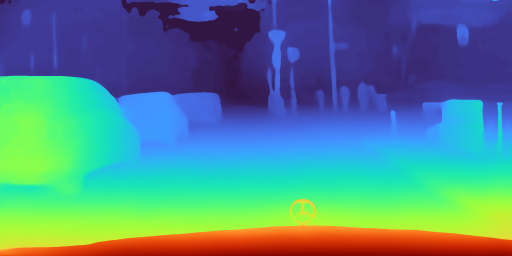}  \\

    \mc{4}{\vspace{-2.0ex}}\\

\resultsfignew{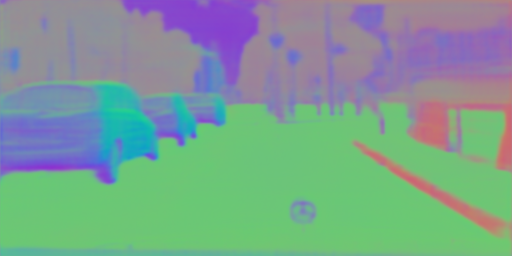} & 
    \resultsfignew{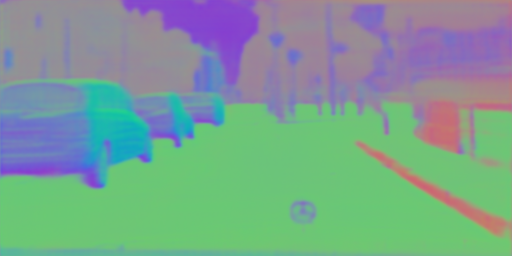} & 
    \resultsfignew{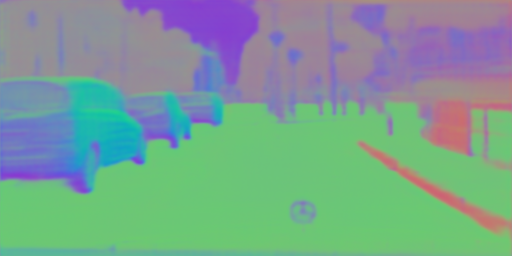} & 
    \resultsfignew{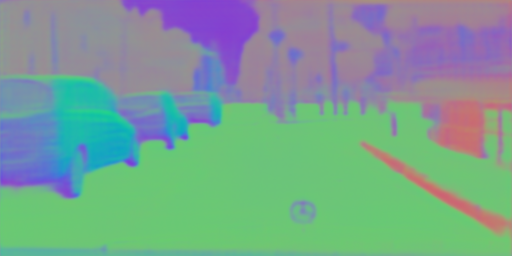}  \\

    \mc{4}{\vspace{-1.5ex}}\\

     $X_{t+39} (2.34s)$ & $X_{t+42} (2.52s)$& $X_{t+45} (2.7s)$ & $X_{t+48} (2.88s)$ \\

\resultsfignew{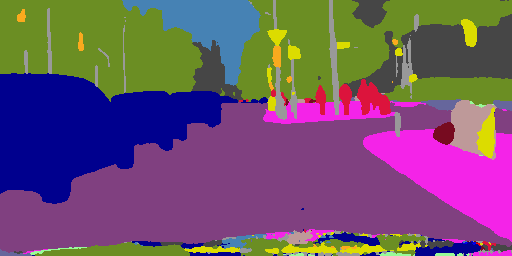} & 
    \resultsfignew{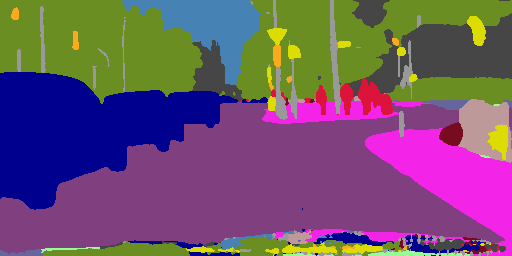} & 
    \resultsfignew{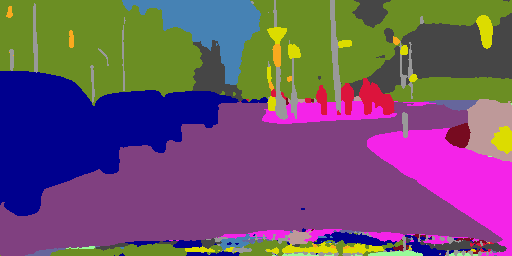} & 
    \resultsfignew{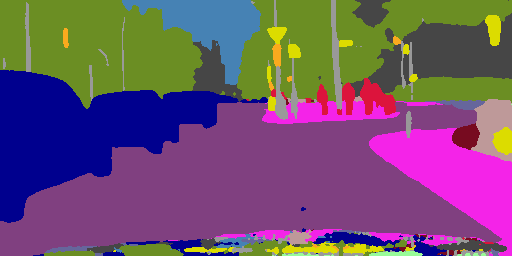}  \\

    \mc{4}{\vspace{-2.0ex}}\\

\resultsfignew{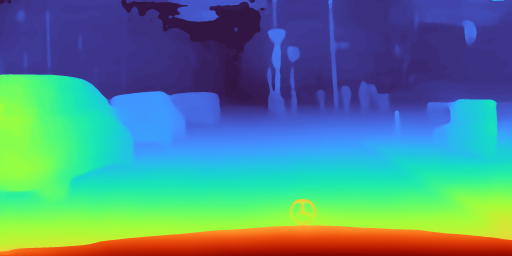} & 
    \resultsfignew{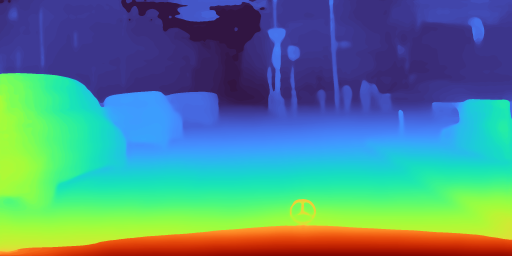} & 
    \resultsfignew{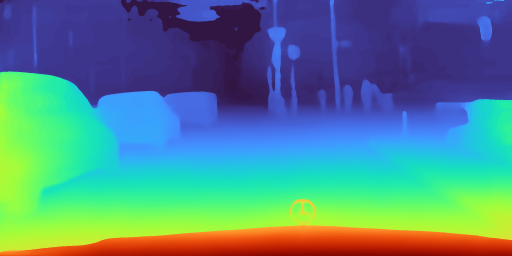} & 
    \resultsfignew{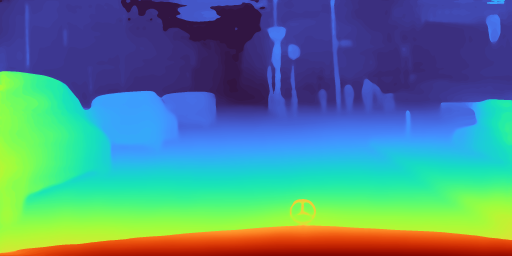}  \\

    \mc{4}{\vspace{-2.0ex}}\\

\resultsfignew{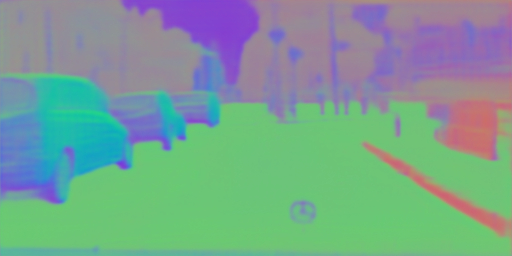} & 
    \resultsfignew{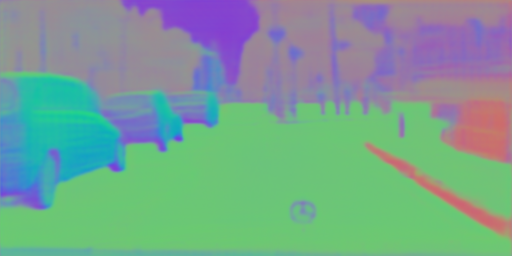} & 
    \resultsfignew{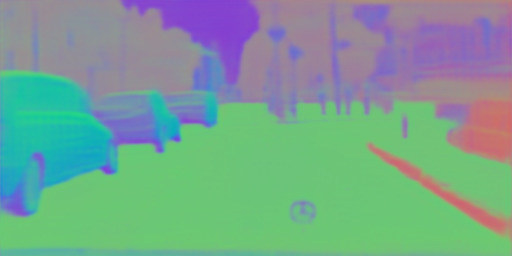} & 
    \resultsfignew{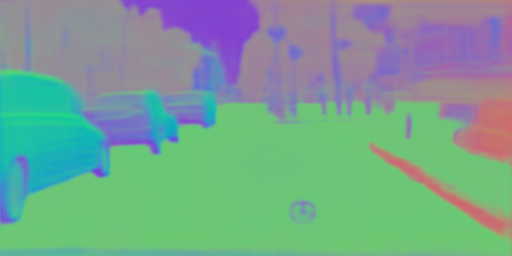}  \\
    
\end{tabular}
}

%% file: sec/supp_fig_rollout_2.tex
{
\small
\centering
\newcommand{\resultsfignew}[1]{\includegraphics[width=0.18\textwidth,valign=c]{#1}}
\setlength{\tabcolsep}{1pt}

\begin{tabular}{@{}cccc@{}}
    
    \mc{4}{\large{Context Frames}} \\
    $X_{t-9}$ & $X_{t-6}$& $X_{t-3}$ & $X_{t}$ \\

    \resultsfignew{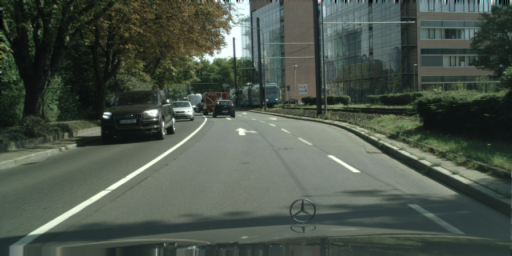} & 
    \resultsfignew{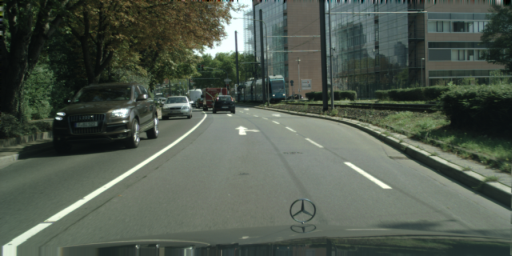} & 
    \resultsfignew{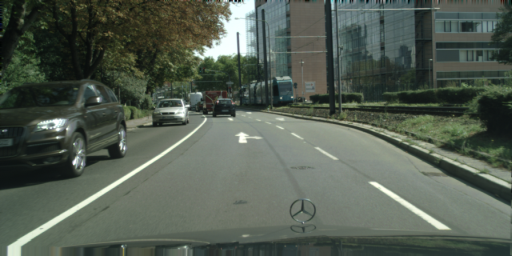} & 
    \resultsfignew{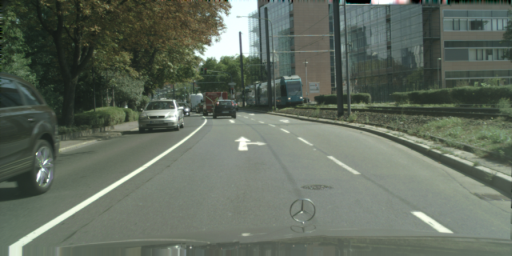}  \\

    \mc{4}{\vspace{-.5ex}}\\

    \mc{4}{\large{Predicted Frames}} \\

    \mc{4}{\vspace{-1.5ex}}\\
    

    $X_{t+3} (0.18s)$ & $X_{t+6} (0.36s)$& $X_{t+9} (0.54s)$ & $X_{t+12} (0.72s)$ \\

    \resultsfignew{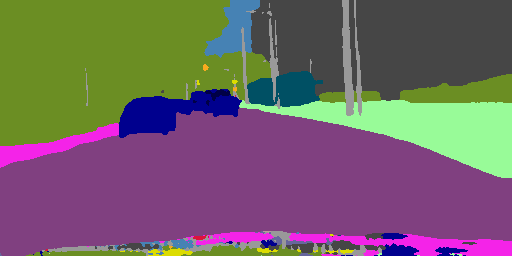} & 
    \resultsfignew{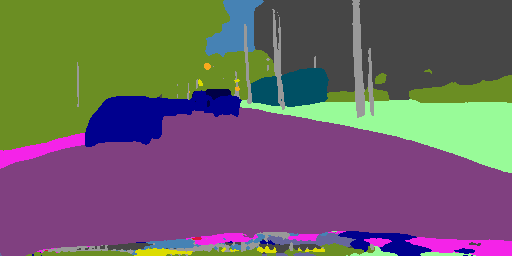} & 
    \resultsfignew{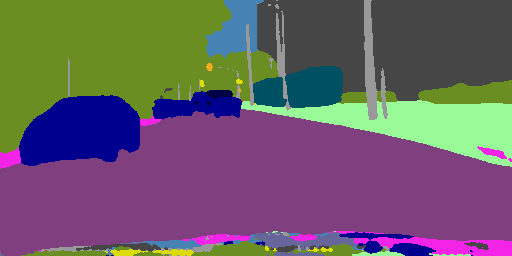} & 
    \resultsfignew{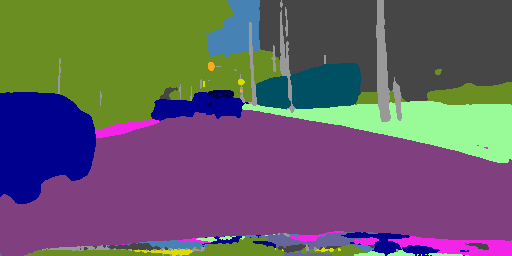}  \\

    \mc{4}{\vspace{-2.0ex}}\\

\resultsfignew{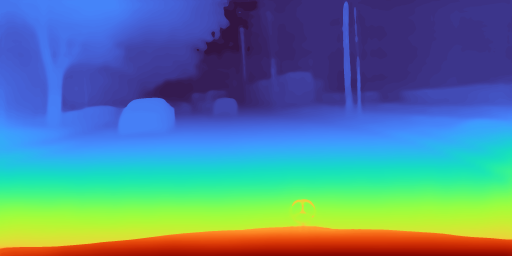} & 
    \resultsfignew{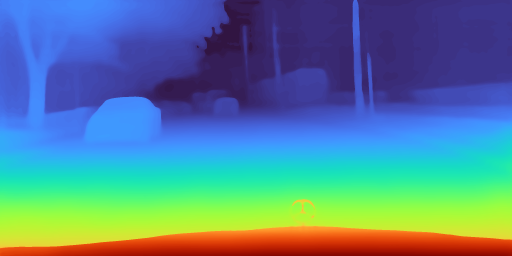} & 
    \resultsfignew{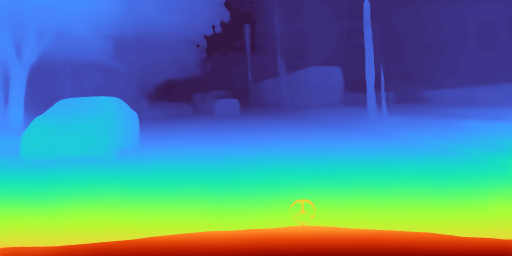} & 
    \resultsfignew{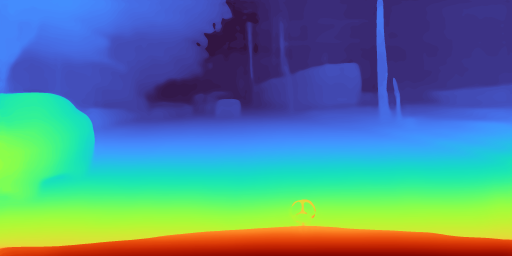}  \\
    
    \mc{4}{\vspace{-2.0ex}}\\

    \resultsfignew{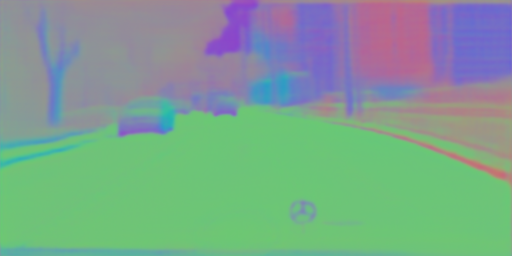} & 
    \resultsfignew{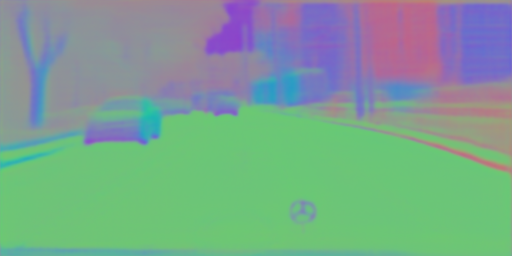} & 
    \resultsfignew{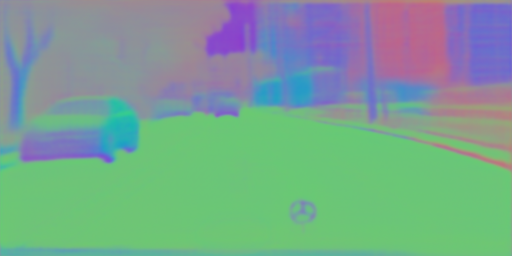} & 
    \resultsfignew{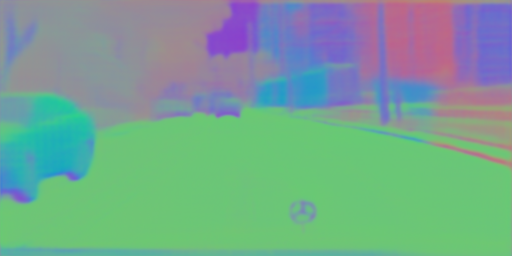}  \\

    \mc{4}{\vspace{-1.5ex}}\\
    
    $X_{t+15} (0.9s)$ & $X_{t+18}(1.08s)$& $X_{t+21} (1.26s)$ & $X_{t+24} (1.44s)$ \\

    \resultsfignew{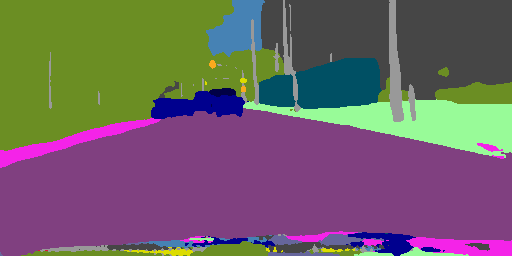} & 
    \resultsfignew{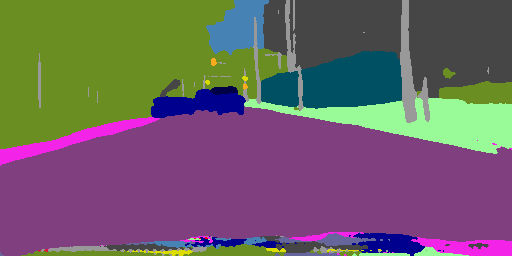} & 
    \resultsfignew{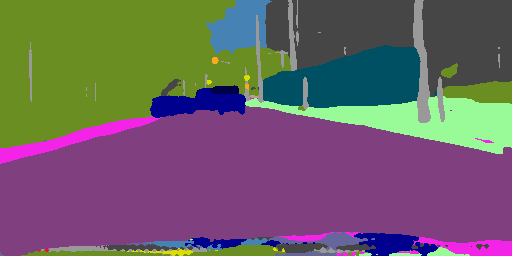} & 
    \resultsfignew{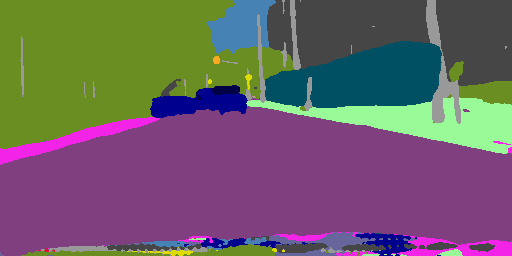}  \\

    \mc{4}{\vspace{-2.0ex}}\\

\resultsfignew{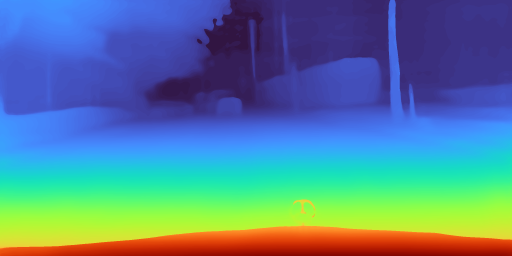} & 
    \resultsfignew{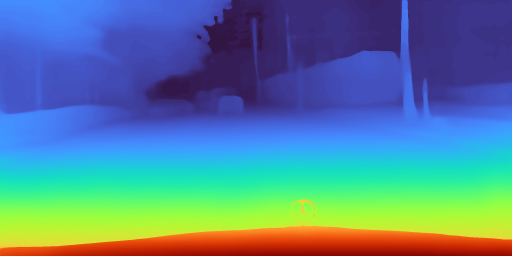} & 
    \resultsfignew{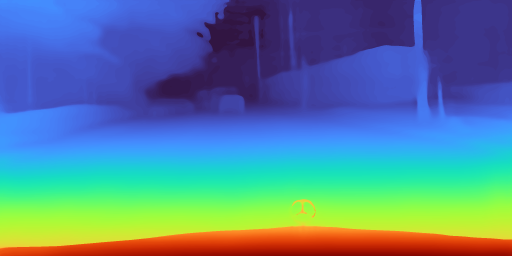} & 
    \resultsfignew{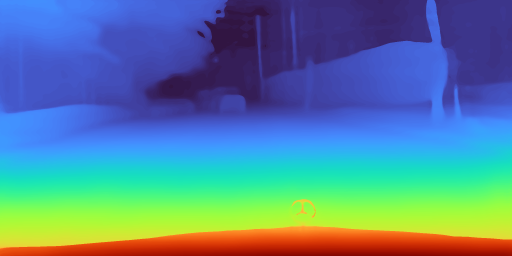}  \\

    \mc{4}{\vspace{-2.0ex}}\\

\resultsfignew{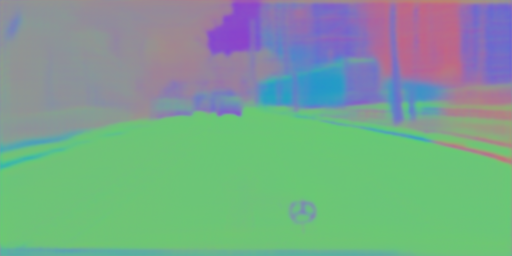} & 
    \resultsfignew{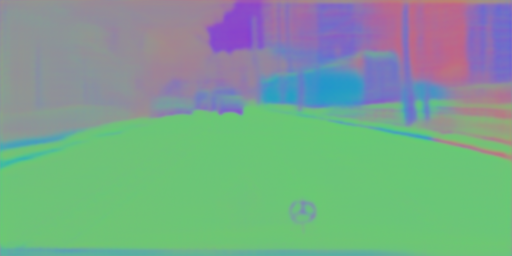} & 
    \resultsfignew{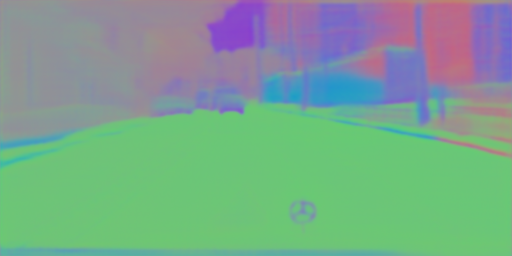} & 
    \resultsfignew{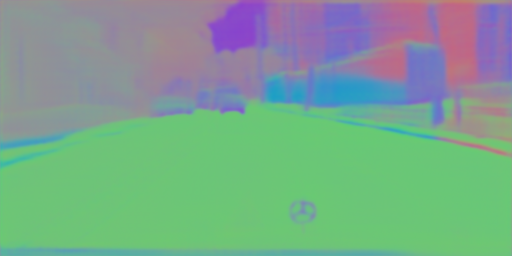}  \\

    \mc{4}{\vspace{-1.5ex}}\\

     $X_{t+27}  (1.62s)$ & $X_{t+30}  (1.8s)$& $X_{t+33}  (1.98s)$ & $X_{t+36}  (2.16s)$ \\

\resultsfignew{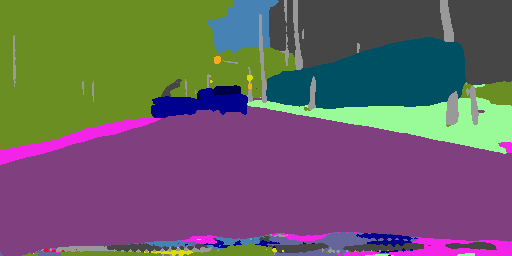} & 
    \resultsfignew{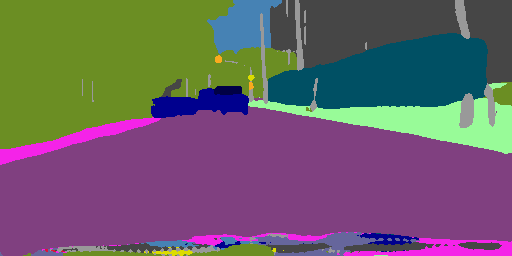} & 
    \resultsfignew{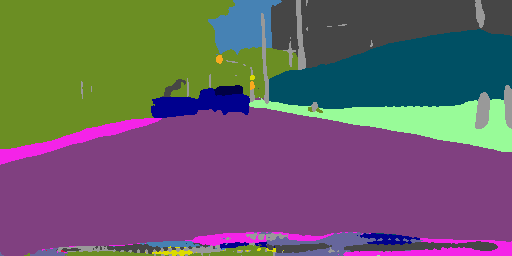} & 
    \resultsfignew{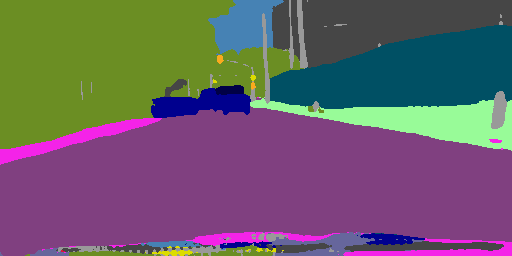}  \\

    \mc{4}{\vspace{-2.0ex}}\\

\resultsfignew{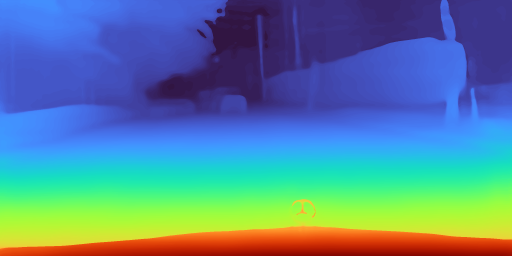} & 
    \resultsfignew{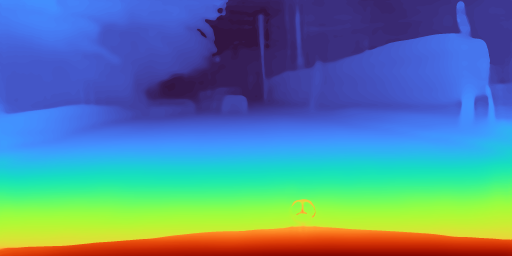} & 
    \resultsfignew{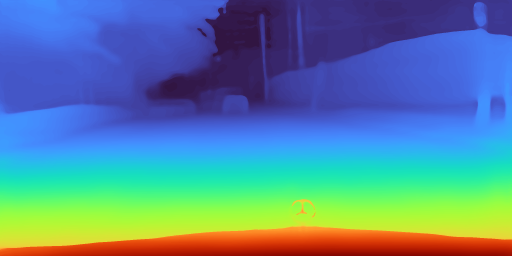} & 
    \resultsfignew{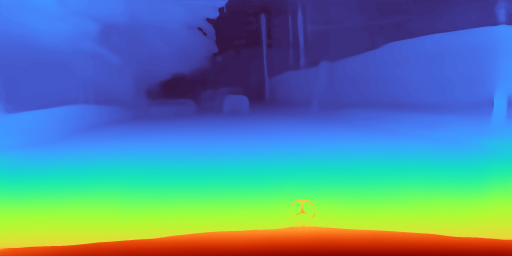}  \\

    \mc{4}{\vspace{-2.0ex}}\\

\resultsfignew{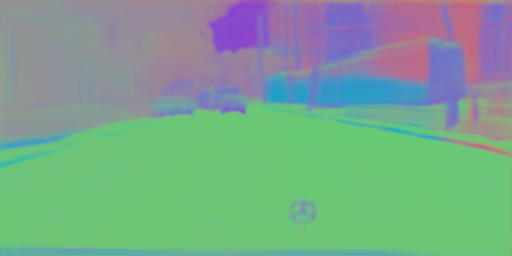} & 
    \resultsfignew{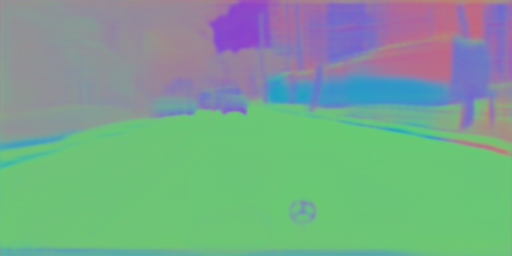} & 
    \resultsfignew{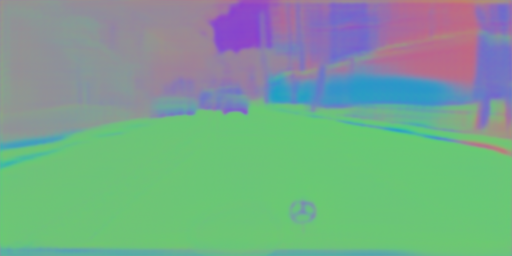} & 
    \resultsfignew{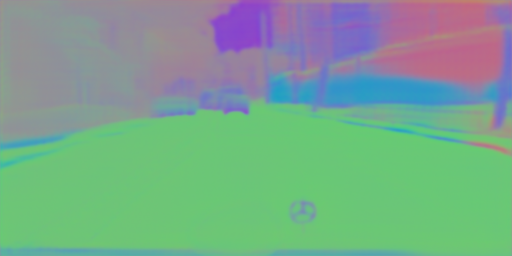}  \\

    \mc{4}{\vspace{-1.5ex}}\\

     $X_{t+39} (2.34s)$ & $X_{t+42} (2.52s)$& $X_{t+45} (2.7s)$ & $X_{t+48} (2.88s)$ \\

\resultsfignew{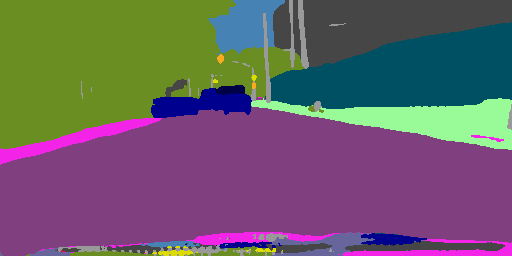} & 
    \resultsfignew{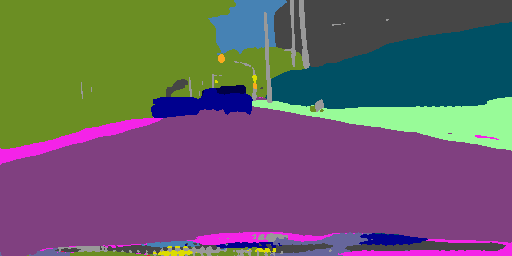} & 
    \resultsfignew{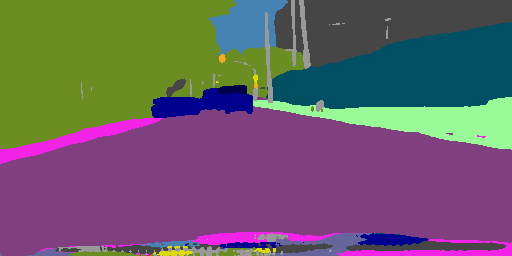} & 
    \resultsfignew{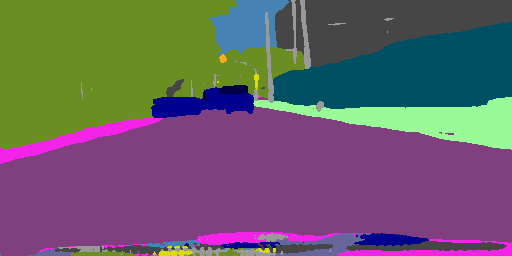}  \\

    \mc{4}{\vspace{-2.0ex}}\\

\resultsfignew{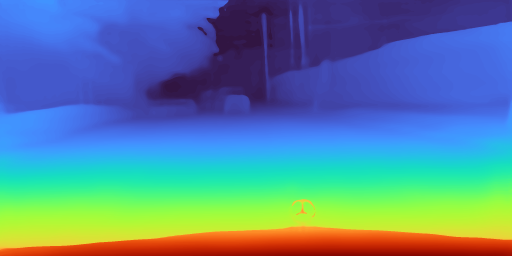} & 
    \resultsfignew{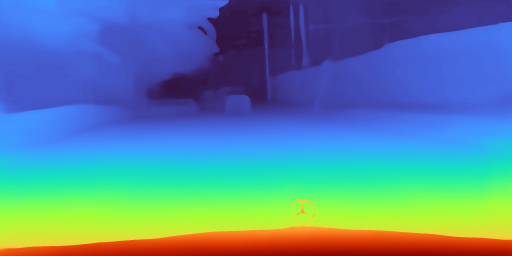} & 
    \resultsfignew{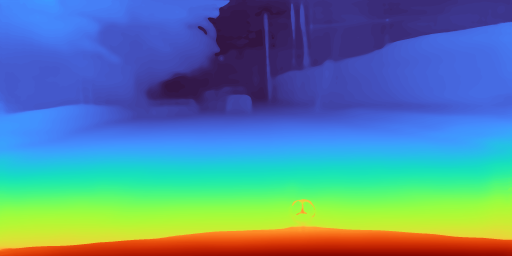} & 
    \resultsfignew{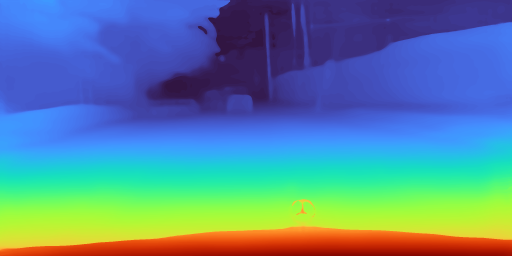}  \\

    \mc{4}{\vspace{-2.0ex}}\\

\resultsfignew{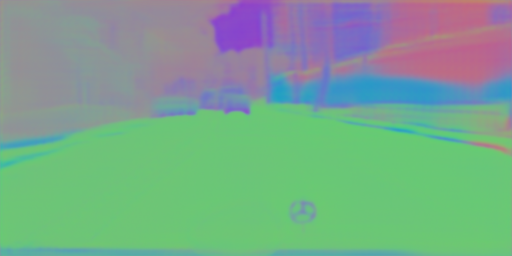} & 
    \resultsfignew{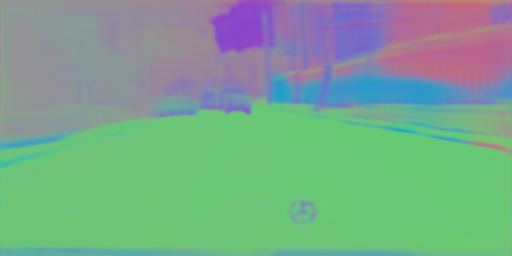} & 
    \resultsfignew{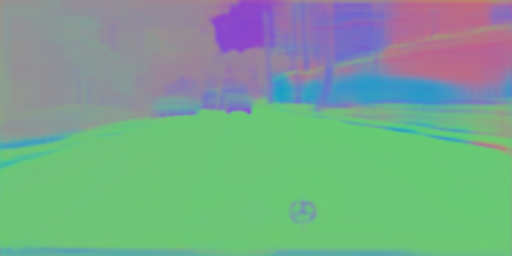} & 
    \resultsfignew{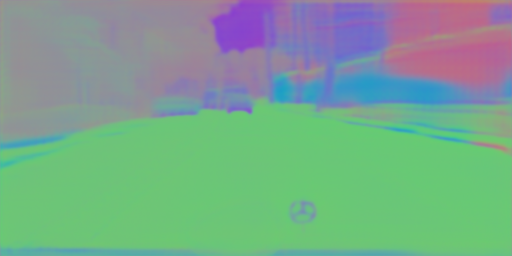}  \\
    
\end{tabular}
}

%% file: main_neurips_2025.bbl
\begin{thebibliography}{113}
\providecommand{\natexlab}[1]{#1}
\providecommand{\url}[1]{\texttt{#1}}
\expandafter\ifx\csname urlstyle\endcsname\relax
  \providecommand{\doi}[1]{doi: #1}\else
  \providecommand{\doi}{doi: \begingroup \urlstyle{rm}\Url}\fi

\bibitem[Aich et~al.(2023)Aich, Schulter, Roy-Chowdhury, Chandraker, and Suh]{aich2023efficient}
Abhishek Aich, Samuel Schulter, Amit~K. Roy-Chowdhury, Manmohan Chandraker, and Yumin Suh.
\newblock Efficient controllable multi-task architectures.
\newblock In \emph{ICCV}, 2023.

\bibitem[Arnab et~al.(2021)Arnab, Dehghani, Heigold, Sun, Lu{\v{c}}i{\'c}, and Schmid]{arnab2021vivit}
Anurag Arnab, Mostafa Dehghani, Georg Heigold, Chen Sun, Mario Lu{\v{c}}i{\'c}, and Cordelia Schmid.
\newblock Vivit: A video vision transformer.
\newblock In \emph{ICCV}, 2021.

\bibitem[Babaeizadeh et~al.(2018)Babaeizadeh, Finn, Erhan, Campbell, and Levine]{babaeizadeh2018stochastic}
Mohammad Babaeizadeh, Chelsea Finn, Dumitru Erhan, Roy~H. Campbell, and Sergey Levine.
\newblock Stochastic variational video prediction.
\newblock In \emph{ICLR}, 2018.

\bibitem[Bachmann et~al.(2024)Bachmann, Kar, Mizrahi, Garjani, Gao, Griffiths, Hu, Dehghan, and Zamir]{4m21}
Roman Bachmann, O\u{g}uzhan~Fatih Kar, David Mizrahi, Ali Garjani, Mingfei Gao, David Griffiths, Jiaming Hu, Afshin Dehghan, and Amir Zamir.
\newblock 4m-21: An any-to-any vision model for tens of tasks and modalities.
\newblock In \emph{NeurIPS}, 2024.

\bibitem[Bardes et~al.(2024)Bardes, Garrido, Ponce, Chen, Rabbat, LeCun, Assran, and Ballas]{bardes2024revisiting}
Adrien Bardes, Quentin Garrido, Jean Ponce, Xinlei Chen, Michael Rabbat, Yann LeCun, Mido Assran, and Nicolas Ballas.
\newblock Revisiting feature prediction for learning visual representations from video.
\newblock \emph{Transactions on Machine Learning Research}, 2024.

\bibitem[Bekoulis et~al.(2018)Bekoulis, Deleu, Demeester, and Develder]{bekoulis-etal-2018-adversarial}
Giannis Bekoulis, Johannes Deleu, Thomas Demeester, and Chris Develder.
\newblock Adversarial training for multi-context joint entity and relation extraction.
\newblock In \emph{NeurIPS}, 2018.

\bibitem[Besnier and Chen(2023)]{besnier2023pytorch}
Victor Besnier and Mickael Chen.
\newblock A pytorch reproduction of masked generative image transformer.
\newblock \emph{arXiv preprint arXiv:2310.14400}, 2023.

\bibitem[Bhattacharyya et~al.(2019)Bhattacharyya, Fritz, and Schiele]{Bayesian_s2s}
Apratim Bhattacharyya, Mario Fritz, and Bernt Schiele.
\newblock Bayesian prediction of future street scenes using synthetic likelihoods.
\newblock In \emph{ICLR}, 2019.

\bibitem[Bragman et~al.(2019)Bragman, Tanno, Ourselin, Alexander, and Cardoso]{bragman2019stochastic}
Felix~JS Bragman, Ryutaro Tanno, Sebastien Ourselin, Daniel~C. Alexander, and Jorge Cardoso.
\newblock Stochastic filter groups for multi-task {CNNs}: Learning specialist and generalist convolution kernels.
\newblock In \emph{{ICCV}}, 2019.

\bibitem[Brooks et~al.(2024)Brooks, Peebles, Holmes, DePue, Guo, Jing, Schnurr, Taylor, Luhman, Luhman, et~al.]{videoworldsimulators2024}
Tim Brooks, Bill Peebles, Connor Holmes, Will DePue, Yufei Guo, Li Jing, David Schnurr, Joe Taylor, Troy Luhman, Eric Luhman, et~al.
\newblock Video generation models as world simulators.
\newblock \emph{OpenAI Blog}, 1:\penalty0 8, 2024.

\bibitem[Caesar et~al.(2020)Caesar, Bankiti, Lang, Vora, Liong, Xu, Krishnan, Pan, Baldan, and Beijbom]{nuscenes}
Holger Caesar, Varun Bankiti, Alex~H. Lang, Sourabh Vora, Venice~Erin Liong, Qiang Xu, Anush Krishnan, Yu Pan, Giancarlo Baldan, and Oscar Beijbom.
\newblock nuscenes: A multimodal dataset for autonomous driving.
\newblock In \emph{CVPR}, 2020.

\bibitem[Caron et~al.(2021)Caron, Touvron, Misra, Jegou, Mairal, Bojanowski, and Joulin]{dino}
Mathilde Caron, Hugo Touvron, Ishan Misra, Hervé Jegou, Julien Mairal, Piotr Bojanowski, and Armand Joulin.
\newblock Emerging properties in self-supervised vision transformers.
\newblock In \emph{{ICCV}}, 2021.

\bibitem[Castrejon et~al.(2019)Castrejon, Ballas, and Courville]{castrejon2019improved}
Lluis Castrejon, Nicolas Ballas, and Aaron Courville.
\newblock Improved conditional vrnns for video prediction.
\newblock In \emph{CVPR}, 2019.

\bibitem[Chang et~al.(2022)Chang, Zhang, Jiang, Liu, and Freeman]{chang2022maskgit}
Huiwen Chang, Han Zhang, Lu Jiang, Ce Liu, and William~T Freeman.
\newblock Maskgit: Masked generative image transformer.
\newblock In \emph{CVPR}, 2022.

\bibitem[Chen et~al.(2017)Chen, Papandreou, Kokkinos, Murphy, and Yuille]{chen2017deeplab}
Liang-Chieh Chen, George Papandreou, Iasonas Kokkinos, Kevin Murphy, and Alan~L Yuille.
\newblock Deeplab: Semantic image segmentation with deep convolutional nets, atrous convolution, and fully connected crfs.
\newblock \emph{IEEE Transactions on Pattern Analysis and Machine Intelligence}, 2017.

\bibitem[Chen et~al.(2020)Chen, Kornblith, Norouzi, and Hinton]{simclr}
Ting Chen, Simon Kornblith, Mohammad Norouzi, and Geoffrey Hinton.
\newblock A simple framework for contrastive learning of visual representations.
\newblock In \emph{ICML}, 2020.

\bibitem[Chen et~al.(2023{\natexlab{a}})Chen, Chen, Du, Rashwan, Yang, Chen, Wang, and Li]{chen2023adamv}
Tianlong Chen, Xuxi Chen, Xianzhi Du, Abdullah Rashwan, Fan Yang, Huizhong Chen, Zhangyang Wang, and Yeqing Li.
\newblock Adamv-moe: Adaptive multi-task vision mixture-of-experts.
\newblock In \emph{{ICCV}}, 2023{\natexlab{a}}.

\bibitem[Chen and Han(2019)]{chen2019multi}
Xin Chen and Yahong Han.
\newblock Multi-timescale context encoding for scene parsing prediction.
\newblock In \emph{2019 IEEE International Conference on Multimedia and Expo (ICME)}, 2019.

\bibitem[Chen et~al.(2023{\natexlab{b}})Chen, Shen, Ding, Chen, Zhao, Learned-Miller, and Gan]{chen2023modsquad}
Zitian Chen, Yikang Shen, Mingyu Ding, Zhenfang Chen, Hengshuang Zhao, Erik~G. Learned-Miller, and Chuang Gan.
\newblock Mod-squad: Designing mixtures of experts as modular multi-task learners.
\newblock In \emph{CVPR}, 2023{\natexlab{b}}.

\bibitem[Cheng et~al.(2022)Cheng, Misra, Schwing, Kirillov, and Girdhar]{cheng2022masked}
Bowen Cheng, Ishan Misra, Alexander~G Schwing, Alexander Kirillov, and Rohit Girdhar.
\newblock Masked-attention mask transformer for universal image segmentation.
\newblock In \emph{CVPR}, 2022.

\bibitem[Chiu et~al.(2020)Chiu, Adeli, and Niebles]{chiu2020segmenting}
Hsu-kuang Chiu, Ehsan Adeli, and Juan~Carlos Niebles.
\newblock Segmenting the future.
\newblock \emph{IEEE Robotics and Automation Letters}, 2020.

\bibitem[Choi and Im(2023)]{choi2023dynamic}
Wonhyeok Choi and Sunghoon Im.
\newblock Dynamic neural network for multi-task learning searching across diverse network topologies.
\newblock In \emph{CVPR}, 2023.

\bibitem[Cordts et~al.(2016)Cordts, Omran, Ramos, Rehfeld, Enzweiler, Benenson, Franke, Roth, and Schiele]{Cordts_2016_CVPR}
Marius Cordts, Mohamed Omran, Sebastian Ramos, Timo Rehfeld, Markus Enzweiler, Rodrigo Benenson, Uwe Franke, Stefan Roth, and Bernt Schiele.
\newblock The cityscapes dataset for semantic urban scene understanding.
\newblock In \emph{CVPR}, 2016.

\bibitem[Darcet et~al.(2024)Darcet, Oquab, Mairal, and Bojanowski]{darcet2024vision}
Timoth{\'e}e Darcet, Maxime Oquab, Julien Mairal, and Piotr Bojanowski.
\newblock Vision transformers need registers.
\newblock In \emph{ICLR}, 2024.

\bibitem[Dosovitskiy and Koltun(2017)]{dosovitskiy2017learning}
Alexey Dosovitskiy and Vladlen Koltun.
\newblock Learning to act by predicting the future.
\newblock In \emph{ICLR}, 2017.

\bibitem[Dosovitskiy et~al.(2020)Dosovitskiy, Beyer, Kolesnikov, Weissenborn, Zhai, Unterthiner, Dehghani, Minderer, Heigold, Gelly, et~al.]{dosovitskiy2020image}
Alexey Dosovitskiy, Lucas Beyer, Alexander Kolesnikov, Dirk Weissenborn, Xiaohua Zhai, Thomas Unterthiner, Mostafa Dehghani, Matthias Minderer, Georg Heigold, Sylvain Gelly, et~al.
\newblock An image is worth 16x16 words: Transformers for image recognition at scale.
\newblock In \emph{ICLR}, 2020.

\bibitem[Esser et~al.(2021)Esser, Rombach, and Ommer]{esser2021taming}
Patrick Esser, Robin Rombach, and Bjorn Ommer.
\newblock Taming transformers for high-resolution image synthesis.
\newblock In \emph{CVPR}, 2021.

\bibitem[Fang et~al.(2023)Fang, Wang, Xie, Sun, Wu, Wang, Huang, Wang, and Cao]{fang2023eva}
Yuxin Fang, Wen Wang, Binhui Xie, Quan Sun, Ledell Wu, Xinggang Wang, Tiejun Huang, Xinlong Wang, and Yue Cao.
\newblock Eva: Exploring the limits of masked visual representation learning at scale.
\newblock In \emph{CVPR}, 2023.

\bibitem[Fang et~al.(2024)Fang, Sun, Wang, Huang, Wang, and Cao]{fang2024eva}
Yuxin Fang, Quan Sun, Xinggang Wang, Tiejun Huang, Xinlong Wang, and Yue Cao.
\newblock Eva-02: A visual representation for neon genesis.
\newblock \emph{Image and Vision Computing}, 2024.

\bibitem[Finn et~al.(2016)Finn, Goodfellow, and Levine]{ULphysinter}
Chelsea Finn, Ian Goodfellow, and Sergey Levine.
\newblock Unsupervised learning for physical interaction through video prediction.
\newblock In \emph{NeurIPS}, 2016.

\bibitem[Gao et~al.(2024)Gao, Yang, Chen, Chitta, Qiu, Geiger, Zhang, and Li]{gao2024vista}
Shenyuan Gao, Jiazhi Yang, Li Chen, Kashyap Chitta, Yihang Qiu, Andreas Geiger, Jun Zhang, and Hongyang Li.
\newblock Vista: A generalizable driving world model with high fidelity and versatile controllability.
\newblock In \emph{NeurIPS}, 2024.

\bibitem[Gao et~al.(2022)Gao, Tan, Wu, and Li]{gao2022simvp}
Zhangyang Gao, Cheng Tan, Lirong Wu, and Stan~Z Li.
\newblock Simvp: Simpler yet better video prediction.
\newblock In \emph{CVPR}, 2022.

\bibitem[Gidaris et~al.(2024)Gidaris, Bursuc, Sim{\'e}oni, Vobeck{\'y}, Komodakis, Cord, and Perez]{gidaris2024moca}
Spyros Gidaris, Andrei Bursuc, Oriane Sim{\'e}oni, Anton{\'\i}n Vobeck{\'y}, Nikos Komodakis, Matthieu Cord, and Patrick Perez.
\newblock {MOCA}: Self-supervised representation learning by predicting masked online codebook assignments.
\newblock \emph{Transactions on Machine Learning Research}, 2024.

\bibitem[Girdhar et~al.(2023)Girdhar, El-Nouby, Singh, Alwala, Joulin, and Misra]{omnimae}
Rohit Girdhar, Alaaeldin El-Nouby, Mannat Singh, Kalyan~Vasudev Alwala, Armand Joulin, and Ishan Misra.
\newblock { OmniMAE: Single Model Masked Pretraining on Images and Videos }.
\newblock In \emph{CVPR}, 2023.

\bibitem[Graber et~al.(2021)Graber, Tsai, Firman, Brostow, and Schwing]{indrnn_stack}
Colin Graber, Grace Tsai, Michael Firman, Gabriel Brostow, and Alexander Schwing.
\newblock { Panoptic Segmentation Forecasting }.
\newblock In \emph{CVPR}, 2021.

\bibitem[Graber et~al.(2022)Graber, Jazra, Luo, Gui, and Schwing]{graber2022joint}
Colin Graber, Cyril Jazra, Wenjie Luo, Liangyan Gui, and Alexander~G Schwing.
\newblock Joint forecasting of panoptic segmentations with difference attention.
\newblock In \emph{CVPR}, 2022.

\bibitem[Grill et~al.(2020)Grill, Strub, Altch\'{e}, Tallec, Richemond, Buchatskaya, Doersch, Pires, Guo, Azar, Piot, Kavukcuoglu, Munos, and Valko]{byol}
Jean-Bastien Grill, Florian Strub, Florent Altch\'{e}, Corentin Tallec, Pierre~H. Richemond, Elena Buchatskaya, Carl Doersch, Bernardo~Avila Pires, Zhaohan~Daniel Guo, Mohammad~Gheshlaghi Azar, Bilal Piot, Koray Kavukcuoglu, R\'{e}mi Munos, and Michal Valko.
\newblock Bootstrap your own latent a new approach to self-supervised learning.
\newblock In \emph{NeurIPS}, 2020.

\bibitem[Guo et~al.(2020)Guo, Lee, and Ulbricht]{pmlr-v119-guo20e}
Pengsheng Guo, Chen-Yu Lee, and Daniel Ulbricht.
\newblock Learning to branch for multi-task learning.
\newblock In \emph{ICML}, 2020.

\bibitem[Gupta et~al.(2023)Gupta, Tian, Zhang, Wu, Mart{\'\i}n-Mart{\'\i}n, and Fei-Fei]{gupta2023maskvit}
Agrim Gupta, Stephen Tian, Yunzhi Zhang, Jiajun Wu, Roberto Mart{\'\i}n-Mart{\'\i}n, and Li Fei-Fei.
\newblock Maskvit: Masked visual pre-training for video prediction.
\newblock In \emph{ICLR}, 2023.

\bibitem[Harvey et~al.(2022)Harvey, Naderiparizi, Masrani, Weilbach, and Wood]{harvey2022flexible}
William Harvey, Saeid Naderiparizi, Vaden Masrani, Christian~Dietrich Weilbach, and Frank Wood.
\newblock Flexible diffusion modeling of long videos.
\newblock In \emph{NeurIPS}, 2022.

\bibitem[He et~al.(2025)He, LI, Yin, Liang, Li, Zhou, Zhang, Liu, and Chen]{he2025lotus}
Jing He, Haodong LI, Wei Yin, Yixun Liang, Leheng Li, Kaiqiang Zhou, Hongbo Zhang, Bingbing Liu, and Ying-Cong Chen.
\newblock Lotus: Diffusion-based visual foundation model for high-quality dense prediction.
\newblock In \emph{ICLR}, 2025.

\bibitem[He et~al.(2022)He, Chen, Xie, Li, Dollár, and Girshick]{mae}
Kaiming He, Xinlei Chen, Saining Xie, Yanghao Li, Piotr Dollár, and Ross Girshick.
\newblock Masked autoencoders are scalable vision learners.
\newblock In \emph{CVPR}, 2022.

\bibitem[Heinrich et~al.(2025)Heinrich, Ranzinger, Yin, Lu, Kautz, Tao, Catanzaro, and Molchanov]{heinrich2025radiov25improvedbaselinesagglomerative}
Greg Heinrich, Mike Ranzinger, Hongxu Yin, Yao Lu, Jan Kautz, Andrew Tao, Bryan Catanzaro, and Pavlo Molchanov.
\newblock Radiov2.5: Improved baselines for agglomerative vision foundation models.
\newblock In \emph{CVPR}, 2025.

\bibitem[Ho et~al.(2022{\natexlab{a}})Ho, Chan, Saharia, Whang, Gao, Gritsenko, Kingma, Poole, Norouzi, Fleet, et~al.]{ho2022imagen}
Jonathan Ho, William Chan, Chitwan Saharia, Jay Whang, Ruiqi Gao, Alexey Gritsenko, Diederik~P Kingma, Ben Poole, Mohammad Norouzi, David~J Fleet, et~al.
\newblock Imagen video: High definition video generation with diffusion models.
\newblock \emph{arXiv preprint arXiv:2210.02303}, 2022{\natexlab{a}}.

\bibitem[Ho et~al.(2022{\natexlab{b}})Ho, Salimans, Gritsenko, Chan, Norouzi, and Fleet]{ho2022video}
Jonathan Ho, Tim Salimans, Alexey Gritsenko, William Chan, Mohammad Norouzi, and David~J Fleet.
\newblock Video diffusion models.
\newblock In \emph{NeurIPS}, 2022{\natexlab{b}}.

\bibitem[Hong et~al.(2023)Hong, Ding, Zheng, Liu, and Tang]{hong2022cogvideo}
Wenyi Hong, Ming Ding, Wendi Zheng, Xinghan Liu, and Jie Tang.
\newblock Cogvideo: Large-scale pretraining for text-to-video generation via transformers.
\newblock In \emph{ICLR}, 2023.

\bibitem[Hu et~al.(2020)Hu, Cotter, Mohan, Gurau, and Kendall]{hu2020probabilistic}
Anthony Hu, Fergal Cotter, Nikhil Mohan, Corina Gurau, and Alex Kendall.
\newblock Probabilistic future prediction for video scene understanding.
\newblock In \emph{ECCV}, 2020.

\bibitem[Hu et~al.(2023)Hu, Russell, Yeo, Murez, Fedoseev, Kendall, Shotton, and Corrado]{hu2023gaia}
Anthony Hu, Lloyd Russell, Hudson Yeo, Zak Murez, George Fedoseev, Alex Kendall, Jamie Shotton, and Gianluca Corrado.
\newblock Gaia-1: A generative world model for autonomous driving.
\newblock \emph{arXiv preprint arXiv:2309.17080}, 2023.

\bibitem[Hu et~al.(2021)Hu, Sun, Lin, Lai, Zeng, and Zheng]{hu2021apanet}
Jian-Fang Hu, Jiangxin Sun, Zihang Lin, Jian-Huang Lai, Wenjun Zeng, and Wei-Shi Zheng.
\newblock Apanet: Auto-path aggregation for future instance segmentation prediction.
\newblock \emph{IEEE Transactions on Pattern Analysis and Machine Intelligence}, 2021.

\bibitem[Jin et~al.(2017)Jin, Li, Xiao, Shen, Lin, Yang, Chen, Dong, Liu, Jie, et~al.]{jin2017video}
Xiaojie Jin, Xin Li, Huaxin Xiao, Xiaohui Shen, Zhe Lin, Jimei Yang, Yunpeng Chen, Jian Dong, Luoqi Liu, Zequn Jie, et~al.
\newblock Video scene parsing with predictive feature learning.
\newblock In \emph{ICCV}, 2017.

\bibitem[Kakogeorgiou et~al.(2022)Kakogeorgiou, Gidaris, Psomas, Avrithis, Bursuc, Karantzalos, and Komodakis]{AttMask}
Ioannis Kakogeorgiou, Spyros Gidaris, Bill Psomas, Yannis Avrithis, Andrei Bursuc, Konstantinos Karantzalos, and Nikos Komodakis.
\newblock What to hide from your students: Attention-guided masked image modeling.
\newblock In \emph{ECCV}, 2022.

\bibitem[Kakogeorgiou et~al.(2024)Kakogeorgiou, Gidaris, Karantzalos, and Komodakis]{spot_cvpr}
Ioannis Kakogeorgiou, Spyros Gidaris, Konstantinos Karantzalos, and Nikos Komodakis.
\newblock Spot: Self-training with patch-order permutation for object-centric learning with autoregressive transformers.
\newblock In \emph{CVPR}, 2024.

\bibitem[Karypidis et~al.(2025)Karypidis, Kakogeorgiou, Gidaris, and Komodakis]{karypidis2025advancing}
Efstathios Karypidis, Ioannis Kakogeorgiou, Spyros Gidaris, and Nikos Komodakis.
\newblock Advancing semantic future prediction through multimodal visual sequence transformers.
\newblock In \emph{CVPR}, 2025.

\bibitem[Kendall et~al.(2018)Kendall, Gal, and Cipolla]{Kendall2018uncertainty}
Alex Kendall, Yarin Gal, and Roberto Cipolla.
\newblock Multi-task learning using uncertainty to weigh losses for scene geometry and semantics.
\newblock In \emph{CVPR}, 2018.

\bibitem[Kingma and Ba(2015)]{adamopt}
Diederik Kingma and Jimmy Ba.
\newblock Adam: A method for stochastic optimization.
\newblock In \emph{ICLR}, 2015.

\bibitem[Kirillov et~al.(2023)Kirillov, Mintun, Ravi, Mao, Rolland, Gustafson, Xiao, Whitehead, Berg, Lo, et~al.]{kirillov2023segment}
Alexander Kirillov, Eric Mintun, Nikhila Ravi, Hanzi Mao, Chloe Rolland, Laura Gustafson, Tete Xiao, Spencer Whitehead, Alexander~C Berg, Wan-Yen Lo, et~al.
\newblock Segment anything.
\newblock In \emph{CVPR}, 2023.

\bibitem[Kolesnikov et~al.(2020)Kolesnikov, Beyer, Zhai, Puigcerver, Yung, Gelly, and Houlsby]{kolesnikov2020big}
Alexander Kolesnikov, Lucas Beyer, Xiaohua Zhai, Joan Puigcerver, Jessica Yung, Sylvain Gelly, and Neil Houlsby.
\newblock Big transfer (bit): General visual representation learning.
\newblock In \emph{ECCV}, 2020.

\bibitem[Kondratyuk et~al.(2024)Kondratyuk, Yu, Gu, Lezama, Huang, Schindler, Hornung, Birodkar, Yan, Chiu, Somandepalli, Akbari, Alon, Cheng, Dillon, Gupta, Hahn, Hauth, Hendon, Martinez, Minnen, Sirotenko, Sohn, Yang, Adam, Yang, Essa, Wang, Ross, Seybold, and Jiang]{kondratyuk2024videopoet}
Dan Kondratyuk, Lijun Yu, Xiuye Gu, Jose Lezama, Jonathan Huang, Grant Schindler, Rachel Hornung, Vighnesh Birodkar, Jimmy Yan, Ming-Chang Chiu, Krishna Somandepalli, Hassan Akbari, Yair Alon, Yong Cheng, Joshua~V. Dillon, Agrim Gupta, Meera Hahn, Anja Hauth, David Hendon, Alonso Martinez, David Minnen, Mikhail Sirotenko, Kihyuk Sohn, Xuan Yang, Hartwig Adam, Ming-Hsuan Yang, Irfan Essa, Huisheng Wang, David~A Ross, Bryan Seybold, and Lu Jiang.
\newblock Videopoet: A large language model for zero-shot video generation.
\newblock In \emph{ICML}, 2024.

\bibitem[Lee et~al.(2018)Lee, Zhang, Ebert, Abbeel, Finn, and Levine]{lee2018stochastic}
Alex~X Lee, Richard Zhang, Frederik Ebert, Pieter Abbeel, Chelsea Finn, and Sergey Levine.
\newblock Stochastic adversarial video prediction.
\newblock \emph{arXiv preprint arXiv:1804.01523}, 2018.

\bibitem[Lee et~al.(2021)Lee, Kim, Choi, Kim, and Ro]{lee2021video}
Sangmin Lee, Hak~Gu Kim, Dae~Hwi Choi, Hyung-Il Kim, and Yong~Man Ro.
\newblock Video prediction recalling long-term motion context via memory alignment learning.
\newblock In \emph{CVPR}, 2021.

\bibitem[Li et~al.(2024)Li, Tian, Li, Deng, and He]{li2024autoregressive}
Tianhong Li, Yonglong Tian, He Li, Mingyang Deng, and Kaiming He.
\newblock Autoregressive image generation without vector quantization.
\newblock \emph{NeurIPS}, 2024.

\bibitem[Liang et~al.(2022)Liang, Wu, Han, Xu, Xu, and Liang]{liang2022effective}
Xiwen Liang, Yangxin Wu, Jianhua Han, Hang Xu, Chunjing Xu, and Xiaodan Liang.
\newblock Effective adaptation in multi-task co-training for unified autonomous driving.
\newblock In \emph{NeurIPS}, 2022.

\bibitem[Lin et~al.(2017)Lin, Doll{\'a}r, Girshick, He, Hariharan, and Belongie]{lin2017feature}
Tsung-Yi Lin, Piotr Doll{\'a}r, Ross Girshick, Kaiming He, Bharath Hariharan, and Serge Belongie.
\newblock Feature pyramid networks for object detection.
\newblock In \emph{CVPR}, 2017.

\bibitem[Lin et~al.(2021)Lin, Sun, Hu, Yu, Lai, and Zheng]{lin2021predictive}
Zihang Lin, Jiangxin Sun, Jian-Fang Hu, Qizhi Yu, Jian-Huang Lai, and Wei-Shi Zheng.
\newblock Predictive feature learning for future segmentation prediction.
\newblock In \emph{CVPR}, 2021.

\bibitem[Liu et~al.(2022)Liu, MA, Tian, He, and Kira]{NEURIPS2022_efb02f96}
Yen-Cheng Liu, CHIH-YAO MA, Junjiao Tian, Zijian He, and Zsolt Kira.
\newblock Polyhistor: Parameter-efficient multi-task adaptation for dense vision tasks.
\newblock In \emph{NeurIPS}, 2022.

\bibitem[Long et~al.(2015)Long, Shelhamer, and Darrell]{Long_2015_CVPR}
Jonathan Long, Evan Shelhamer, and Trevor Darrell.
\newblock Fully convolutional networks for semantic segmentation.
\newblock In \emph{CVPR}, 2015.

\bibitem[Lu et~al.(2017)Lu, Kumar, Zhai, Cheng, Javidi, and Feris]{8099609}
Yongxi Lu, Abhishek Kumar, Shuangfei Zhai, Yu Cheng, Tara Javidi, and Rogerio Feris.
\newblock Fully-adaptive feature sharing in multi-task networks with applications in person attribute classification.
\newblock In \emph{CVPR}, 2017.

\bibitem[Luc et~al.(2017)Luc, Neverova, Couprie, Verbeek, and LeCun]{luc2017predicting}
Pauline Luc, Natalia Neverova, Camille Couprie, Jakob Verbeek, and Yann LeCun.
\newblock Predicting deeper into the future of semantic segmentation.
\newblock In \emph{ICCV}, 2017.

\bibitem[Luc et~al.(2018)Luc, Couprie, Lecun, and Verbeek]{luc2018predicting}
Pauline Luc, Camille Couprie, Yann Lecun, and Jakob Verbeek.
\newblock Predicting future instance segmentation by forecasting convolutional features.
\newblock In \emph{ECCV}, 2018.

\bibitem[Maninis et~al.(2019)Maninis, Radosavovic, and Kokkinos]{MRK19}
Kevis-Kokitsi Maninis, Ilija Radosavovic, and Iasonas Kokkinos.
\newblock Attentive single-tasking of multiple tasks.
\newblock In \emph{CVPR}, 2019.

\bibitem[Misra et~al.(2016)Misra, Shrivastava, Gupta, and Hebert]{Misra2016cross}
Ishan Misra, Abhinav Shrivastava, Abhinav Gupta, and Martial Hebert.
\newblock Cross-stitch networks for multi-task learning.
\newblock In \emph{CVPR}, 2016.

\bibitem[Nabavi et~al.(2018)Nabavi, Rochan, and Wang]{nabavi2018future}
Seyed~Shahabeddin Nabavi, Mrigank Rochan, and Yang Wang.
\newblock Future semantic segmentation with convolutional lstm.
\newblock In \emph{BMVC}, 2018.

\bibitem[Neseem et~al.(2023)Neseem, Agiza, and Reda]{adamtl2023}
Marina Neseem, Ahmed Agiza, and Sherief Reda.
\newblock { AdaMTL: Adaptive Input-dependent Inference for Efficient Multi-Task Learning}.
\newblock In \emph{CVPRW}, 2023.

\bibitem[Oquab et~al.(2024)Oquab, Darcet, Moutakanni, Vo, Szafraniec, Khalidov, Fernandez, HAZIZA, Massa, El-Nouby, Assran, Ballas, Galuba, Howes, Huang, Li, Misra, Rabbat, Sharma, Synnaeve, Xu, Jegou, Mairal, Labatut, Joulin, and Bojanowski]{oquab2024dinov}
Maxime Oquab, Timoth{\'e}e Darcet, Th{\'e}o Moutakanni, Huy~V. Vo, Marc Szafraniec, Vasil Khalidov, Pierre Fernandez, Daniel HAZIZA, Francisco Massa, Alaaeldin El-Nouby, Mido Assran, Nicolas Ballas, Wojciech Galuba, Russell Howes, Po-Yao Huang, Shang-Wen Li, Ishan Misra, Michael Rabbat, Vasu Sharma, Gabriel Synnaeve, Hu Xu, Herve Jegou, Julien Mairal, Patrick Labatut, Armand Joulin, and Piotr Bojanowski.
\newblock {DINO}v2: Learning robust visual features without supervision.
\newblock \emph{Transactions on Machine Learning Research}, 2024.

\bibitem[Radford et~al.(2021)Radford, Kim, Hallacy, Ramesh, Goh, Agarwal, Sastry, Askell, Mishkin, Clark, et~al.]{radford2021learning}
Alec Radford, Jong~Wook Kim, Chris Hallacy, Aditya Ramesh, Gabriel Goh, Sandhini Agarwal, Girish Sastry, Amanda Askell, Pamela Mishkin, Jack Clark, et~al.
\newblock Learning transferable visual models from natural language supervision.
\newblock In \emph{ICML}, 2021.

\bibitem[Ranftl et~al.(2021)Ranftl, Bochkovskiy, and Koltun]{ranftl2021vision}
Ren{\'e} Ranftl, Alexey Bochkovskiy, and Vladlen Koltun.
\newblock Vision transformers for dense prediction.
\newblock In \emph{CVPR}, 2021.

\bibitem[Razavi et~al.(2019)Razavi, van~den Oord, and Vinyals]{vqvae_2}
Ali Razavi, Aaron van~den Oord, and Oriol Vinyals.
\newblock Generating diverse high-fidelity images with vq-vae-2.
\newblock In \emph{NeurIPS}, 2019.

\bibitem[Rombach et~al.(2022)Rombach, Blattmann, Lorenz, Esser, and Ommer]{rombach2022high}
Robin Rombach, Andreas Blattmann, Dominik Lorenz, Patrick Esser, and Bj{\"o}rn Ommer.
\newblock High-resolution image synthesis with latent diffusion models.
\newblock In \emph{CVPR}, 2022.

\bibitem[Ruder et~al.(2019)Ruder, Bingel, Augenstein, and S\o{}gaard]{Ruder_Bingel_Augenstein_Søgaard_2019}
Sebastian Ruder, Joachim Bingel, Isabelle Augenstein, and Anders S\o{}gaard.
\newblock Latent multi-task architecture learning.
\newblock In \emph{AAAI}, 2019.

\bibitem[Ryali et~al.(2023)Ryali, Hu, Bolya, Wei, Fan, Huang, Aggarwal, Chowdhury, Poursaeed, Hoffman, Malik, Li, and Feichtenhofer]{ryali2023hiera}
Chaitanya Ryali, Yuan-Ting Hu, Daniel Bolya, Chen Wei, Haoqi Fan, Po-Yao Huang, Vaibhav Aggarwal, Arkabandhu Chowdhury, Omid Poursaeed, Judy Hoffman, Jitendra Malik, Yanghao Li, and Christoph Feichtenhofer.
\newblock Hiera: a hierarchical vision transformer without the bells-and-whistles.
\newblock In \emph{ICML}, 2023.

\bibitem[{\v{S}}ari{\'c} et~al.(2019){\v{S}}ari{\'c}, Or{\v{s}}i{\'c}, Antunovi{\'c}, Vra{\v{z}}i{\'c}, and {\v{S}}egvi{\'c}]{vsaric2019single}
Josip {\v{S}}ari{\'c}, Marin Or{\v{s}}i{\'c}, Ton{\'c}i Antunovi{\'c}, Sacha Vra{\v{z}}i{\'c}, and Sini{\v{s}}a {\v{S}}egvi{\'c}.
\newblock Single level feature-to-feature forecasting with deformable convolutions.
\newblock In \emph{41st DAGM German Conference on Pattern Recognition}, 2019.

\bibitem[Saric et~al.(2020)Saric, Orsic, Antunovic, Vrazic, and Segvic]{saric2020warp}
Josip Saric, Marin Orsic, Tonci Antunovic, Sacha Vrazic, and Sinisa Segvic.
\newblock Warp to the future: Joint forecasting of features and feature motion.
\newblock In \emph{CVPR}, 2020.

\bibitem[Sener and Koltun(2018)]{NEURIPS2018_432aca3a}
Ozan Sener and Vladlen Koltun.
\newblock Multi-task learning as multi-objective optimization.
\newblock In \emph{NeurIPS}, 2018.

\bibitem[Sirko-Galouchenko et~al.(2025)Sirko-Galouchenko, Gidaris, Vobecky, Bursuc, and Thome]{sirko2025dip}
Sophia Sirko-Galouchenko, Spyros Gidaris, Antonin Vobecky, Andrei Bursuc, and Nicolas Thome.
\newblock Dip: Unsupervised dense in-context post-training of visual representations.
\newblock In \emph{ICCV}, 2025.

\bibitem[Strudel et~al.(2021)Strudel, Garcia, Laptev, and Schmid]{strudel2021segmenter}
Robin Strudel, Ricardo Garcia, Ivan Laptev, and Cordelia Schmid.
\newblock Segmenter: Transformer for semantic segmentation.
\newblock In \emph{CVPR}, 2021.

\bibitem[Sun et~al.(2019)Sun, Xie, Hu, Lin, Lai, Zeng, and Zheng]{sun2019predicting}
Jiangxin Sun, Jiafeng Xie, Jian-Fang Hu, Zihang Lin, Jianhuang Lai, Wenjun Zeng, and Wei-Shi Zheng.
\newblock Predicting future instance segmentation with contextual pyramid convlstms.
\newblock In \emph{Proceedings of the 27th ACM International Conference on Multimedia}, 2019.

\bibitem[Sun et~al.(2023)Sun, Fang, Wu, Wang, and Cao]{sun2023eva}
Quan Sun, Yuxin Fang, Ledell Wu, Xinlong Wang, and Yue Cao.
\newblock Eva-clip: Improved training techniques for clip at scale.
\newblock \emph{arXiv preprint arXiv:2303.15389}, 2023.

\bibitem[Terwilliger et~al.(2019)Terwilliger, Brazil, and Liu]{terwilliger2019recurrent}
Adam Terwilliger, Garrick Brazil, and Xiaoming Liu.
\newblock Recurrent flow-guided semantic forecasting.
\newblock In \emph{WACV}, 2019.

\bibitem[Tong et~al.(2022)Tong, Song, Wang, and Wang]{tong2022videomae}
Zhan Tong, Yibing Song, Jue Wang, and Limin Wang.
\newblock Video{MAE}: Masked autoencoders are data-efficient learners for self-supervised video pre-training.
\newblock In \emph{NeurIPS}, 2022.

\bibitem[Touvron et~al.(2019)Touvron, Vedaldi, Douze, and Jegou]{touvron2019fixing}
Hugo Touvron, Andrea Vedaldi, Matthijs Douze, and Herve Jegou.
\newblock Fixing the train-test resolution discrepancy.
\newblock In \emph{NeurIPS}, 2019.

\bibitem[Tschannen et~al.(2024)Tschannen, Eastwood, and Mentzer]{tschannen2024givt}
Michael Tschannen, Cian Eastwood, and Fabian Mentzer.
\newblock Givt: Generative infinite-vocabulary transformers.
\newblock In \emph{ECCV}, 2024.

\bibitem[Vandenhende et~al.(2022)Vandenhende, Georgoulis, Van~Gansbeke, Proesmans, Dai, and Van~Gool]{9336293}
Simon Vandenhende, Stamatios Georgoulis, Wouter Van~Gansbeke, Marc Proesmans, Dengxin Dai, and Luc Van~Gool.
\newblock Multi-task learning for dense prediction tasks: A survey.
\newblock \emph{IEEE Transactions on Pattern Analysis and Machine Intelligence}, 2022.

\bibitem[Venkataramanan et~al.(2025)Venkataramanan, Pariza, Salehi, Knobel, Gidaris, Ramzi, Bursuc, and Asano]{venkataramanan2025franca}
Shashanka Venkataramanan, Valentinos Pariza, Mohammadreza Salehi, Lukas Knobel, Spyros Gidaris, Elias Ramzi, Andrei Bursuc, and Yuki~M Asano.
\newblock Franca: Nested matryoshka clustering for scalable visual representation learning.
\newblock \emph{arXiv preprint arXiv:2507.14137}, 2025.

\bibitem[Vondrick et~al.(2016{\natexlab{a}})Vondrick, Pirsiavash, and Torralba]{vondrick2016anticipating}
Carl Vondrick, Hamed Pirsiavash, and Antonio Torralba.
\newblock Anticipating visual representations from unlabeled video.
\newblock In \emph{CVPR}, 2016{\natexlab{a}}.

\bibitem[Vondrick et~al.(2016{\natexlab{b}})Vondrick, Pirsiavash, and Torralba]{vondrick2016generating}
Carl Vondrick, Hamed Pirsiavash, and Antonio Torralba.
\newblock Generating videos with scene dynamics.
\newblock In \emph{NeurIPS}, 2016{\natexlab{b}}.

\bibitem[Vora et~al.(2018)Vora, Mahjourian, Pirk, and Angelova]{FeatReproj3D}
Suhani Vora, Reza Mahjourian, Soeren Pirk, and Anelia Angelova.
\newblock Future semantic segmentation using 3d structure.
\newblock \emph{arXiv preprint arXiv:1811.11358}, 2, 2018.

\bibitem[Wang et~al.(2023)Wang, Huang, Zhao, Tong, He, Wang, Wang, and Qiao]{wang2023videomaev2}
Limin Wang, Bingkun Huang, Zhiyu Zhao, Zhan Tong, Yinan He, Yi Wang, Yali Wang, and Yu Qiao.
\newblock Videomae v2: Scaling video masked autoencoders with dual masking.
\newblock In \emph{CVPR}, 2023.

\bibitem[Wang et~al.(2024)Wang, Zhang, Luo, Sun, Cui, Wang, Zhang, Wang, Li, Yu, et~al.]{wang2024emu3}
Xinlong Wang, Xiaosong Zhang, Zhengxiong Luo, Quan Sun, Yufeng Cui, Jinsheng Wang, Fan Zhang, Yueze Wang, Zhen Li, Qiying Yu, et~al.
\newblock Emu3: Next-token prediction is all you need.
\newblock \emph{arXiv preprint arXiv:2409.18869}, 2024.

\bibitem[Wang et~al.(2018)Wang, Jiang, Yang, Li, Long, and Fei-Fei]{wang2018eidetic}
Yunbo Wang, Lu Jiang, Ming-Hsuan Yang, Li-Jia Li, Mingsheng Long, and Li Fei-Fei.
\newblock Eidetic 3d lstm: A model for video prediction and beyond.
\newblock In \emph{ICLR}, 2018.

\bibitem[Wei et~al.(2022)Wei, Fan, Xie, Wu, Yuille, and Feichtenhofer]{wei2022masked}
Chen Wei, Haoqi Fan, Saining Xie, Chao-Yuan Wu, Alan Yuille, and Christoph Feichtenhofer.
\newblock Masked feature prediction for self-supervised visual pre-training.
\newblock In \emph{CVPR}, 2022.

\bibitem[Wu et~al.(2021)Wu, Yao, Wang, and Long]{wu2021motionrnn}
Haixu Wu, Zhiyu Yao, Jianmin Wang, and Mingsheng Long.
\newblock Motionrnn: A flexible model for video prediction with spacetime-varying motions.
\newblock In \emph{CVPR}, 2021.

\bibitem[Wu et~al.(2019)Wu, Kirillov, Massa, Lo, and Girshick]{wu2019detectron2}
Yuxin Wu, Alexander Kirillov, Francisco Massa, Wan-Yen Lo, and Ross Girshick.
\newblock Detectron2.
\newblock \url{https://github.com/facebookresearch/detectron2}, 2019.

\bibitem[Xu et~al.(2018)Xu, Ni, Li, Cheng, and Yang]{Xu_2018_CVPR}
Jingwei Xu, Bingbing Ni, Zefan Li, Shuo Cheng, and Xiaokang Yang.
\newblock Structure preserving video prediction.
\newblock In \emph{CVPR}, 2018.

\bibitem[Yan et~al.(2021)Yan, Zhang, Abbeel, and Srinivas]{yan2021videogpt}
Wilson Yan, Yunzhi Zhang, Pieter Abbeel, and Aravind Srinivas.
\newblock Videogpt: Video generation using vq-vae and transformers.
\newblock \emph{arXiv preprint arXiv:2104.10157}, 2021.

\bibitem[Yang et~al.(2024{\natexlab{a}})Yang, Kang, Huang, Xu, Feng, and Zhao]{yang_depth}
Lihe Yang, Bingyi Kang, Zilong Huang, Xiaogang Xu, Jiashi Feng, and Hengshuang Zhao.
\newblock Depth anything: Unleashing the power of large-scale unlabeled data.
\newblock In \emph{CVPR}, 2024{\natexlab{a}}.

\bibitem[Yang et~al.(2024{\natexlab{b}})Yang, Kang, Huang, Zhao, Xu, Feng, and Zhao]{yang2024depth}
Lihe Yang, Bingyi Kang, Zilong Huang, Zhen Zhao, Xiaogang Xu, Jiashi Feng, and Hengshuang Zhao.
\newblock Depth anything v2.
\newblock In \emph{NeurIPS}, 2024{\natexlab{b}}.

\bibitem[Yu et~al.(2023)Yu, Cheng, Sohn, Lezama, Zhang, Chang, Hauptmann, Yang, Hao, Essa, et~al.]{yu2023magvit}
Lijun Yu, Yong Cheng, Kihyuk Sohn, Jos{\'e} Lezama, Han Zhang, Huiwen Chang, Alexander~G Hauptmann, Ming-Hsuan Yang, Yuan Hao, Irfan Essa, et~al.
\newblock Magvit: Masked generative video transformer.
\newblock In \emph{CVPR}, 2023.

\bibitem[Yu et~al.(2024)Yu, Lezama, Gundavarapu, Versari, Sohn, Minnen, Cheng, Gupta, Gu, Hauptmann, Gong, Yang, Essa, Ross, and Jiang]{yu2024language}
Lijun Yu, Jose Lezama, Nitesh~Bharadwaj Gundavarapu, Luca Versari, Kihyuk Sohn, David Minnen, Yong Cheng, Agrim Gupta, Xiuye Gu, Alexander~G Hauptmann, Boqing Gong, Ming-Hsuan Yang, Irfan Essa, David~A Ross, and Lu Jiang.
\newblock Language model beats diffusion - tokenizer is key to visual generation.
\newblock In \emph{ICLR}, 2024.

\bibitem[Zhang et~al.(2022)Zhang, Liu, and Guan]{NEURIPS2022_dd3bd4e3}
Lijun Zhang, Xiao Liu, and Hui Guan.
\newblock Automtl: A programming framework for automating efficient multi-task learning.
\newblock In \emph{NeurIPS}, 2022.

\bibitem[Zhao et~al.(2017)Zhao, Shi, Qi, Wang, and Jia]{zhao2017pyramid}
Hengshuang Zhao, Jianping Shi, Xiaojuan Qi, Xiaogang Wang, and Jiaya Jia.
\newblock Pyramid scene parsing network.
\newblock In \emph{CVPR}, 2017.

\bibitem[Zheng et~al.(2024)Zheng, Song, Guo, Zhang, and Chen]{zheng2024genad}
Wenzhao Zheng, Ruiqi Song, Xianda Guo, Chenming Zhang, and Long Chen.
\newblock Genad: Generative end-to-end autonomous driving.
\newblock In \emph{ECCV}, 2024.

\bibitem[Zhong et~al.(2023)Zhong, Schneider, Voit, Stiefelhagen, and Beyerer]{zhong2023anticipative}
Zeyun Zhong, David Schneider, Michael Voit, Rainer Stiefelhagen, and J{\"u}rgen Beyerer.
\newblock Anticipative feature fusion transformer for multi-modal action anticipation.
\newblock In \emph{WACV}, 2023.

\bibitem[Zhou et~al.(2025)Zhou, Pan, LeCun, and Pinto]{zhou2025dinowm}
Gaoyue Zhou, Hengkai Pan, Yann LeCun, and Lerrel Pinto.
\newblock {DINO}-{WM}: World models on pre-trained visual features enable zero-shot planning.
\newblock In \emph{ICML}, 2025.

\end{thebibliography}
